\documentclass[10pt,twocolumn,letterpaper]{article}
\newcommand{\datasetname}{SAGI-D }

\usepackage[pagenumbers]{iccv} 
\usepackage[accsupp]{axessibility}

\usepackage{pifont} 
\usepackage{dashrule} 
\usepackage{svg}
\usepackage{soul}
\definecolor{trufor_color}{rgb}{0, 0, 1} 
\definecolor{catnet_color}{rgb}{1, 0, 0} 
\definecolor{mmfusion_color}{rgb}{0, 0.5, 0} 
\definecolor{pscc_color}{rgb}{0.5, 0, 0.5} 

\usepackage{verbatim}
\usepackage{graphicx}
\usepackage{subcaption}
\usepackage{multirow}

\usepackage[most]{tcolorbox}
\usepackage{makecell}
\usepackage{tabularx}

\definecolor{iccvblue}{rgb}{0.21,0.49,0.74}
\usepackage[pagebackref,breaklinks,colorlinks,allcolors=iccvblue]{hyperref}
\usepackage{balance}

\def\paperID{8326} 
\def\confName{ICCV}
\def\confYear{2025}

\title{SAGI: Semantically Aligned and Uncertainty Guided AI Image Inpainting}

\author{
Paschalis Giakoumoglou$^{1,2}$ \quad Dimitrios Karageorgiou$^{2}$ \\ Symeon Papadopoulos$^{2}$ \quad Panagiotis C. Petrantonakis$^{1}$ \vspace{4pt}\\
$^{1}$Department of Electrical and Computer Engineering, Aristotle University of Thessaloniki \\
$^{2}$Information Technologies Institute, CERTH \\
{\tt\small \{giakoupg,ppetrant\}@ece.auth.gr, \{giakoupg,dkarageo,papadop\}@iti.gr}
}

\begin{document}
\maketitle
\begin{abstract}

Recent advancements in generative AI have made text-guided image inpainting—adding, removing, or altering image regions using textual prompts—widely accessible. However, generating semantically correct photorealistic imagery, typically requires carefully-crafted prompts and iterative refinement by evaluating the realism of the generated content - tasks commonly performed by humans. To automate the generative process, we propose  Semantically Aligned and Uncertainty Guided AI Image Inpainting (SAGI), a model-agnostic pipeline,  to sample prompts from a distribution that closely aligns with human perception and to evaluate the generated content and discard instances that deviate from such a distribution, which we approximate using pretrained large language models and vision-language models. By applying this pipeline on multiple state-of-the-art inpainting models, we create the SAGI \textbf{D}ataset (\emph{SAGI-D}), currently the largest and most diverse dataset of AI-generated inpaintings, comprising over 95k inpainted images and a human-evaluated subset. Our experiments show that semantic alignment significantly improves image quality and aesthetics, while uncertainty guidance effectively identifies realistic manipulations — human ability to distinguish inpainted images from real ones drops from 74\% to 35\% in terms of accuracy, after applying our pipeline. Moreover, using \emph{SAGI-D} for training several image forensic approaches increases in-domain detection performance on average by 37.4\% and out-of-domain generalization by 26.1\% in terms of IoU, also demonstrating its utility in countering malicious exploitation of generative AI. Code and dataset are available at \url{https://mever-team.github.io/SAGI/}

\end{abstract}

\section{Introduction}
\label{sec:intro}

\begin{figure*}[!htb]
    \centering
    \includegraphics[width=1\textwidth]{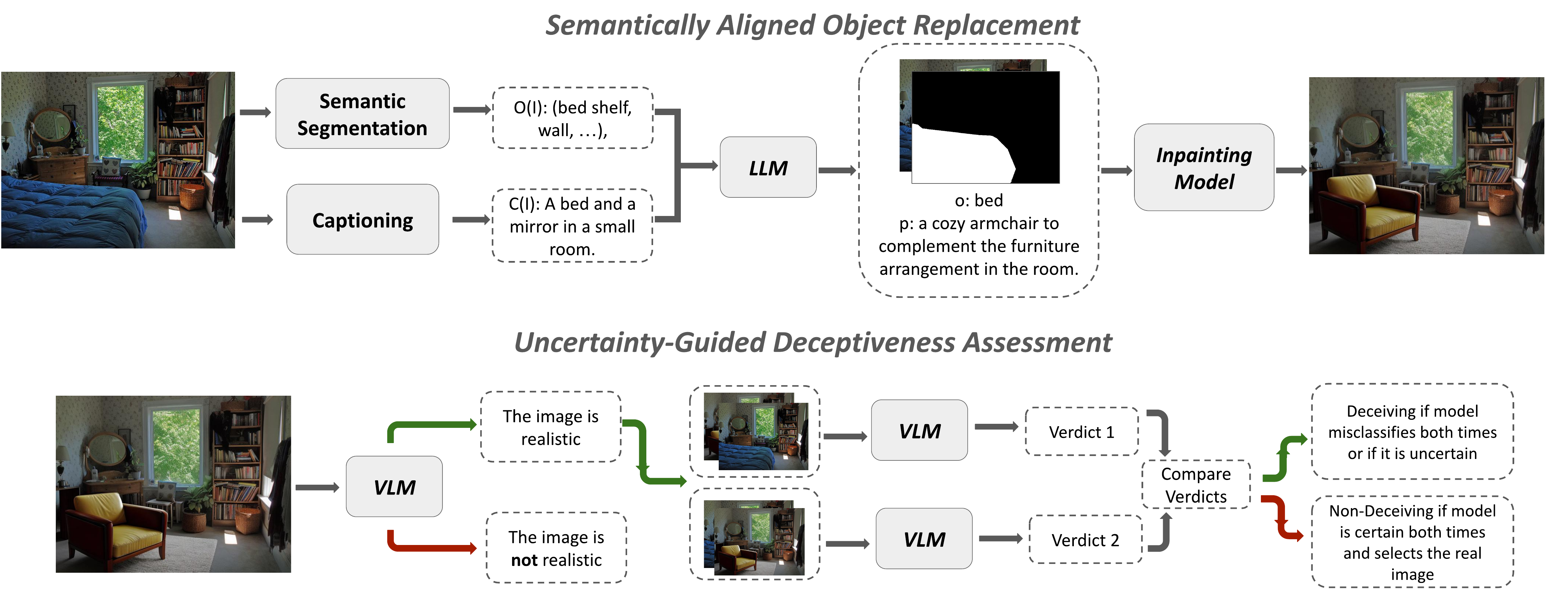}  
    \caption{The first row illustrates SAOR: an input image is processed to identify objects (via labels or segmentation) and generate a caption. The language model selects an object and generates a prompt, which, along with the image and mask, guides the inpainting model to produce the result. The second row depicts UGDA: if the inpainted image passes an initial realism check, it undergoes a second evaluation. The VLM compares the inpainted and original images twice, in reverse order. If the VLM's responses differ (indicating uncertainty) or both favor the inpainted image, it is labeled as deceiving; otherwise, non-deceiving.}
    \label{fig:full_pipeline} 
\vspace{-12pt}
\end{figure*}

Image inpainting—reconstructing image regions—became highly accessible with powerful generative AI tools, enabling non-experts to create photorealistic edits \cite{zhang2024texttoimagediffusionmodelsgenerative}. Text-guided inpainting, that adds, removes, or alters regions using textual prompts, has advanced significantly through models like Stable Diffusion \cite{rombach2022high}, DALL-E \cite{ramesh2021zero}, and Imagen \cite{saharia2022photorealistic}. Achieving high-quality results often requires several attempts and iterative refinement \cite{goloujeh2024isitaiorme}, as advanced models struggle with complex prompts \cite{hessen2024advancesinaigen}. However, successful generations can be so realistic that humans struggle to distinguish them from real photographs \cite{hessen2024advancesinaigen}. \looseness=-1 

Automated inpainting generation faces two key challenges. First, while state-of-the-art models excel at manipulation, they require detailed prompts for realistic results—basic object labels often lack sufficient context \cite{mahajan2023promptinghardhardlyprompting}. Second, automating realism assessment is difficult, as standard quality metrics may not reflect perceived realism. To address these, we propose SAGI (Semantically Aligned and Uncertainty Guided AI Image Inpainting), a unified framework integrating Semantically Aligned Object Replacement (SAOR) and Uncertainty Guided Deceptiveness Assessment (UGDA). SAOR leverages image semantics and large language models (LLMs) to create context-aware prompts, producing higher aesthetic quality than simple object labels and captions. UGDA is a realism assessment method that uses vision-language models (VLMs) to compare inpainted images with originals, identifying convincing manipulations. We validate this approach through a user study, showing it aligns with human perception of image realism. An overview of the approach is presented in \cref{fig:full_pipeline}. 

Using this framework along with multiple state-of-the-art inpainting models, we create the \datasetname dataset, the largest and most diverse collection of AI-generated inpaintings to date, establishing a new benchmark for inpainting detection research. The dataset contains 95,839 inpainted images from 78,684 originals across three datasets: MS-COCO \cite{lin2014coco}, RAISE \cite{dang2015raise}, and OpenImages \cite{kuznetsova2020open}. We provide each inpainted image with its original version, inpainting mask, and text prompt. For evaluation, we create both in-domain and out-of-domain testing splits. The in-domain split uses the same LLM and source images as the training set, for familiar data assessment. The out-of-domain split uses different source images and a different LLM, testing generalization to new data. Beyond advancing inpainting generation, the resulting dataset provides a valuable resource for training and evaluating forensic detection methods that can address challenges regarding potential malicious misuse \cite{verdoliva2020media, wu2022defakehop}.

Our main contributions are summarized as follows:

\begin{itemize} 
\item We propose semantically aligned and uncertainty guided image inpainting, a model-agnostic framework, to automate the generation of realistic inpainted images.
\item We introduce semantically aligned object replacement to generate semantically coherent prompts.
\item We propose uncertainty guided deceptiveness assessment, to estimate the realism of an image at test time. 
\item We experimentally show that the proposed  framework increases quality and aesthetics across several generative models. Through human studies we show that human ability to distinguish inpainted images from real ones drops from 74\% to 35\% after applying our framework.
\item We present the \datasetname dataset, the largest and most diverse dataset of AI-generated inpainted images to date
\item We demonstrate the efficacy of our approach on retraining image forensic methods, enhancing their average in-domain localization performance by 37.4\% and out-of-domain generalization by 26.1\% in terms of IoU.
\end{itemize}

\section{Related Work}
\label{sec:related}


\subsection{Image Inpainting}

Early image inpainting methods used diffusion for image restoration \cite{bertalmio2000pdeinpainting, bertalmio2001pde2, bertalmio2005pde3, chan2001pde} and exemplar-based approaches for object removal \cite{criminisi2004patch, jin2015patch2, kawai2016patch3, guo2018patch4}. Recent methods are deep learning-based, using CNNs \cite{lecun1997cnn}, Fourier Convolutions \cite{suvorov2021lama}, Transformers \cite{vaswani2017attentionneed}, or Diffusion Models \cite{chang2023designfundamentalsdiffusionmodels}. GAN-based methods are also widely used, to generate coherent images, with techniques such as Context Encoders \cite{pathak2016context} and dilated convolutions improving results \cite{yu2018generativeimageinpaintingcontextual}. Diffusion models have emerged as a powerful approach, reconstructing missing regions via noise removal, with models like GLIDE, DALL-E \cite{ramesh2021zeroshottexttoimagegeneration} \cite{nichol2022glidephotorealisticimagegeneration} and Stable Diffusion \cite{rombach2022stablediffusion} incorporating additional guidance, such as text, for improved control. 


\subsection{Inpainting Detection}

Inpainting detection became crucial with inpainting advancements. Early approaches used patch comparison and connectivity analysis to identify inconsistencies \cite{wu2008detection, chang2013forgerydetection, liang2015efficientforgerydetection, zhang2018robustforgery}. Deep learning, particularly CNNs, improved detection via filtering, segmentation networks, and encoder-decoder architectures \cite{zhu2018deeplearningapproachpatchbasedcnn, li2019highpasssfullyconvulutionalresnetcnn, lu2020detectionlstmcnn, kumar2021semanticsegmentioninpainting, simonyan2015deepconvolutionalnetworkslargescalevgg, luy2022pscc}. Transformer-CNN hybrids enhanced long-range dependency and texture analysis \cite{zhu2023transformercnn, zhang2023localizationinpaintingfeatureenhancemnet, wu2022iid}. Recent methods integrate frequency-domain, compression artifacts, noise, and semantic features \cite{kwon2022catnet, triaridis2023mmfusion, guillaro2023trufor, karageorgiou2024fusion}. While effective for traditional inpainting and forgeries like copy-move and splicing, few have been tested on AI-based inpainting, except for evaluations on GLIDE-generated forgeries \cite{guillaro2023trufor, triaridis2023mmfusion, karageorgiou2024fusion}, leaving a gap in assessing performance on modern generative techniques.


\subsection{Datasets for Image Inpainting Detection}

The development and evaluation of inpainting detection models rely on datasets that capture diverse inpainting techniques and scenarios. Early inpainting datasets used outdated inpainting techniques that produced very poor results \cite{dong2013casia, amerini2011micc, christlein2012cmfd, tralic2013cofomod, wu2022iid, mahfoudi2019defacto}. Several recent datasets  \cite{zhong2023aigcdetect, bammey2024synthbuster, chen2023twigmadatasetaigeneratedimages} employ state-of-the-art generative AI models, but they focus on fully synthetic images rather than inpaintings. This creates a significant gap in the available resources for inpainting research. Among these, CocoGlide \cite{guillaro2023trufor} and TGIF \cite{mareen2024tgiftextguidedinpaintingforgery} stand out as the most relevant, utilizing text-to-image inpainting models such as GLIDE, Stable Diffusion, and Adobe Firefly. However, these datasets have notable limitations: CocoGlide is limited in scale, while TGIF, though larger, does not account for more recent and complex inpainting pipelines that produce higher-quality results.

\subsection{Image Quality and Aesthetic Assessment}
Image quality assessment (IQA) evaluates how distortions affect human perception, evolving from handcrafted features \cite{wang2004imagequalityassessment, mittal2012noreferencimagequalityassessment} to deep learning methods like NIMA \cite{talebi2018nima} and HyperIQA \cite{su2020hyperiqablindlyassess}. Recent approaches  utilize CLIP \cite{radford2021clip, zhang2023liqe, wang2022clipiqa}, leverage multi-scale inputs and vision-language correspondence but often underperform purely visual methods due to reliance on visual-text similarity, which prioritizes semantic alignment over visual fidelity. Image aesthetic assessment (IAA) is more complex, focusing on attributes like composition and emotional impact \cite{murray2012alargescale}. Deep learning dominates IAA, with methods like NIMA \cite{talebi2018nima} and VILA \cite{ke2023vilalearningimageaesthetics} achieving state-of-the-art results. QALIGN \cite{wu2023qalignteachinglmmsvisual} advances both IQA and IAA by leveraging VLMs to improve generalization without extensive fine-tuning. However, most approaches are task-specific and struggle with generalization, and none of them focus on realism assessment.
\section{Semantically Aligned and Uncertainty Guided AI Image Inpainting}
\label{sec:dataset}

Recent studies have shown that detailed prompts in AI image generation lead to better inpainting results \cite{rosenman2024neuropromptsadaptiveframeworkoptimize, manas2024improvingtexttoimageconsistencyautomatic}, with users employing complex, carefully crafted prompts \cite{goloujeh2024isitaiorme}, evaluating image semantics \cite{pasupathy2015humansegmentation} and using multiple models and attempts \cite{tang2024exploringimpactaigeneratedimage} until achieving desired results. Inspired by these observations, we argue that automatically generating high-quality inpainted images, deceiving to the human eye, requires addressing two key challenges: ensuring semantic coherence in manipulations and assessing the realism of generated images. To this end, we propose SAGI, a model-agnostic pipeline for a) sampling prompts from a distribution that aligns with human perception of reality and b) assessing and discarding generated samples that deviate from such distribution—techniques notably absent in previous works. We consider pre-trained LLMs and VLMs to be the best available approximations for such distribution, due to their training data scale. We employ them in our automated pipeline with two main components: 1) Semantically Aligned Object Replacement for contextually appropriate manipulations and 2) Uncertainty-Guided Deceptiveness Assessment for evaluating realism.

\begin{figure*}[!htb]
    \centering
    \begin{subfigure}{\textwidth}
        \begin{subfigure}{0.32\textwidth}
            \includegraphics[width=0.49\linewidth]{examples/000_final_paper_good_000000002549_val.jpg}
            \includegraphics[width=0.49\linewidth]{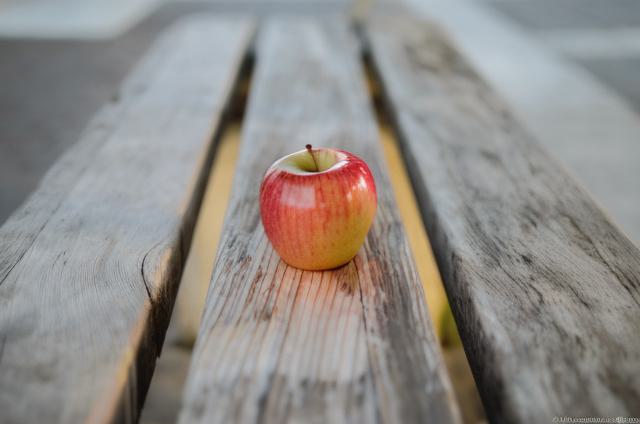}
            \caption{``a delicious apple to add a pop of color to the scene''}
        \end{subfigure}
        \hfill
        \begin{subfigure}{0.32\textwidth}
            \includegraphics[width=0.49\linewidth]{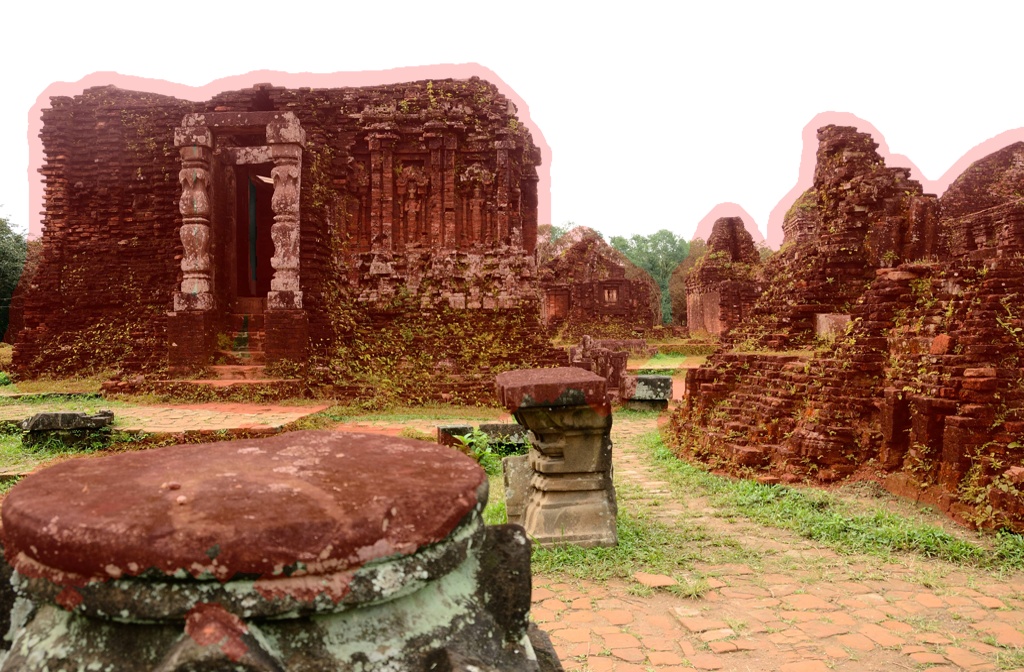}
            \includegraphics[width=0.49\linewidth]{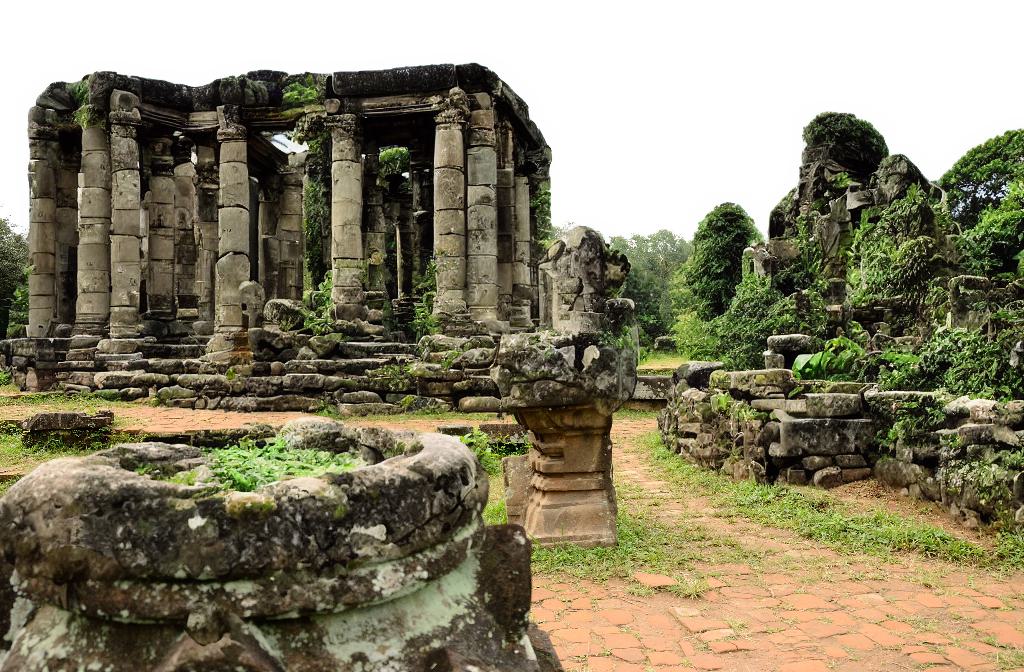}
            \caption{``lush green vines and foliage, intertwining around ancient stone pillars''}
        \end{subfigure}
        \hfill
        \begin{subfigure}{0.32\textwidth}
            \includegraphics[width=0.49\linewidth]{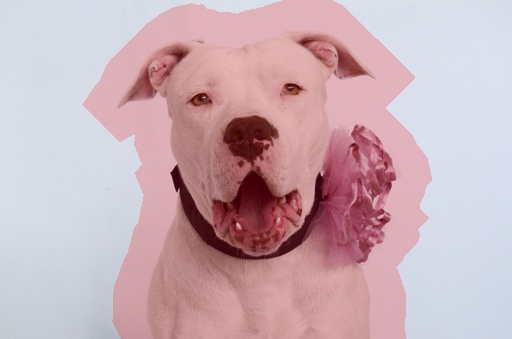}
            \includegraphics[width=0.49\linewidth]{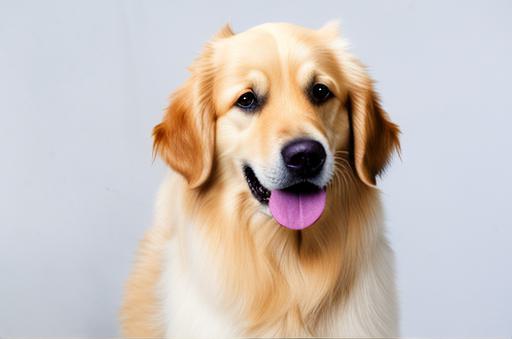}
            \caption{``a fluffy golden retriever''}
        \end{subfigure}
    \end{subfigure}
    
    \vspace{1em}
    
    \begin{subfigure}{\textwidth}
        \begin{subfigure}{0.32\textwidth}
            \includegraphics[width=0.49\linewidth]{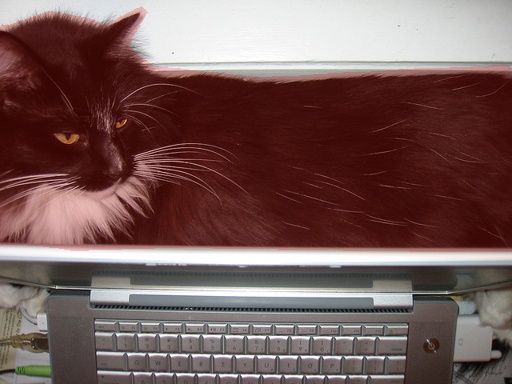}
            \includegraphics[width=0.49\linewidth]{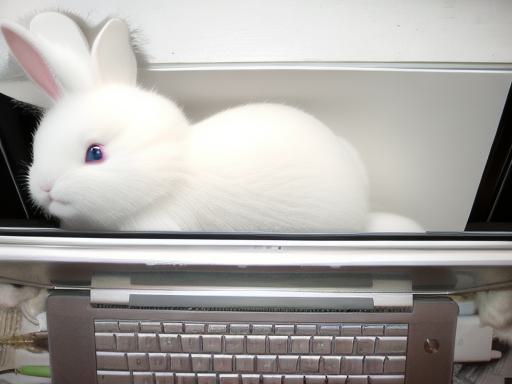}
            \caption{``a fluffy white bunny sitting inside a laptop.''}
        \end{subfigure}
        \hfill
        \begin{subfigure}{0.32\textwidth}
            \includegraphics[width=0.49\linewidth]{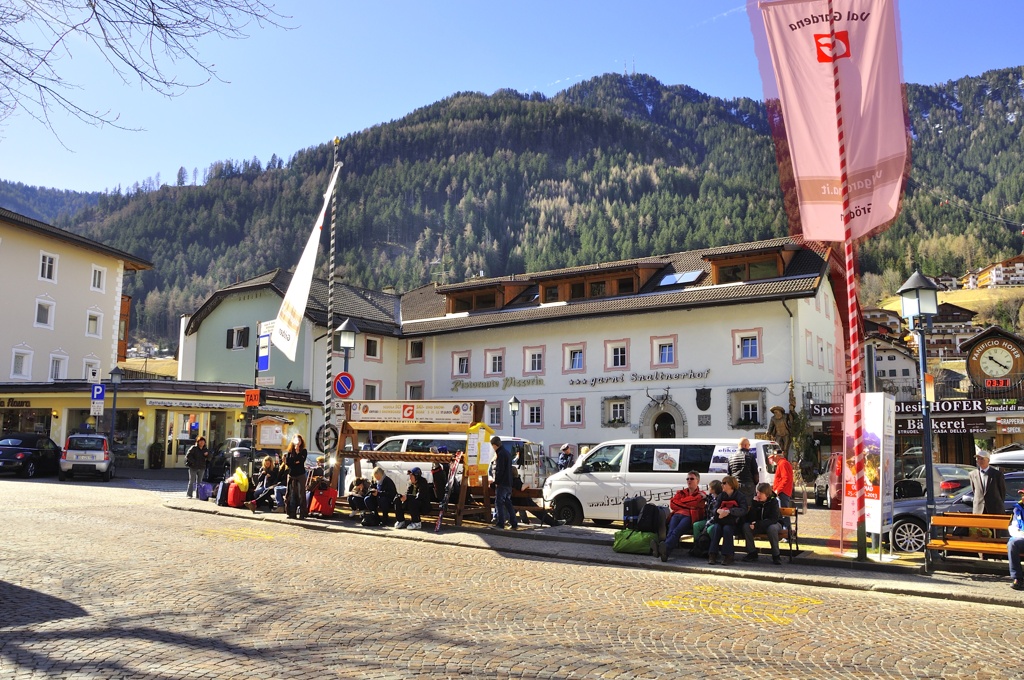}
            \includegraphics[width=0.49\linewidth]{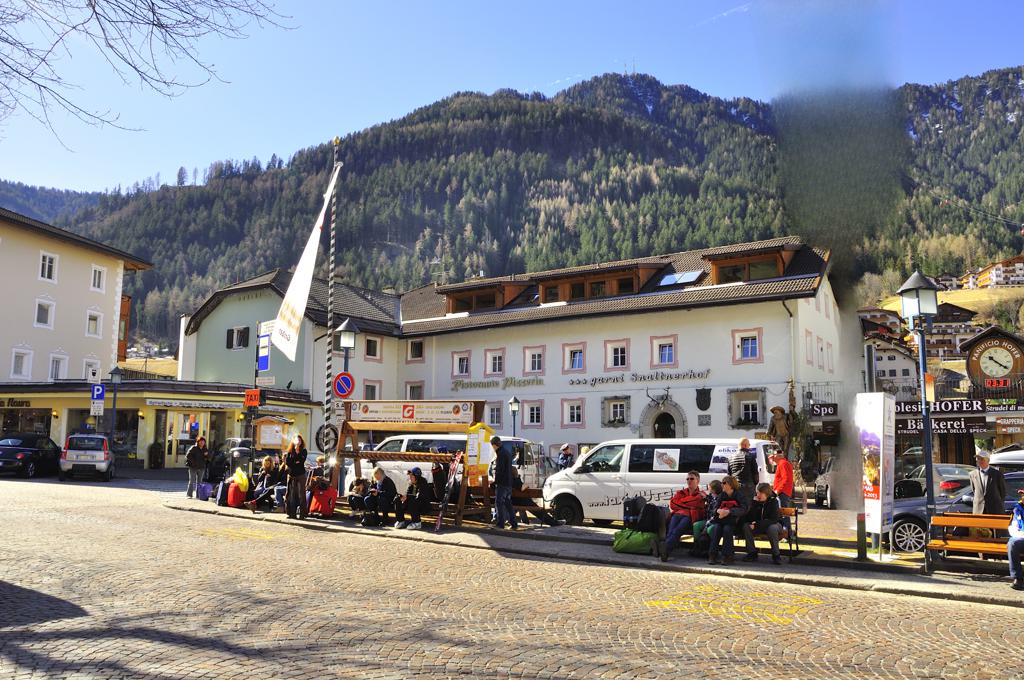}
            \caption{No prompt}
        \end{subfigure}
        \hfill
        \begin{subfigure}{0.32\textwidth}
            \includegraphics[width=0.49\linewidth]{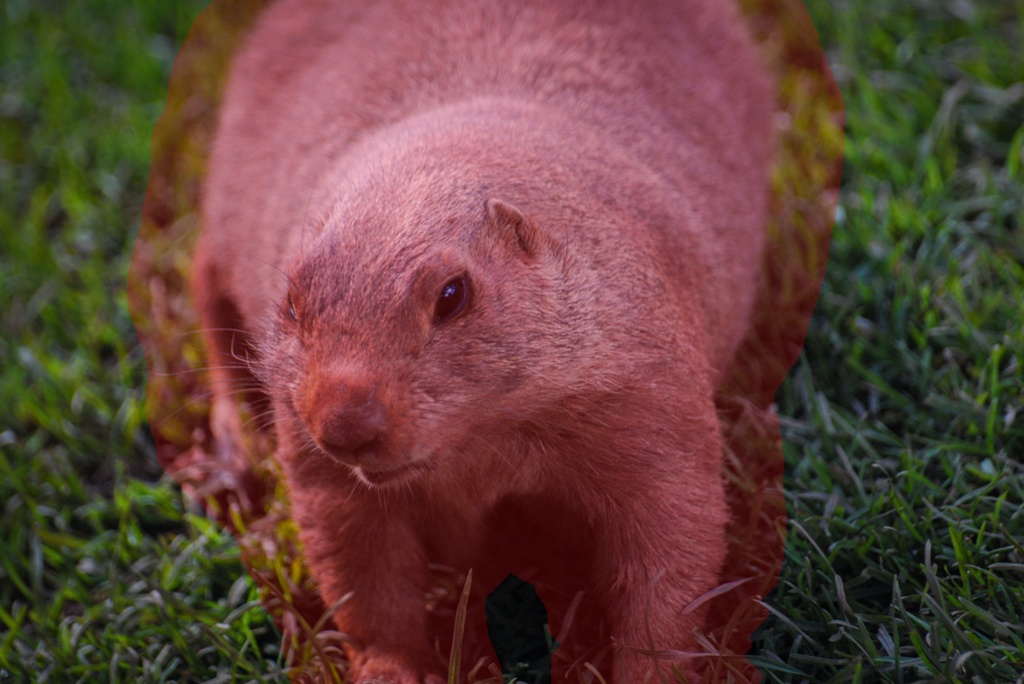}
            \includegraphics[width=0.49\linewidth]{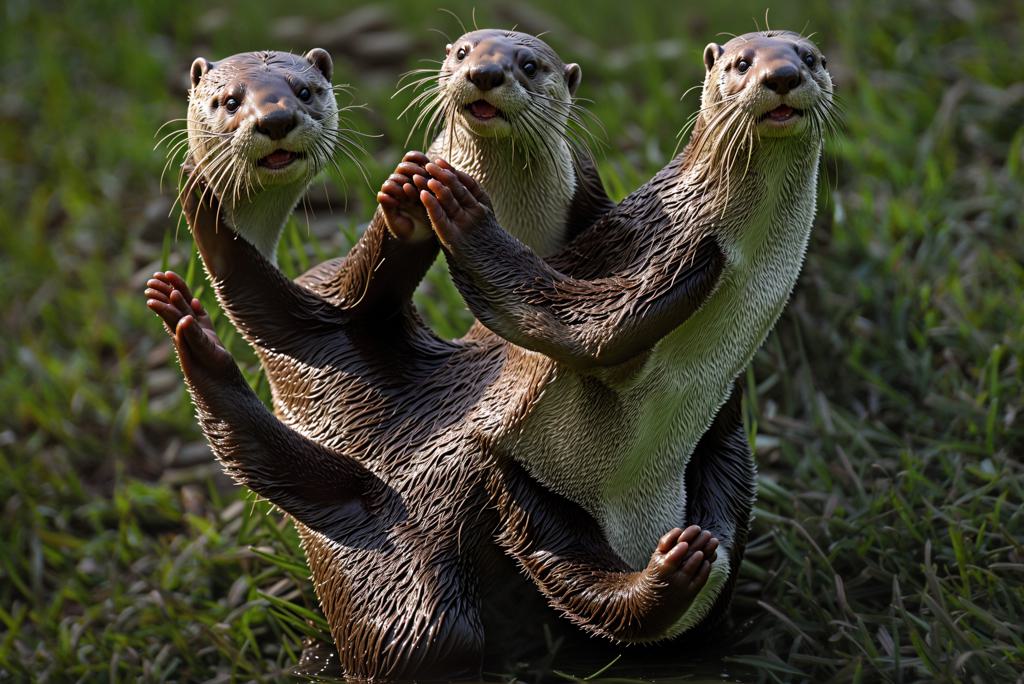}
            \caption{``a curious otter exploring the grassy field''}
        \end{subfigure}
    \end{subfigure}

    \vspace{-4pt}
    \caption{Original (with semi-transparent red inpainting mask) and inpainted images from three datasets, with prompts shown below each pair for text-guided models. The first row shows images classified as deceiving by UGDA \ref{sec:ugda}, and the second row shows non-deceiving images. Each column corresponds to a different dataset: MS-COCO~\cite{lin2014coco} (first), RAISE~\cite{dang2015raise} (second), and OpenImages~\cite{kuznetsova2020open} (third).}
    \label{fig:deceptive_dataset_examples}
    \vspace{-6pt}
\end{figure*}


\subsection{Semantically Aligned Object Replacement}
\label{sec:SAOR}

Generating realistic inpainted imagery requires selecting appropriate objects for replacement and creating contextually relevant prompts that maintain semantic consistency. To address this challenge, we propose Semantically Aligned Object Replacement (SAOR), a method that automates object selection and prompt generation. Formally, let $I$ be an image and $S(I)$ represent its semantic content. We aim to sample a prompt $p$ and object $o$ under the distribution $\mathcal{P}(p, o \mid S(I))$ that characterizes how humans would select regions and generate prompts given image semantics. We approximate $S(I)$ through two components: the image caption $C(I)$ and the set of displayed objects $O(I)$. This decomposition captures both local semantics (through objects) and global relationships (through captions), providing a descriptive representation of image semantics.

\begin{align}
S(I) \approx \{C(I), O(I)\}
\end{align}

Object masks $M(I) = \{m_1, m_2, \ldots, m_n\}$ and their respective labels $O(I) = \{o_1, o_2, \ldots, o_n\}$ are obtained  from labeled data if available; otherwise, a panoptic segmentation model $\Phi_\text{seg}: I \rightarrow (O(I), M(I))$ is used. Similarly, captions $C(I)$ are taken directly from labeled data or generated by a captioning model $\Psi_\text{cap}$. To approximate $\mathcal{P}$, we leverage LLMs, which have been extensively trained on human text, making them ideal candidates to model human reasoning about visual semantics. Specifically:

\begin{align}
\mathcal{P}(p, o \mid S(I)) \approx \Theta_\text{llm}(p, o \mid O(I), C(I))
\end{align}

\noindent where $\Theta_\text{llm}$ represents the LLM's conditional probability distribution over generated text. The language model, leveraging the caption to understand the image content, is instructed to produce complex, contextually appropriate prompts, maintaining semantic consistency with the surrounding content. The generated prompt $p$, the mask $m$ corresponding to the object selected by the LLM, and the original image $I$ are then provided to an inpainting model $\Gamma$ to generate the inpainted image $\hat{I} = \Gamma(I, m, p)$. The SAOR pipeline is presented at the top of \cref{fig:full_pipeline}



\subsection{Uncertainty-Guided Deceptiveness Assessment}
\label{sec:ugda}

Assessing the realism of inpainted images at test-time is crucial for discarding unrealistic samples and refining the generation process. However, quantifying realism is challenging due to its subjective nature \cite{theis2024makesimagerealistic}. To approximate this subjective judgment, we propose Uncertainty-Guided Deceptiveness Assessment (UGDA), which leverages pretrained VLMs, which have proven successful in synthetic image detection \cite{ye2024lokicomprehensivesyntheticdata}, to compare inpainted images with their originals.

Let $I$ denote the original image and $\hat{I}$ the inpainted version of $I$. Using a VLM $\Omega_\text{vlm}$, we perform a two-stage assessment. Based on empirical observations, unrealistic manipulations are often easily identifiable, while realistic ones require a more thorough evaluation. In the first stage, $\hat{I}$ is assessed for realism. Images passing this check undergo a second, more rigorous evaluation. The second stage leverages this key idea: realistic inpainted images align closer with the distribution of semantically coherent images that a pretrained VLM represents. Therefore, aleatoric uncertainty should increase in a proxy task of ranking the original and its corresponding inpainted image according to their semantic alignment with the learned distribution of the VLM: the bigger the aleatoric uncertainty in this proxy task is, the closer the inpainted image is expected to align with the distribution of semantically correct images. To capture this uncertainty increase, we assume that a confident assessment (i.e.low uncertainty), generated by the VLM, should remain stable under semantically invariant perturbations. If the VLM's assessment varies under such perturbations, it indicates uncertainty about the inpainted image's realism. We argue that a suitable perturbation is input reordering. Specifically, we perform two assessments with reversed image order: 

\vspace{-10pt}
\begin{align}
s_1 = \Omega_\text{vlm}(I, \hat{I}) \quad \text{and} \quad s_2 = \Omega_\text{vlm}(\hat{I}, I),
\end{align}

\noindent where $s_1, s_2 \in \{c_I, c_{\hat{I}}, c_\text{both}\}$ represents the image assessed as more realistic, with $c_I$ indicating the original, $c_{\hat{I}}$ the inpainted image, and $c_\text{both}$ equal realism. Variation in responses or consistently choosing the inpainted image indicates uncertainty. The classification rule: $\hat{I}$ is deceiving if $(s_1 = c_{\hat{I}} \vee s_2 = c_{\hat{I}}) \vee (s_1 = s_2 = c_\text{both})$. Otherwise, it is non-deceiving. This approach enhances reliability in distinguishing between deceiving and easily identifiable synthetic images through uncertainty-aware, order-based evaluations. \cref{fig:deceptive_dataset_examples} shows representative examples from UGDA across different sources. The first row depicts high-quality inpainting examples classified as deceiving by UGDA, where the manipulations are seamlessly integrated with the original content (see suppl. for detailed classification process). The UGDA pipeline is shown in \cref{fig:full_pipeline}

\subsection{\datasetname}
\label{sec:datasource}

Using the proposed pipeline, we introduce \datasetname, the first semantically aligned deceptive dataset for AI-generated inpainting detection, designed to evaluate the effectiveness of our components. The dataset leverages Semantically Aligned Object Replacement (SAOR) for context-aware prompt generation and Uncertainty-Guided Deceptiveness Assessment (UGDA) for realism evaluation, ensuring high-quality and diverse inpainted images.
\vspace{-10pt} 
\paragraph{Sources of authentic images.}
To ensure diversity and robustness, we leverage datasets spanning multiple domains: general object detection, high-resolution photography, and large-scale segmentation. Specifically, we utilize three publicly available datasets for authentic image sources:
(1) MS-COCO~\cite{lin2014coco}, which provides images with captions and object masks across 80 categories,
(2) RAISE~\cite{dang2015raise}, a high-resolution dataset of 8,156 uncompressed RAW images designed for forgery detection evaluation and
(3) OpenImages~\cite{kuznetsova2020open}, which offers extensive object segmentation data with over 2.7 million segmentations across 350 categories.
\vspace{-14pt} 
\paragraph{SAOR configuration.}
We use dataset-provided masks and captions for MS-COCO and OpenImages, while for RAISE, we use OneFormer \cite{jain2022oneformertransformerruleuniversal} as the segmentation model $\Phi_\text{seg}$ and BLIP-2 \cite{li2023blip2bootstrappinglanguageimagepretraining} as the captioning model $\Psi_\text{cap}$. To generate prompts, we employ as the language model $\Theta_\text{llm}$ ChatGPT 3.5 \cite{openai2023chatgpt3.5} for MS-COCO and RAISE as well as Claude Sonnet 3.5 \cite{claude2024sonnet} for OpenImages. The prompt engineering methodology is detailed in the supplementary material.
\vspace{-14pt} 
\paragraph{UGDA configuration.}
We use GPT-4o \cite{openai2023chatgpt4} as the VLM $\Omega_\text{vlm}$, chosen for its effectiveness in synthetic image detection \cite{ye2024lokicomprehensivesyntheticdata} and for achieving the best performance in our respective ablation studies presented in \cref{tab:vlm_ablation}. Empirically, we observed that QAlign effectively filters out low-quality images, so we applied UGDA to approximately half of the test inpainted images, selecting those with the highest QAlign scores. The complete prompt engineering methodology is detailed in the supplementary material.
\vspace{-14pt}
\paragraph{Inpainting Models.}
Diverse inpainting models are crucial for creating robust detection datasets and high-quality images, capturing varied artifacts, styles, and visual traits. We use five pipelines: HD-Painter \cite{manukyan2024hdpainter}, BrushNet \cite{ju2024brushnetplugandplayimageinpainting}, PowerPaint \cite{zhuang2024powerpaint}, ControlNet \cite{zhang2023addingconditionalcontroltexttoimage}, and Inpaint-Anything \cite{yu2023inpaintanythingsegmentmeets}, including its Remove-Anything variant for object removal, with their original padding and normalization strategies maintained. These pipelines support eight models, primarily based on Stable Diffusion \cite{rombach2022stablediffusion}, except Remove-Anything, which uses LaMa \cite{suvorov2021lama}. One-sixth of the images underwent a second inpainting round with different masks and prompts, simulating real-life multi-edit scenarios. Also, due to memory limits, images were resized to a maximum dimension of 2048 pixels, except for Inpaint-Anything, that preserves original dimensions via cropping and resizing. 
\vspace{-14pt} 
\paragraph{Preservation of unmasked area.}
We categorize inpainted images based on how models handle unmasked regions: if the unmasked region is preserved, we refer to them as Spliced (SP) images; if regenerated, as Fully Regenerated (FR) images. Inpaint-Anything preserves unmasked regions through copy-paste (SP), while ControlNet regenerates the full image (FR). BrushNet, PowerPaint, and HD-Painter can produce both SP and FR images depending on post-processing settings (e.g., blending or upscaling). Remove-Anything inherently preserves unmasked regions, thus producing SP images. This diversity in processing approaches contributes to a more comprehensive dataset, as FR images are typically more challenging to detect than SP images \cite{tailanian2024diffusionmodelsmeetimage}.\looseness=-1
\vspace{-14pt} 
\paragraph{Dataset splits.}
Our dataset considers both in-domain performance as well as generalization to new data. Regarding in-domain evaluation, we use MS-COCO (60,000 training and nearly all 5,000 validation images) and RAISE (7,735 images processed with $\Phi_\text{seg}$, yielding 25,674 image-mask-model combinations). To test generalization across different image sources and LLMs, we create an out-of-domain split from OpenImages (6k test images), by also using a different language model $\Theta_\text{llm}$, i.e. Claude versus ChatGPT for COCO and RAISE. Detailed statistics are provided in the supplementary material.

\section{Experimental Evaluation}
\label{sec:experiments}

Our experimental evaluation focuses on two main directions: (1) assessing the quality, aesthetics, and realism of images inpainted with \textit{SAGI}, and (2) evaluating the effectiveness of \datasetname as a benchmark for inpainting detection.


\subsection{Image Quality and Realism Evaluation}

We conduct two ablations to validate our design choices. First, we evaluate SAOR's use of language models by comparing three prompting approaches: (1) using just object labels, (2) combining object labels with image captions, and (3) feeding both labels and captions to an LLM to generate enhanced prompts. Second, we assess UGDA's effectiveness through a human study with 42 participants on 1,000 images, comparing human perception against model performance on images classified as \textit{deceiving} or \textit{non-deceiving}. We also ablate the VLM employed in UGDA, by evaluating the ability of several popular VLMs to align with human perception of image realism.
\vspace{-14pt} 
\paragraph{SAOR evaluation.}

\begin{table}[ht]
\centering
\setlength{\tabcolsep}{0.7mm}
\scalebox{0.9}{
\begin{tabular}{l|ccc|ccccc}
\hline
\multirow{2}{*}{\textbf{Mdl}} & \multicolumn{3}{c|}{\textbf{SAOR}} & \multirow{2}{*}{\textbf{AS}} & \multirow{2}{*}{\textbf{CS Ae}} & \multirow{2}{*}{\textbf{QA Ae}} & \multirow{2}{*}{\textbf{CS Qlt}} & \multirow{2}{*}{\textbf{QA Qlt}} \\
\cline{2-4}
 & \textbf{Obj} & \textbf{Cap} & \textbf{LLM} & & & & & \\
\hline
BN & \checkmark & & & 5.55 & -0.01 & 2.63 & 0.68 & 3.90 \\
\multirow{2}{*}{\cite{ju2024brushnet}} & \checkmark & \checkmark & & 5.69 & 0.14 & 2.68 & 0.69 & 3.94 \\
& \checkmark & \checkmark & \checkmark & \textbf{5.79} & \textbf{0.38} & \textbf{2.75} & \textbf{0.69} & \textbf{3.95} \\
\hline
CN & \checkmark & & & 5.35 & -0.07 & 2.65 & 0.69 & 3.88 \\
\multirow{2}{*}{\cite{zhang2023addingconditionalcontroltexttoimage}} & \checkmark & \checkmark & & 5.40 & 0.01 & 2.68 & 0.70 & 3.92 \\
& \checkmark & \checkmark & \checkmark & \textbf{5.46} & \textbf{0.14} & \textbf{2.71} & \textbf{0.70} & \textbf{3.92} \\
\hline
HDP & \checkmark & & & 5.74 & 0.51 & 2.8 & 0.71 & \textbf{4.12} \\
\multirow{2}{*}{\cite{manukyan2024hdpainter}} & \checkmark & \checkmark & & 5.80 & 0.49 & 2.8 & 0.71 & \textbf{4.12} \\
& \checkmark & \checkmark & \checkmark & \textbf{5.9} & \textbf{0.7} & \textbf{2.83} & \textbf{0.71} & 4.11 \\
\hline
IA & \checkmark & & & 5.49 & -0.12 & 2.44 & 0.66 & 3.76 \\
\multirow{2}{*}{\cite{yu2023inpaintanything}} & \checkmark & \checkmark & & 5.60 & -0.08 & 2.50 & 0.67 & \textbf{3.83} \\
& \checkmark & \checkmark & \checkmark & \textbf{5.68} & \textbf{0.17} & \textbf{2.51} & \textbf{0.67} & 3.81 \\
\hline
PPt & \checkmark & & & 5.68 & 0.46 & 2.75 & \textbf{0.71} & \textbf{4.06} \\
\multirow{2}{*}{\cite{zhuang2024powerpaint}} & \checkmark & \checkmark & & 5.76 & 0.44 & 2.78 & 0.70 & 4.05 \\
& \checkmark & \checkmark & \checkmark & \textbf{5.89} & \textbf{0.61} & \textbf{2.83} & 0.70 & 4.02 \\
\hline
\multirow{3}{*}{Avg} & \checkmark & & & 5.56 & 0.15 & 2.66 & 0.69 & 3.96 \\
 & \checkmark & \checkmark & & 5.65 & 0.20 & 2.69 & 0.69 & \textbf{3.98} \\
& \checkmark & \checkmark & \checkmark & \textbf{5.74} & \textbf{0.40} & \textbf{2.73} & \textbf{0.70} & 3.96 \\
\hline
\end{tabular}
}
\caption{Comparison of metrics across inpainting models with different prompt types. The SAOR columns indicate the employed prompting method: object labels (Obj), image captions (Cap), and LLM-generated prompts (LLM). Bold values indicate better performance per model-metric pair. Pipelines: BN (BrushNet), CN (ControlNet), HDP (HD-Painter), IA (Inpaint-Anything), PPt (PowerPaint). Metrics: AS (Aesthetics Score), CS Ae (Clip Aesthetics), QA Ae (QAlign Aesthetics), CS Qlt (Clip Quality), QA Qlt (QAlign Quality). LLM prompts notably enhance aesthetics, while maintaining performance even in quality metrics like CS Qlt, that are typically less sensitive to semantic variations.}
\label{tab:metrics_llm_vs_obj}
\end{table}

\begin{table}[!htb]
\centering
\setlength{\tabcolsep}{1.35mm}
\begin{tabular}{lcccccc}
\hline
& \multicolumn{3}{c}{\textbf{Accuracy}} & \multicolumn{3}{c}{\textbf{Mean IoU}} \\
\textbf{Metric} & \textbf{Top} & \textbf{Bottom} & \textbf{Diff} & \textbf{Top} & \textbf{Bottom} & \textbf{Diff} \\
\hline
AS & 54.7 & 55.1 &  \textcolor{green!50!black}{+0.4} & 26.6 & 26.3 &  \textcolor{green!50!black}{+0.2}\\
QA Qlt & 56.2 & 53.5 &  \textcolor{green!50!black}{+2.9} & 28.1 & 24.8 &  \textcolor{green!50!black}{+3.3}\\
CS Qlt& 61.2 & 48.5 &  \textcolor{green!50!black}{+12.7} & 30.6 & 22.3 &  \textcolor{green!50!black}{+8.3}\\
CS Ae& 58.9 & 50.9 &  \textcolor{green!50!black}{+8.0} & 28.0 & 24.39 &  \textcolor{green!50!black}{+3.1}\\
QA Ae & 61.3 & 48.4 &  \textcolor{green!50!black}{+12.9} & 30.4 & 22.4 &  \textcolor{green!50!black}{+8.0}\\
\hline
& \textbf{Dec} & \textbf{Non-Dec} & \textbf{Diff} & \textbf{Dec} & \textbf{Non-Dec} & \textbf{Diff} \\
UGDA & 35.2 & 73.7 & \textbf{ \textcolor{green!50!black}{+38.5}} & 12.7 & 39.9 & \textbf{ \textcolor{green!50!black}{+26.8}}\\
\hline
\end{tabular}
\caption{Comparison of human accuracy and mean IoU on 500 inpainted images, separated into two groups for each metric. For AS, QA Qlt, CS, and QA Ae, groups are formed by ranking images and splitting them into the top 50\% and bottom 50\% based on their metric scores; for UGDA, groups are based on its assessment (deceiving/non-deceiving). UGDA can clearly separate deceiving from non-deceiving images.}
\label{tab:ugda_vs_metrics}
\end{table}

\begin{table}[!htb]
\centering
\begin{tabular}{lcc}
\specialrule{\heavyrulewidth}{0.4pt}{0.4pt}  
VLM / Human Eval. & Acc $\downarrow$ & IoU $\downarrow$ \\
\specialrule{\lightrulewidth}{0.5pt}{0.5pt}  
\specialrule{\lightrulewidth}{0.5pt}{0.5pt} 
w/o UGDA & $54.9$ & $26.4$ \\
\specialrule{\lightrulewidth}{0.5pt}{0.5pt}  
Gemma-3-27b \cite{gemmateam2025gemma3technicalreport} & $45.4^{\textcolor{green!50!black}{+9.5}}$ & $20.0^{\textcolor{green!50!black}{+6.4}}$ \\
Mistral-S-3.1-24B \cite{mistral2025small31}& $45.2^{\textcolor{green!50!black}{+9.7}}$ & $20.4^{\textcolor{green!50!black}{+6.0}}$ \\
Qwen2.5-VL-7B \cite{bai2025qwen25vltechnicalreport}& $45.2^{\textcolor{green!50!black}{+9.7}}$ & $21.0^{\textcolor{green!50!black}{+5.4}}$ \\
Claude-3-7-sonnet \cite{anthropic2025claude37} & $43.4^{\textcolor{green!50!black}{+11.5}}$ & $17.8^{\textcolor{green!50!black}{+8.6}}$ \\
Gemini-2.5-flash \cite{comanici2025gemini25pushingfrontier}& $38.6^{\textcolor{green!50!black}{+16.3}}$ & $16.5^{\textcolor{green!50!black}{+9.9}}$ \\
GPT-4o \cite{openai2023chatgpt4} & $\textbf{35.2}^{\textbf{\textcolor{green!50!black}{+19.7}}}$ & $\textbf{12.7}^{\textbf{\textcolor{green!50!black}{+13.7}}}$ \\
\specialrule{\heavyrulewidth}{0.4pt}{0.4pt}  
\end{tabular}
\caption{VLM ablation for UGDA. Human accuracy and mean IoU for distinguishing inpainted from real images after filtering with different VLMs. Lower values indicate more deceiving inpaintings. All VLMs improve over the unfiltered baseline, with GPT-4o achieving the strongest alignment with human perception.}
\label{tab:vlm_ablation}
\vspace{-8pt}
\end{table}

\begin{table*}[!htb]
\centering
\setlength{\tabcolsep}{0.5mm}
\scalebox{0.97}{
\begin{tabular}{lcccccccccccccc}
\hline
\multirow{2}{*}{Model} & Orig. & Orig. & Inp. & Inp. & Inp. & Double & \multirow{2}{*}{Type} & \multirow{2}{*}{Resolution} & \multirow{2}{*}{QA Qlt $\uparrow$} & \multirow{2}{*}{QA Ae $\uparrow$} & \multirow{2}{*}{CS Qlt $\uparrow$} & \multirow{2}{*}{CS Ae $\uparrow$} & \multirow{2}{*}{AS $\uparrow$}\\
 & Datasets & Imgs & Imgs & Mdls & Pipes & Inp. &  &  \\
\hline
CocoGlide\cite{guillaro2023trufor} & 1 & 512 & 512 & 1 & \ding{55} & \ding{55} & AI & $256\times256$ & 2.88 & 1.79 & \textbf{0.68} & -1.27 & 5.40\\
TGIF\cite{mareen2024tgif} & 1 & 3,124 & 74,976 & 3 & \ding{55} & \ding{55} & AI & up to $1024p$ & 3.94 & 2.53 & 0.63 & -0.69 & 5.65\\
\datasetname (ours) & 3 & 77,900 & 95,839 & 8 & 5 & \checkmark & AI/OR & up to $2048p$* & \textbf{4.06} & \textbf{2.84} & \textbf{0.68} & \textbf{0.27} & \textbf{5.69}\\
\hline
\end{tabular}
}
\vspace{-4pt}
\caption{Comparison of inpainting datasets characteristics. Our dataset surpasses existing ones in scale (number of images), diversity (source datasets, models, pipelines). Resolution varies based on source dataset. AI: AI-Generated Content, OR: Object Removal.}
\label{tab:cocoglide_vs_tgif_vs_ours}
\vspace{-8pt}
\end{table*} 

To compare LLM-generated prompts with object labels or labels and captions, we evaluated 900 original images across five inpainting models, generating 4,500 images per prompt type. We compared aesthetics and quality metrics, including CLIP Similarity \cite{radford2021clip} for aesthetics (CS Ae) \cite{hentschel2022clipimageaeshetics}  and quality (CS Qlt) \cite{wang2022clipiqa} , QAlign \cite{wu2023qalignteachinglmmsvisual} for quality (QA Qlt) and aesthetics (QA Ae), and Aesthetic Score (AS) \cite{schuhmann2022laion5bopenlargescaledataset}. \cref{tab:metrics_llm_vs_obj} shows that LLM prompts consistently outperform object labels and caption prompts across most metrics and models, enhancing both image quality and aesthetics. All models achieve higher aesthetic metrics with LLM prompts. BrushNet and ControlNet excelled across all metrics, while other models showed slight variations in quality metrics. Quality improvements were marginal compared to aesthetic gains, as quality metrics primarily focus on technical aspects of the image which are less influenced by prompt's content. Overall, these results highlight the richer semantic guidance of LLM prompts, significantly enhancing inpainting aesthetics, while also improving some technical aspects of the generated visual content.

\vspace{-14pt} 
\paragraph{UGDA  evaluation.}

We validated UGDA through a user study with 42 participants on 1,000 images: 250 inpainted ones classified by UGDA as deceiving, 250 as non-deceiving and 500 authentic. Images were selected to avoid redundancy, i.e. no authentic-inpainted overlap and no multiple inpainted versions per authentic source image. Participants evaluated batches of 20 images, with each image receiving 3-5 independent assessments. Detailed participant statistics are provided in the supplementary material.

\cref{tab:ugda_vs_metrics} compares human accuracy and mean IoU on 500 inpainted images, split into two groups per metric. For AS, QA Qlt, CS, and QA Ae, splits are based on top/bottom 50\% rankings; for UGDA, on its deceiving/non-deceiving assessment. Existing quality and aesthetic metrics (AS, QA Qlt, CS, QA Ae) show minimal differences among groups (accuracy: 0.4–12.9; IoU: 0.2–8.3), indicating limited correlation with human perception of realism. In contrast, UGDA yields much larger gaps (accuracy: 38.5; IoU: 26.8), effectively identifying realistic inpaintings. This underscores the need for uncertainty guidance to capture realism.

To further validate the efficacy and model-agnostic nature of UGDA, we compare the performance of several VLMs. \cref{tab:vlm_ablation} compares human ability to distinguish inpainted from real images, both in terms of accuracy and mean IoU, after non-deceiving samples have been filtered out using our methodology. Results demonstrate that our method enables all tested VLMs to better align with human preference and discard non-realistic inpaintings, as evidenced by the positive differences over a baseline without UGDA-based filtering. As GPT-4o shows the strongest alignment with human perception, we employ it as the primary model in our benchmark creation.

\vspace{-14pt} 
\paragraph{Quantitative comparison with state-of-the-art.} 

To the best of our knowledge, \datasetname is the largest collection of AI-generated inpainted images. \cref{tab:cocoglide_vs_tgif_vs_ours} compares \datasetname with existing datasets. With 95,839 inpainted images from 77,900 originals, it surpasses TGIF (74,976 from 3,124) and CocoGlide (512 from 512) in scale. Unlike single-source datasets, \datasetname integrates COCO, RAISE, and OpenImages, enhancing diversity. It employs eight inpainting models across five pipelines, considers advanced manipulations such as double inpainting, AI generated content, and object removal, and allows resolutions up to 2048p—exceeding TGIF's 1024p and CocoGlide's 256p. \datasetname uniquely includes human benchmark and out-of-domain test subsets, making it valuable for evaluating detection under diverse conditions. It outperforms existing datasets in aesthetic and quality metrics, demonstrating superior perceptual alignment and visual fidelity. This positions \datasetname as a superior resource for advancing detection models and sets a new standard for benchmarking in the field.

\begin{figure}[!htb]
    \centering
    \begin{subfigure}{0.49\columnwidth}
        \includegraphics[width=\linewidth]{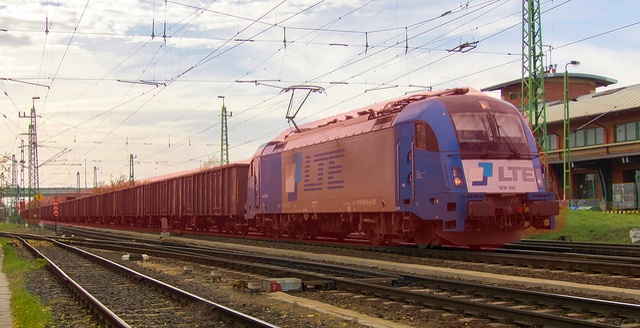}
    \end{subfigure}
    \hfill
    \begin{subfigure}{0.49\columnwidth}
        \includegraphics[width=\linewidth]{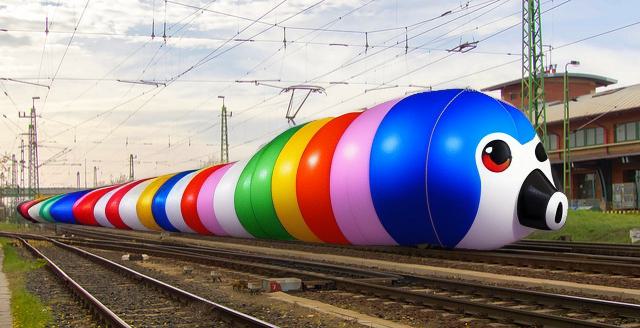}
    \end{subfigure}
    \\[0.3em]
    \begin{subfigure}{0.49\columnwidth}
        \includegraphics[width=\linewidth]{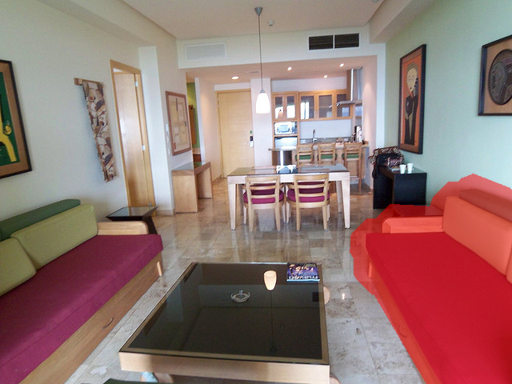}
    \end{subfigure}
    \hfill
    \begin{subfigure}{0.49\columnwidth}
        \includegraphics[width=\linewidth]{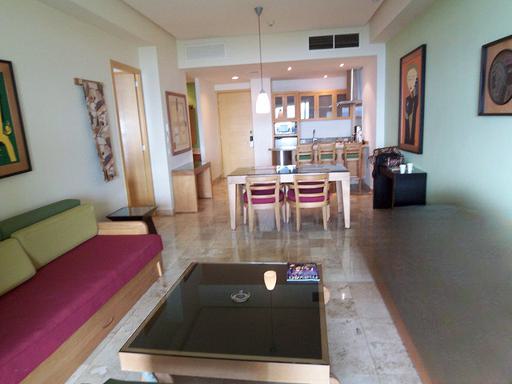}
    \end{subfigure}
    
    \caption{Failure cases for SAOR and UGDA. Top row shows SAOR generating semantically inappropriate content (a colorful hot air balloon floating in the sky). Bottom row demonstrates UGDA's failure: 1st assessment gets fixated on ``unnatural duplication of couch'' without recognizing the blur, while 2nd assessment correctly identifies ``couch appears altered with blurry texture''.}
    \label{fig:failure_cases_main}
\end{figure}

\paragraph{Failure Cases}
While our framework demonstrates strong overall performance, certain limitations remain. A representative failure case for SAOR occurs when, despite the context provided, the LLM generates semantically incoherent prompts. For example, in the first row of \cref{fig:failure_cases_main}, a hot air balloon is suggested as a replacement for a train object. For UGDA, the VLM can exhibit inconsistent behavior by fixating on specific details while ignoring obvious artifacts. As shown in the second row of \cref{fig:failure_cases_main}, in the first permutation, the VLM insists on ``couch duplication'' and fails to recognize clear blurring artifacts, whereas in the second permutation, it correctly identifies the same blur. These cases highlight the dependency of our approach on the underlying capabilities and biases of foundation models. More examples are discussed in the supplementary material.

\subsection{Inpainting Detection Benchmark}

To establish a comprehensive benchmark for the presented \datasetname dataset, we evaluate the performance of several state-of-the-art image forgery detection models. 
\vspace{-10pt} 
\paragraph{Problem definition.} Given an RGB image \( x^{rgb} \in \mathbb{R}^{(H \times W \times 3)} \), the inpainting detection model aims to predict a pixel-level inpainting localization mask \( \hat{y}^{loc} \in (0,1)^{(H \times W \times 1)} \) and/or an image-level inpainting detection probability \( \hat{y}^{det} \in (0,1) \). The former will be referred to as the localization task, and the latter as the detection task.
\vspace{-14pt} 
\paragraph{Forensics models.} We evaluated four inpainting detectors: PSCC-Net \cite{luy2022pscc}, CAT-Net \cite{kwon2022catnet}, TruFor \cite{guillaro2023trufor}, and MMFusion \cite{triaridis2023mmfusion}. CAT-Net provides only pixel-level masks, so we used the maximum probability from these masks for image-level detection. The other models output both pixel localization and image-level detection probabilities.
\vspace{-14pt} 
\paragraph{Training protocol.} We evaluated the pretrained models, the versions retrained on the \datasetname and on TGIF \cite{mareen2024tgif} following the training protocol on the original papers. 
\vspace{-14pt} 
\paragraph{Implementation Details.} We retrained all models from scratch, following the training protocol outlined in their original papers. CAT-Net was trained on an NVIDIA A100 GPU, while PSCC-Net, TruFor, and MMFusion were trained on an NVIDIA RTX 4090.
\vspace{-14pt} 
\paragraph{Evaluation metrics.} We evaluated model performance at image and pixel levels using distinct metrics. With the positive class representing inpainted regions, for image-level detection, we use accuracy, and for pixel level we measure Intersection over Union (IoU), with a threshold of $0.5$. To compare models across datasets, we used the threshold-agnostic Area Under the Curve (AUC) metric. For localization AUC, we resized and flattened localization maps and ground truths into vectors for ROC computation (see supplementary for additional results, qualitative comparisons, and human performance analysis).


\subsection{Localization and Detection Results}

\begin{table*}[!htb]
\centering
\setlength{\tabcolsep}{1mm}
\scalebox{1}{
    \begin{tabular}{clcccccccccccccccc}
\hline
\multirow{2}{*}{Data} & \multirow{2}{*}{Model} & \multicolumn{8}{c}{Mean IoU} & \multicolumn{8}{c}{Accuracy} \\
 & & ID &  & OOD &  & SP &  & FR &  & ID &  & OOD &  & SP &  & FR &  \\
\cmidrule(lr){1-2} \cmidrule(lr){3-10} \cmidrule(lr){11-18}
\multirow{4}{*}{\rotatebox[origin=c]{90}{\textbf{Original}}} & CAT-Net\cite{kwon2022catnet} & 39.3 &  & 20.9 &  & 41.4 &  & 5.9 &  & 59.8 &  & 42.4 &  & 84.5 &  & 39.9 &  \\
 & PSCC-Net\cite{luy2022pscc} & 35.1 &  & 24.0 &  & 20.1 &  & 9.2 &  & 59.9 &  & 53.0 &  & 40.4 &  & 33.4 &  \\
 & MMFusion\cite{triaridis2023mmfusion} & 19.6 &  & 19.1 &  & 46.3 &  & 16.9 &  & 61.7 &  & 63.4 &  & 62.9 &  & 25.5 &  \\
 & TruFor\cite{guillaro2023trufor} & 12.2 &  & 19.1 &  & 40.0 &  & 19.6 &  & 58.2 &  & 60.1 &  & 41.0 &  & 12.5 &  \\
\cmidrule(lr){1-2} \cmidrule(lr){3-10} \cmidrule(lr){11-18}
\multirow{4}{*}{\rotatebox[origin=c]{90}{\textbf{SAGI-D}}} & CAT-Net\cite{kwon2022catnet} & 69.2 & {\footnotesize \textcolor{green!50!black}{+29.9}} & 23.0 & {\footnotesize \textcolor{green!50!black}{+2.1}} & 46.5 & {\footnotesize \textcolor{green!50!black}{+5.1}} & 36.2 & {\footnotesize \textcolor{green!50!black}{+30.3}} & 98.8 & {\footnotesize \textcolor{green!50!black}{+39.0}} & 53.0 & {\footnotesize \textcolor{green!50!black}{+10.6}} & 99.9 & {\footnotesize \textcolor{green!50!black}{+15.4}} & 99.9 & {\footnotesize \textcolor{green!50!black}{+60.0}} \\
 & PSCC-Net\cite{luy2022pscc} & 58.1 & {\footnotesize \textcolor{green!50!black}{+23.0}} & 57.6 & {\footnotesize \textcolor{green!50!black}{+33.6}} & 43.2 & {\footnotesize \textcolor{green!50!black}{+23.1}} & 19.6 & {\footnotesize \textcolor{green!50!black}{+10.4}} & 63.4 & {\footnotesize \textcolor{green!50!black}{+3.5}} & 71.8 & {\footnotesize \textcolor{green!50!black}{+18.8}} & 50.1 & {\footnotesize \textcolor{green!50!black}{+9.8}} & 35.6 & {\footnotesize \textcolor{green!50!black}{+2.2}} \\
 & MMFusion\cite{triaridis2023mmfusion} & 63.7 & {\footnotesize \textcolor{green!50!black}{+44.1}} & 41.4 & {\footnotesize \textcolor{green!50!black}{+22.3}} & 73.1 & {\footnotesize \textcolor{green!50!black}{+26.8}} & 50.3 & {\footnotesize \textcolor{green!50!black}{+33.4}} & 92.8 & {\footnotesize \textcolor{green!50!black}{+31.1}} & 80.9 & {\footnotesize \textcolor{green!50!black}{+17.5}} & 85.9 & {\footnotesize \textcolor{green!50!black}{+23.0}} & 81.5 & {\footnotesize \textcolor{green!50!black}{+56.0}} \\
 & TruFor\cite{guillaro2023trufor} & 64.7 & {\footnotesize \textcolor{green!50!black}{+52.5}} & 65.5 & {\footnotesize \textcolor{green!50!black}{+46.4}} & 86.8 & {\footnotesize \textcolor{green!50!black}{+46.8}} & 72.4 & {\footnotesize \textcolor{green!50!black}{+52.8}} & 96.2 & {\footnotesize \textcolor{green!50!black}{+38.0}} & 92.8 & {\footnotesize \textcolor{green!50!black}{+32.7}} & 95.5 & {\footnotesize \textcolor{green!50!black}{+54.5}} & 94.4 & {\footnotesize \textcolor{green!50!black}{+81.9}} \\
\hline
\end{tabular}
}
\vspace{-2pt}
\caption{Evaluation of SAGI on training image forgery detection and localization models. The first column indicates training data, while Mean IoU is reported for localization and Accuracy for detection. Results include in-domain (ID) and out-of-domain (OOD) performance for spliced (SP) and fully-regenerated (FR) images. ID and OOD performance is computed across inpainted and authentic images, while SP and FR highlight the performance on the corresponding type of inpaintings. \textcolor{green!50!black}{Green} numbers show improvements over original models. Retraining on \datasetname significantly enhances performance, underscoring the value of semantic alignment for high-quality synthetic data.}
\label{tab:performance_comparison}
\vspace{-2pt}
\end{table*}

\begin{table}[!htb]
\centering
\setlength{\tabcolsep}{0.6mm}
\scalebox{0.86}{
\begin{tabular}{clcccccccc}
\hline
\multirow{2}{*}{Data} & \multirow{2}{*}{Model} & \multicolumn{4}{c}{AUC (det)} & \multicolumn{4}{c}{AUC (loc)} \\
 & & TGIF & & \emph{\datasetname} & & TGIF & & \emph{\datasetname} & \\
\cmidrule(lr){1-2} \cmidrule(lr){3-6} \cmidrule(lr){7-10}
\multirow{4}{*}{\rotatebox[origin=c]{90}{\textbf{Original}}} & CN \cite{kwon2022catnet} & 75.1 & & 53.2 & & 74.9 & & 56.3 & \\
 & MM \cite{triaridis2023mmfusion} & 86.5 & & 75.3 & & 67.4 & & 67.3 & \\
 & PS \cite{luy2022pscc} & 73.8 & & 62.7 & & 39.0 & & 66.0 & \\
 & TF \cite{guillaro2023trufor} & 89.4 & & 79.5 & & 70.5 & & 67.0 & \\
 & \textbf{Average} & \textbf{81.2} & & \textbf{67.7} & & \textbf{63.0} & & \textbf{64.2} & \\
\cmidrule(lr){1-2} \cmidrule(lr){3-6} \cmidrule(lr){7-10}
\multirow{4}{*}{\rotatebox[origin=c]{90}{\textbf{TGIF}}} & CN\cite{kwon2022catnet} & 69.5 & {\scriptsize \textcolor{red}{-5.6}} & 68.1 & {\scriptsize \textcolor{green!50!black}{+14.9}} & 81.0 & {\scriptsize \textcolor{green!50!black}{+6.1}} & 62.6 & {\scriptsize \textcolor{green!50!black}{+6.3}} \\
 & MM \cite{triaridis2023mmfusion} & 95.9 & {\scriptsize \textcolor{green!50!black}{+9.4}} & 82.2 & {\scriptsize \textcolor{green!50!black}{+6.9}} & 93.7 & {\scriptsize \textcolor{green!50!black}{+26.3}} & 82.1 & {\scriptsize \textcolor{green!50!black}{+14.8}} \\
 & PS \cite{luy2022pscc} & 73.5 & {\scriptsize \textcolor{red}{-0.2}} & 69.6 & {\scriptsize \textcolor{green!50!black}{+7.0}} & 60.9 & {\scriptsize \textcolor{green!50!black}{+21.9}} & 49.6 & {\scriptsize \textcolor{red}{-16.4}} \\
 & TF \cite{guillaro2023trufor} & 58.4 & {\scriptsize \textcolor{red}{-31.0}} & 74.4 & {\scriptsize \textcolor{red}{-5.1}} & 95.0 & {\scriptsize \textcolor{green!50!black}{+24.5}} & 89.7 & {\scriptsize \textcolor{green!50!black}{+22.7}} \\
 & \textbf{Average} & \textbf{74.3} & {\scriptsize \textcolor{red}{-6.9}} & \textbf{73.6} & {\scriptsize \textcolor{green!50!black}{+5.9}} & \textbf{82.7} & {\scriptsize \textcolor{green!50!black}{+19.7}} & \textbf{71.0} & {\scriptsize \textcolor{green!50!black}{+6.8}} \\
\cmidrule(lr){1-2} \cmidrule(lr){3-6} \cmidrule(lr){7-10}
\multirow{4}{*}{\rotatebox[origin=c]{90}{\textbf{SAGI-D}}} & CN \cite{kwon2022catnet} & 84.6 & {\scriptsize \textcolor{green!50!black}{+9.5}} & 92.7 & {\scriptsize \textcolor{green!50!black}{+39.4}} & 73.2 & {\scriptsize \textcolor{red}{-1.7}} & 84.9 & {\scriptsize \textcolor{green!50!black}{+28.6}} \\
 & MM \cite{triaridis2023mmfusion} & 71.9 & {\scriptsize \textcolor{red}{-14.6}} & 97.1 & {\scriptsize \textcolor{green!50!black}{+21.8}} & 54.6 & {\scriptsize \textcolor{red}{-12.8}} & 95.9 & {\scriptsize \textcolor{green!50!black}{+28.6}} \\
 & PS \cite{luy2022pscc} & 65.5 & {\scriptsize \textcolor{red}{-8.3}} & 82.9 & {\scriptsize \textcolor{green!50!black}{+20.2}} & 83.5 & {\scriptsize \textcolor{green!50!black}{+44.6}} & 75.9 & {\scriptsize \textcolor{green!50!black}{+9.9}} \\
 & TF \cite{guillaro2023trufor} & 91.0 & {\scriptsize \textcolor{green!50!black}{+1.6}} & 99.5 & {\scriptsize \textcolor{green!50!black}{+20.0}} & 91.0 & {\scriptsize \textcolor{green!50!black}{+20.5}} & 98.8 & {\scriptsize \textcolor{green!50!black}{+31.8}} \\
 & \textbf{Average} & \textbf{78.3} & {\scriptsize \textcolor{red}{-2.9}} & \textbf{93.1} & {\scriptsize \textcolor{green!50!black}{+25.4}} & \textbf{75.6} & {\scriptsize \textcolor{green!50!black}{+12.6}} & \textbf{88.9} & {\scriptsize \textcolor{green!50!black}{+24.7}} \\
\hline
\end{tabular}
}
\caption{Performance comparison of image forensics methods CAT-Net (CN), MMFusion (MM), PSCC-Net (PS), and TruFor (TF) across TGIF and \datasetname datasets. Metrics include detection and localization AUC. \emph{Data} indicates training source. \textcolor{green!50!black}{Green} and \textcolor{red}{red} numbers show performance improvements and decreases over original models. Retraining on \datasetname yields more consistent improvements, even across datasets, compared to training on TGIF.}
\label{tab:combined_metrics}
\vspace{-10pt}
\end{table}

We evaluate \datasetname using four state-of-the-art inpainting detection models: PSCC-Net, CAT-Net, TruFor, and MMFusion. We assess both pre-trained models and versions retrained on our dataset. Table \ref{tab:performance_comparison} presents both the localization (IoU) and detection (accuracy) results for in-domain and out-of-domain testing sets and SP and FR images. Retraining on \datasetname leads to significant performance improvements across all models. TruFor shows the largest gains, with IoU increasing by +52.5\% (in-domain) and +46.4\% (out-of-domain), while CAT-Net achieves the highest in-domain IoU (69.2\%, +29.9\% improvement). FR regions remain challenging for original models (IoUs between 5.9\% and 19.6\%), but retrained models show substantial improvements, with TruFor reaching 72.4\% IoU (+52.8\%). SP detection is easier, with retrained TruFor achieving 86.8\% IoU. TruFor and PSCC-Net maintain consistent cross-domain performance, while CAT-Net and MMFusion exhibit variability due to their designs (e.g., JPEG-specific artifacts and added input complexities).
Overall, \datasetname proves highly effective for improving image forgery detection and localization models.

To further assess \datasetname as a benchmark, we compare models trained on their original data, TGIF and SAGI-D, and evaluate them on both TGIF and \datasetname test sets. \cref{tab:combined_metrics} shows localization and detection AUC results. Training on \datasetname consistently improves performance, outperforming TGIF-trained models. Notably, SAGI-D-trained models sometimes surpass TGIF-trained ones even on TGIF (78.3 vs 74.3 detection AUC average). CAT-Net trained on \datasetname achieves +28.6 and +21.8 improvements on \datasetname and TGIF respectively, while TGIF-trained CAT-Net drops -5.6 on TGIF. PSCC-Net shows +44.6 and +20.2 localization AUC gains, and SAGI-D-trained TruFor reaches AUCs above 90 on both datasets. Original models perform better on TGIF than SAGI-D, highlighting its greater complexity and benchmark value. These results affirm the superiority of \datasetname for training detection models.


\section{Conclusions}
\label{sec:conclusions}
In this work we proposed a model-agnostic framework for generating and evaluating high-quality inpaintings, based on the key ideas of semantic alignment and uncertainty guidance. Looking ahead, our framework can benefit from advances in foundation models to further refine prompt generation and realism assessment and extend uncertainty mechanisms to enhance LLM/VLM performance in tasks involving reference samples, such as image-to-image translation, audio/video editing, and quality assessment of generated content. We aim for this framework to advance both generative quality and forensic detection, supporting safer applications of generative AI.

\vspace{-10pt}
\paragraph{Limitations:} The limited capacity and finite training data of any pre-trained LLM and VLM impose inherent constraints on tasks like semantic alignment and realism assessment. Despite these limitations, we consider them the best approximations available for bridging the gap between the real world and computer-generated visual representations.
\vspace{-10pt}
 \paragraph{Ethical Concerns:} While advances in generative AI provide significant societal benefits by automating previously labor-intensive tasks, they also enable potential malicious exploitation, such as spreading misinformation, or manipulating visual evidence. Yet, the fact that our proposed framework can significantly enhance the robustness of image forensic tools, allows mitigating such misuse risks.
 \vspace{-8pt}

\paragraph{Acknowledgments:} This work was supported by the Horizon Europe projects vera.ai (grant no. 101070093), ELLIOT (101214398) and AI4TRUST (101070190). Compute resources were granted by GRNET and the HPC infrastructure of the Aristotle University of Thessaloniki.
\clearpage
{
    \small
    \balance
    \bibliographystyle{ieeenat_fullname}
    \bibliography{main}
}
\clearpage
\maketitlesupplementary

\section{Implementation details}
This section provides additional implementation details of our approach to ensure reproducibility. Code is available on \url{https://github.com/mever-team/SAGI}.

\subsection{Source of Authentic Images}
Since RAISE \cite{dang2015raise} contains RAW images, we processed these images before using them for inpainting experiments. We utilized the RAISE dataset as described in \cite{kwon2022catnet}.

\subsection{Dataset Splits}

As shown in Table \ref{tab:dataset_splits}, we structure our dataset to evaluate both in-domain performance and generalization to new data. For in-domain evaluation, we use COCO (60,000 randomly selected training images and nearly all 5,000 validation images for validation and testing) and RAISE (7,735 images processed with $\Phi_\text{seg}$, yielding 25,674 image-mask-model combinations through 1-7 masks or prompts per image, with derived images kept in the same split, as each image was inpainted up to 4 times only in this dataset). To test generalization, we create an out-of-domain testing split using OpenImages \cite{benenson2022openimages}—a dataset not used during training—comprising 6,000 randomly selected test images. This split uses a different language model $\Theta_\text{llm}$ (Claude) than COCO and RAISE (ChatGPT), providing a way to evaluate how well models perform on both new data and different prompting approaches. Throughout our experiments, we refer to the COCO and RAISE test splits as in-domain and the OpenImages test split as out-of-domain.

\begin{table}[!htbp]
\centering
\begin{tabular}{lccc}
\hline
 & Training & Validation & Testing \\
\hline
\multirow{2}{*}{COCO \cite{lin2014coco}} & 59,708 & 1,950 & 2,922 \\
 & (75\%) & (31\%) & (29\%) \\
\multirow{2}{*}{RAISE \cite{dang2015raise}} & 19,741 & 4,262 & 1,671 \\
 & (25\%) & (69\%) & (16\%) \\
\multirow{2}{*}{OpenImages \cite{benenson2022openimages}} & \multirow{2}{*}{\textit{N/A}} & \multirow{2}{*}{\textit{N/A}} & 5,585 \\
 & & & (55\%) \\
 \hline
Inpainted & 79,449 & 6,212 & 10,178 \\
Authentic & 79,449 & 6,212 & 9,071 \\
\hline
\end{tabular}
\caption{Overview of dataset splits across COCO, RAISE, and OpenImages. The table shows the number of images in each split. The total number of images, including authentic and inpainted versions, is provided. Percentages represent the distribution of each dataset within the total split for inpainted images.}
\label{tab:dataset_splits}
\end{table}

\subsection{SAOR configuration}
The specific API endpoints used in our implementation were gpt-3.5-turbo \cite{openai2023chatgpt3.5} (as of June 2024) and claude-3-5-sonnet-20240620 \cite{claude2024sonnet}. The system prompt used for the LLMs in SAOR was configured as shown in \cref{fig:llm-prompts}. For double inpainting cases, where two objects needed to be sequentially modified, we used an adapted system prompt to select a second object and generate a prompt as shown in \cref{fig:llm-prompts}. For images designated for object removal, we used a simplified system prompt focused solely on object selection that is shown in \cref{fig:llm-prompts}.

\begin{figure*}[ht]
\centering
\small 
\begin{tabular}{|p{0.9\textwidth}|}
\hline
\textbf{LLM System Prompt (1st Inpainting)} \\
\hline
You write prompts for text-to-image image inpainting models (AI-inpainting). In these models, you give an image, a mask of an area that will be inpainted, and a text prompt to tell the model what to inpaint the masked area with. You will be given a caption of the original image (the whole image) to understand the context and a list of objects. Then you choose an object, \textbf{THAT EXISTS IN THE LIST GIVEN TO YOU}. You need to generate a suitable prompt to alter the masked area of the image that covers the object you chose. 

Remember to make a prompt that alters the image. If you decide to replace the said object, replace it with something that makes sense given the object that is to be replaced and the caption. Also, do not mention the original object in the prompt unless you want to replace the said object with one of the same class. Generate the prompt like this:

\textbf{Object:} \{object on the original list\}

\textbf{Prompt:} Inpaint the masked area with... \\
\hline
\end{tabular}

\vspace{0.5cm} 

\begin{tabular}{|p{0.9\textwidth}|}
\hline
\textbf{LLM System Prompt (2nd Inpainting)} \\
\hline
You write prompts for text-to-image image inpainting models (AI-inpainting). In these models, you give an image, a mask of an area that will be inpainted, and a text prompt to tell the model what to inpaint the masked area with. An object has already been replaced in the image, and we need to generate a \textbf{DIFFERENT} prompt for a second object. 

You will be given a caption of the image to understand the context, the class of the 1st object, and the prompt of the 1st object. You will then \textbf{SELECT A 2ND OBJECT} from the image that is to be inpainted. You need to generate a suitable prompt to alter the masked area of the image that covers the 2nd object.

Remember to make a prompt that alters the image. If you decide to replace the said object, replace it with something that makes sense given the object that is to be replaced and the caption. Also, do not mention the original object in the prompt unless you want to replace the said object with one of the same class. Generate the prompt like this:

\textbf{Object:} \{name of the 2nd object\}

\textbf{Prompt:} Inpaint the masked area with... \\
\hline
\end{tabular}

\vspace{0.5cm} 

\begin{tabular}{|p{0.9\textwidth}|}
\hline
\textbf{LLM System Prompt (Removal)} \\
\hline
You will be given a list of objects that exist in an image. You must choose an object to be removed with inpainting methods. Choose an object that makes sense.\\
Answer like this:\\
\textbf{Object:} \{object in the list\} \\
\hline
\end{tabular}

\caption{System Prompts for selecting objects and generating prompts for inpainting and removal. The first prompt is for the 1st inpainting, the second for the 2nd inpainting, and the third for object removal.}
\label{fig:llm-prompts}
\end{figure*}

All LLM interactions were configured with hyperparameters including a temperature of 1.2 to encourage creative variations in the generated prompts, a top-p (nucleus sampling threshold) of 0.8, and a maximum token limit of 40 for prompt length. Some prompts from initial experiments, conducted without the maximum token restriction, were retained in our final dataset. The prefix ``Inpaint the masked area with...'' was included in the system prompts to maintain a consistent format in the LLMs' responses but was omitted from the actual saved prompts to avoid potential misinterpretation by diffusion models.

\subsection{Inpainting Pipelines Configuration}
The text-guided inpainting models support Stable Diffusion \cite{rombach2022stablediffusion} by default, along with certain community versions. Specifically, HD-Painter \cite{manukyan2024hdpainter} supports Stable Diffusion v1.5\footnote{\url{https://huggingface.co/stable-diffusion-v1-5/stable-diffusion-v1-5}}, Stable Diffusion v2\footnote{\url{https://huggingface.co/stabilityai/stable-diffusion-2}}, and DreamShaper v8\footnote{\url{https://civitai.com/models/4384/dreamshaper}}. BrushNet \cite{ju2024brushnet} supports Stable Diffusion v1.5, Stable Diffusion XL\footnote{\url{https://huggingface.co/stabilityai/stable-diffusion-xl-base-1.0}}, DreamShaper v8, Realistic Vision\footnote{\url{https://civitai.com/models/4201/realistic-vision-v60-b1}}, epiCRealism\footnote{\url{https://civitai.com/models/25694/epicrealism}}, and Juggernaut\-XL\footnote{\url{https://civitai.com/models/133005/juggernaut-xl}}. PowerPaint \cite{zhuang2024powerpaint} combines Realistic Vision and BrushNet, while ControlNetInpaint \cite{zhang2023addingconditionalcontroltexttoimage} supports Stable Diffusion v1.5. Inpaint-Anything \cite{yu2023inpaintanything} supports Stable Diffusion v2.

Each inpainting pipeline received an equal number of images for processing, with Remove-Anything being treated as a separate pipeline, and the settings for each pipeline, such as diffusion models and post-processing techniques, were distributed uniformly. Despite efforts to maintain uniformity, small discrepancies occurred due to constraints such as excluding NSFW images flagged by Stable Diffusion.

\subsection{UGDA Configuration}
The Uncertainty-Guided Deceptiveness Assessment (UGDA) was implemented using the chatgpt-4o-latest \cite{openai2023chatgpt4} API endpoint (as of October 2024). We conducted preliminary experiments with various prompt formulations and found that allowing the model to explain its reasoning before providing a verdict led to more reliable assessments. The final implementation uses a two-stage process with carefully engineered prompts. The initial realism check uses the system prompt shown in \cref{fig:vlm-prompts}. For images passing the initial check, a comparative assessment is performed using the system prompt shown in \cref{fig:vlm-prompts}.

\begin{figure*}[ht]
\centering
\small 
\begin{tabular}{|p{0.9\textwidth}|}
\hline
\textbf{VLM System Prompt (First Realism Check)} \\
\hline
You are an expert in analyzing images for signs of modification by AI models and assessing their realism. Your task is to examine an image and determine if it is realistic. In addition, provide a brief explanation highlighting the key elements that informed your judgment, such as unusual textures, unnatural elements, unusual biological phenomena (animals with multiple limbs, heads), unusual arrangement of objects (weirdly clustered objects), or any visible artifacts.

\textbf{\# Output Format} \\
\textbf{Assessment:} (your explanation goes here) \\
\textbf{Verdict:} (final verdict, answer ``Yes, it is realistic'' or ``No, it is not realistic'') \\
\hline
\end{tabular}

\vspace{0.5cm} 

\begin{tabular}{|p{0.9\textwidth}|}
\hline
\textbf{VLM System Prompt (Second Realism Check)} \\
\hline
You are an expert in analyzing images for signs of modification by AI models and assessing their realism. Your task is to examine two images and determine which one is more realistic. In addition, provide a brief explanation highlighting the key elements that informed your judgment, such as unusual textures, unnatural elements, unusual biological phenomena (animals with multiple limbs, heads), unusual arrangement of objects (weirdly clustered objects), or any visible artifacts.

\textbf{\# Output Format} \\
\textbf{Assessment:} (your explanation goes here) \\
\textbf{Verdict:} (final verdict, answer ``First is more realistic'' or ``Second is more realistic'' or ``Both look realistic'') \\
\hline
\end{tabular}

\caption{VLM system prompts for realism checks. The first prompt is for assessing the realism of a single image, and the second prompt is for comparing the realism of two images, original and inpainted.}
\label{fig:vlm-prompts}
\end{figure*}

The prompt design choices were made based on empirical observations from a small validation set. Through our experiments, we found that requiring explanations before verdicts significantly improved assessment quality. Additionally, explicit mention of common artifacts (e.g., unusual textures, biological anomalies) helped focus the model's attention on relevant features.

The VLM was configured with conservative hyperparameters to ensure consistent responses, using a temperature of 0.1 to minimize response variability, a top-p of 1.0 with no nucleus sampling, and a maximum token limit of 2048 to allow for detailed explanations.

\subsection{Detailed Classification Process of UGDA}

Here we present the classification process of the second stage of UGDA in more detail:

\begin{itemize}
    \item Case 1: The VLM identifies $I_i$ in one order and $\hat{I}_i$ in the other, indicating order-dependent bias
    \item Case 2: The VLM consistently identifies $\hat{I}_i$ as more realistic ($s_1 = s_2 = \hat{I}_i$)
    \item Case 3: The VLM responds with ``both'' in one case and chooses $\hat{I}_i$ in the other
    \item Case 4: The VLM consistently responds that both images are equally realistic ($s_1 = s_2 = \text{both}$)
\end{itemize}

In all other response combinations, $\hat{I}_i$ is classified as \textit{non-deceiving}. This classification scheme captures cases where the VLM either consistently prefers the inpainted image or shows uncertainty in its assessment, all of which indicate potential deceptiveness in the synthetic content. Case 1 indicates model uncertainty manifested through order sensitivity, Case 2 represents clear preference for synthetic content, Case 3 captures uncertainty biased toward synthetic content, and Case 4 reflects complete inability to distinguish between real and synthetic content. These patterns suggest varying degrees of image deceptiveness that warrant classification as deceiving. 

\subsection{Human Benchmark}
The application implementing the human benchmark was developed using Gradio \cite{abid2019gradiohasslefreesharingtesting}. At the start of the demo, participants were provided with clear instructions on how to proceed. They were asked to evaluate whether an image had been inpainted and to draw bounding boxes around the areas they believed to be inpainted. Additionally, participants were asked to complete a short demographics questionnaire before beginning the task. The questions included are shown in \cref{tab:questionnaire}.

\begin{table*}[ht]
\centering
\small
\setlength{\tabcolsep}{0.5em} 
\begin{tabular}{|p{0.3\textwidth}|p{0.3\textwidth}|p{0.3\textwidth}|}
\hline
\textbf{Gender} & \textbf{Age Range} & \textbf{Highest Education Completed} \\
\hline
$\square$ Male & $\square$ Under 18 & $\square$ EQF 1-4 (Primary/Upper Secondary) \\
$\square$ Female & $\square$ 18-24 & $\square$ EQF 5 (Post-Secondary Diploma) \\
$\square$ Other & $\square$ 25-34 & $\square$ EQF 6 (Bachelor's Degree) \\
$\square$ Prefer not to say & $\square$ 35-44 & $\square$ EQF 7 (Master's Degree) \\
& $\square$ 45-54 & $\square$ EQF 8 (Doctorate) \\
& $\square$ 55-64 & $\square$ Prefer not to say \\
& $\square$ 65+ & \\
& $\square$ Prefer not to say & \\
\hline
\textbf{Current Education Status} & \textbf{Familiarity with AI-Generated Images} & \textbf{Knowledge of Digital Photography} \\
\hline
$\square$ EQF 1-4 (Primary/Upper Secondary) & $\square$ Very familiar & $\square$ Professional level \\
$\square$ EQF 5 (Post-Secondary Diploma) & $\square$ Somewhat familiar & $\square$ Advanced \\
$\square$ EQF 6 (Bachelor's Degree) & $\square$ Slightly familiar & $\square$ Intermediate \\
$\square$ EQF 7 (Master's Degree) & $\square$ Not familiar & $\square$ Basic \\
$\square$ EQF 8 (Doctorate) & $\square$ Prefer not to say & $\square$ No experience \\
$\square$ Not currently studying & & $\square$ Prefer not to say \\
$\square$ Prefer not to say & & \\
\hline
\end{tabular}
\caption{Demographic and background questionnaire.}
\label{tab:questionnaire}
\end{table*}

\section{Unmasked Area Preservation}

 Fidelity metrics such as Mean Squared Error (MSE), Mean Absolute Error (MAE), Peak Signal-to-Noise Ratio (PSNR), and Learned Perceptual Image Patch Similarity (LPIPS) \cite{zhang2018unreasonableeffectivenessdeepfeatures} assess the preservation of the non-inpainted area. Fidelity metrics are most meaningful for FR images, whereas for SP images, where the compared areas are nearly identical, they provide limited insight. The results are presented in  \cref{tab:metrics_fidelity}. When comparing our dataset with existing alternatives, our \datasetname significantly outperforms TGIF across all FR image fidelity metrics. We achieve a PSNR of 25.79 compared to TGIF's 14.41, with substantially better LPIPS (44.24 vs 289.55), MSE (5.08 vs 60.43), and MAE (41.16 vs 173.97). These improvements indicate that our inpainting approach better preserves the original image context while implementing the intended modifications. CocoGlide is not included in \cref{tab:metrics_fidelity} since it contains only SP images.\looseness=-1

\begin{table}[!htb]
\centering
\small
\begin{tabular}{lccccc}
\hline
\textbf{Dataset} & \textbf{PSNR↑} & \textbf{LPIPS↓} & \textbf{MSE↓} & \textbf{MAE↓} & \textbf{SSIM↑} \\
\hline
TGIF & 14.4 & 289.6 & 60.4 & 174.0 & 0.53 \\
Ours & \textbf{25.8} & \textbf{44.2} & \textbf{5.1} & \textbf{41.2} & \textbf{0.81} \\
\hline
\end{tabular}
\caption{Comparison based on fidelity metrics for FR images. Top: object labels vs. Caption prompts vs LLM prompts. Bottom: our dataset vs. TGIF. LPIPS, MSE, and MAE values are $\times 10^3$. CocoGlide is omitted as it contains only SP images.}
\label{tab:metrics_fidelity}
\vspace{-10pt}
\end{table}

\section{Localization and Detection Results}
In this section, we present extended results on localization and detection, studying various cases for forensic models PSCC-Net \cite{luy2022pscc}, CAT-Net \cite{kwon2022catnet}, TruFor \cite{guillaro2023trufor}, and MMFusion (MMFus) \cite{triaridis2023mmfusion}.

Since mean IoU and detection Accuracy require a threshold, we also report AUC metrics in Table \ref{tab:performance_comparison_auc} as they are threshold-agnostic. We calculate AUC at both pixel level (localization) and image level (detection). For localization AUC, we resize and flatten all localization maps and their ground truths into two vectors for ROC computation in each group. Note that detection AUC cannot be calculated for SP and FR sets, as they contain only forged images.
The AUC metrics further confirm that retraining improves performance significantly. TruFor's localization AUC increases from 68.9\% to 99.5\% for in-domain and 79.9\% to 99.6\% for out-of-domain testing. Similarly, CAT-Net shows strong in-domain gains (60.0\% to 95.6\%) but smaller out-of-domain improvement (51.7\% to 90.8\%). Domain generalization varies across models. While retrained CAT-Net achieves high in-domain detection AUC (99.6\%), it drops to 76\% for out-of-domain. In contrast, retrained TruFor maintains consistent performance across domains in both localization (99.5\%/99.6\%) and detection (99.2\%/98.0\%).
SP localization remains easier for both original and retrained models than FR, with all retrained models achieving localization AUCs above 90.0\% for SP tasks. For FR images, original models perform poorly (AUCs 52.7\%-74.1\%) but show clear improvements after retraining, with TruFor reaching 98.6\% AUC.

\begin{figure}[!htb]
\centering
\begin{subfigure}{\columnwidth}
   \includegraphics[width=0.49\linewidth]{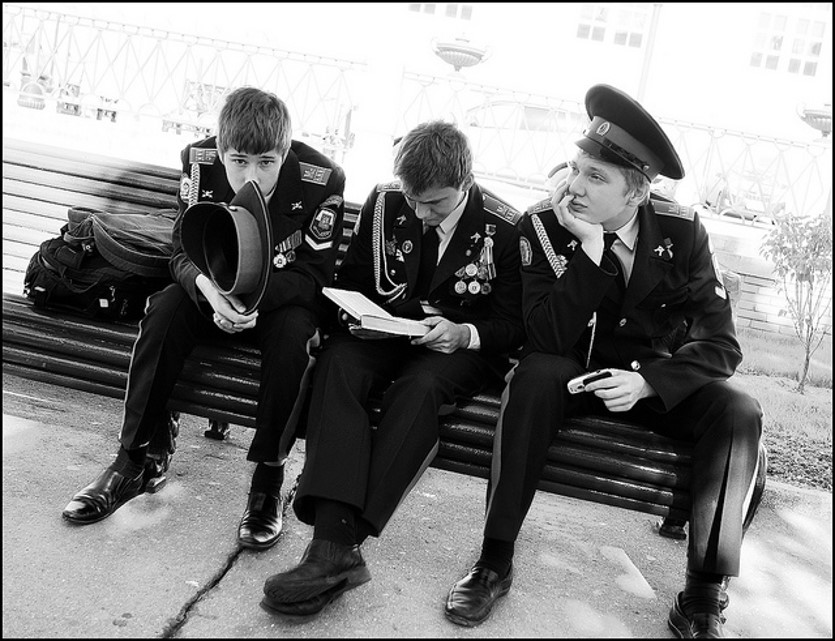}
   \includegraphics[width=0.49\linewidth]{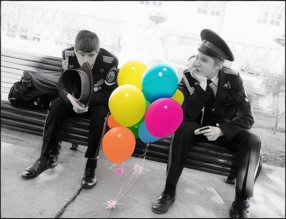}
\end{subfigure}
\caption{Examples of inpainted images using HDPainter. Left is the original image and right is the inpainted result. HDPainter's inpainting significantly affects regions beyond the masked area, evident in the faces of the people next to the balloons.}
\label{fig:examples}
\end{figure}

Tables \ref{tab:performance_comparison_inp_mdl} demonstrate model performance across inpainting methods. The SP/FR performance gap persists across methods - e.g., TruFor† achieves 89.9 IoU on BrushNet-SP versus 77.6 on BrushNet-FR, with similar patterns for PowerPaint (90.9 SP, 78.1 FR). HDPainter presents the most challenging case, with TruFor† achieving only 55.4 IoU compared to 76.6-78.1 for other FR methods. BrushNet and PowerPaint FR manipulations are more detectable, likely due to distinctive inpainting artifacts. This trend holds across models, with MMFusion† achieving 58.0 IoU on PowerPaint-FR but only 42.5 on HDPainter-FR.
For SP cases, InpaintAnything is well-detected even by original models (34.9-63.6 IoU), likely due to its traditional copy-paste operations. HDPainter remains challenging in SP scenarios, showing consistently lower scores. HDPainter's difficulty could stem from its greater impact beyond masked regions in FR cases (Figure \ref{fig:examples}) and its blending/upscaling post-processing in SP cases.

Table \ref{tab:double_inpainting} compares model performance between single and double inpainting cases. Original models show decreased performance on double inpainting, particularly evident in CAT-Net's IoU drop from 27.2 to 7.0. This suggests that multiple manipulations make detection more challenging for models not specifically trained for such cases. Interestingly, retrained models show more robust performance across both scenarios. TruFor† maintains similar IoU scores (81.0/79.0) while slightly improving in accuracy (95.0/99.0). CAT-Net† even shows a small improvement in IoU for double inpainting (42.4 to 49.0) while maintaining near-perfect accuracy (99.9/100.0), suggesting that retraining helps models adapt to more complex manipulation patterns. \looseness=-1

\begin{table}[!htb]
\centering
\setlength{\tabcolsep}{1.2mm}
\begin{tabular}{clcccc}
\hline
\multirow{2}{*}{Data} & \multirow{2}{*}{Model} & \multicolumn{2}{c}{Mean IoU} & \multicolumn{2}{c}{Accuracy} \\
 & & Single & Double & Single & Double \\
\cmidrule(lr){1-2} \cmidrule(lr){3-4} \cmidrule(lr){5-6}
\multirow{4}{*}{\rotatebox[origin=c]{90}{\textbf{Original}}} & PSCC-Net & 15.8 & 16.0 & 37.6 & 39.0 \\
 & CAT-Net & 27.2 & 7.0 & 66.6 & 49.0 \\
 & MMFusion & 34.5 & 22.0 & 47.9 & 42.0 \\
 & TruFor & 31.8 & 21.0 & 29.6 & 32.0 \\
\cmidrule(lr){1-2} \cmidrule(lr){3-4} \cmidrule(lr){5-6}
\multirow{8}{*}{\rotatebox[origin=c]{90}{\textbf{\datasetname}}} & PSCC-Net & 33.7 & 27.0 & 44.3 & 41.0 \\
 & & {\footnotesize \textcolor{green!50!black}{+17.9}} & {\footnotesize \textcolor{green!50!black}{+11.0}} & {\footnotesize \textcolor{green!50!black}{+6.7}} & {\footnotesize \textcolor{green!50!black}{+2.0}} \\
 & CAT-Net & 42.4 & 49.0 & 99.9 & 100.0 \\
 & & {\footnotesize \textcolor{green!50!black}{+15.2}} & {\footnotesize \textcolor{green!50!black}{+42.0}} & {\footnotesize \textcolor{green!50!black}{+33.3}} & {\footnotesize \textcolor{green!50!black}{+51.0}} \\
 & MMFusion & 64.0 & 59.0 & 84.1 & 88.0 \\
 & & {\footnotesize \textcolor{green!50!black}{+29.5}} & {\footnotesize \textcolor{green!50!black}{+37.0}} & {\footnotesize \textcolor{green!50!black}{+36.2}} & {\footnotesize \textcolor{green!50!black}{+46.0}} \\
 & TruFor & 81.0 & 79.0 & 95.0 & 99.0 \\
 & & {\footnotesize \textcolor{green!50!black}{+49.2}} & {\footnotesize \textcolor{green!50!black}{+58.0}} & {\footnotesize \textcolor{green!50!black}{+65.4}} & {\footnotesize \textcolor{green!50!black}{+67.0}} \\
\hline
\end{tabular}
\caption{Performance comparison between original models and models retrained on \datasetname. The table shows Mean IoU and Accuracy for both single and double inpainting manipulations. \textcolor{green!50!black}{Green} numbers indicate improvements compared to the original models. Retraining on \datasetname~yields significant performance improvements across all metrics and models, with TruFor showing the most substantial gains in both localization (Mean IoU) and detection (Accuracy).}
\label{tab:double_inpainting}
\end{table}

\begin{table}[!htb]
\centering
\setlength{\tabcolsep}{1mm}
\begin{tabular}{clcccccc}
\hline
\multirow{2}{*}{Data} & \multirow{2}{*}{Model} & \multicolumn{4}{c}{AUC (loc)} & \multicolumn{2}{c}{AUC (det)} \\
 & & id & ood & SP & FR & id & ood \\
\cmidrule(lr){1-2} \cmidrule(lr){3-6} \cmidrule(lr){7-8}
\multirow{4}{*}{\rotatebox[origin=c]{90}{\textbf{Original}}} & CAT-Net & 60.0 & 51.7 & 69.9 & 58.7 & 67.2 & 50.8 \\
 & PSCC-Net & 71.6 & 59.2 & 64.8 & 52.7 & 83.4 & 55.8 \\
 & MMFusion & 76.5 & 76.0 & 84.9 & 70.5 & 70.5 & 65.6 \\
 & TruFor & 68.9 & 79.9 & 81.1 & 74.1 & 72.3 & 65.1 \\
\cmidrule(lr){1-2} \cmidrule(lr){3-6} \cmidrule(lr){7-8}
\multirow{8}{*}{\rotatebox[origin=c]{90}{\textbf{\datasetname}}} & CAT-Net & 95.6 & 90.8 & 93.3 & 88.8 & 99.6 & 76.7 \\
 & & {\footnotesize \textcolor{green!50!black}{+35.5}} & {\footnotesize \textcolor{green!50!black}{+39.1}} & {\footnotesize \textcolor{green!50!black}{+23.4}} & {\footnotesize \textcolor{green!50!black}{+30.1}} & {\footnotesize \textcolor{green!50!black}{+32.4}} & {\footnotesize \textcolor{green!50!black}{+26.0}} \\
 & PSCC-Net & 83.5 & 84.2 & 90.0 & 68.7 & 80.8 & 74.2 \\
 & & {\footnotesize \textcolor{green!50!black}{+11.8}} & {\footnotesize \textcolor{green!50!black}{+25.0}} & {\footnotesize \textcolor{green!50!black}{+25.2}} & {\footnotesize \textcolor{green!50!black}{+16.1}} & {\footnotesize \textcolor{red}{-2.6}} & {\footnotesize \textcolor{green!50!black}{+18.4}} \\
 & MMFusion & 96.8 & 95.0 & 98.5 & 90.9 & 98.2 & 89.9 \\
 & & {\footnotesize \textcolor{green!50!black}{+20.3}} & {\footnotesize \textcolor{green!50!black}{+19.0}} & {\footnotesize \textcolor{green!50!black}{+13.6}} & {\footnotesize \textcolor{green!50!black}{+20.4}} & {\footnotesize \textcolor{green!50!black}{+27.8}} & {\footnotesize \textcolor{green!50!black}{+24.3}} \\
 & TruFor & 99.5 & 99.6 & 99.9 & 98.6 & 99.2 & 98.0 \\
 & & {\footnotesize \textcolor{green!50!black}{+30.5}} & {\footnotesize \textcolor{green!50!black}{+19.7}} & {\footnotesize \textcolor{green!50!black}{+18.8}} & {\footnotesize \textcolor{green!50!black}{+24.6}} & {\footnotesize \textcolor{green!50!black}{+26.9}} & {\footnotesize \textcolor{green!50!black}{+33.0}} \\
\hline
\end{tabular}
\caption{Performance comparison of image forensics methods across different domains. The table shows AUC scores for both localization and detection tasks, comparing original models with those retrained on our dataset. ``ID'' indicates in-domain and ``OOD'' indicates out-of-domain performance, while SP (Splicing) and FR (Fully Regenerated) represent different forgery types. \textcolor{green!50!black}{Green} numbers show improvements and \textcolor{red}{red} numbers show decreases compared to original models. Retraining on our dataset yields significant performance improvements across most metrics and models.}
\label{tab:performance_comparison_auc}
\end{table}

\begin{table*}[!htb]
\centering
\setlength{\tabcolsep}{1mm}
\scalebox{0.96}{
\begin{tabular}{clcccccccccccccccccc}
\hline
\multirow{3}{*}{Data} & \multirow{3}{*}{Model} & \multicolumn{8}{c}{FR} & \multicolumn{10}{c}{SP} \\
\cmidrule(lr){3-10} \cmidrule(lr){11-20}
 & & \multicolumn{2}{c}{BN} & \multicolumn{2}{c}{CN} & \multicolumn{2}{c}{HDP} & \multicolumn{2}{c}{PPt} & \multicolumn{2}{c}{BN} & \multicolumn{2}{c}{HDP} & \multicolumn{2}{c}{IA} & \multicolumn{2}{c}{PPt} & \multicolumn{2}{c}{RA} \\
\cmidrule(lr){3-4} \cmidrule(lr){5-6} \cmidrule(lr){7-8} \cmidrule(lr){9-10} \cmidrule(lr){11-12} \cmidrule(lr){13-14} \cmidrule(lr){15-16} \cmidrule(lr){17-18} \cmidrule(lr){19-20}
 & & IoU & Acc & IoU & Acc & IoU & Acc & IoU & Acc & IoU & Acc & IoU & Acc & IoU & Acc & IoU & Acc & IoU & Acc \\
\hline
\multirow{4}{*}{\rotatebox[origin=c]{90}{\textbf{Original}}} & CN & 3.0 & 36.8 & 9.0 & 40.4 & 5.0 & 34.4 & 4.1 & 48.0 & 31.6 & 72.2 & 1.5 & 58.3 & 62.6 & 98.7 & 22.4 & 67.0 & 52.4 & 96.2 \\
 & PS & 12.2 & 36.6 & 7.9 & 35.9 & 5.7 & 25.9 & 12.7 & 33.8 & 16.5 & 40.3 & 17.3 & 53.8 & 34.9 & 48.4 & 10.6 & 23.2 & 14.6 & 35.5 \\
 & MM & 20.2 & 24.1 & 18.5 & 29.2 & 10.9 & 24.8 & 17.2 & 20.9 & 67.7 & 75.3 & 26.7 & 50.2 & 63.6 & 88.7 & 46.4 & 57.2 & 30.5 & 43.6 \\
 & TF & 22.5 & 14.0 & 21.7 & 14.3 & 12.1 & 9.9 & 21.3 & 10.6 & 56.7 & 49.8 & 23.7 & 26.6 & 60.7 & 70.1 & 42.5 & 35.7 & 20.9 & 21.0 \\
\cmidrule(lr){1-20}
\multirow{8}{*}{\rotatebox[origin=c]{90}{\textbf{\datasetname}}} & CN & 38.3 & 99.9 & 38.9 & 100.0 & 28.3 & 99.9 & 37.9 & 99.7 & 41.5 & 100.0 & 34.3 & 100.0 & 48.5 & 99.8 & 42.0 & 100.0 & 54.0 & 99.9 \\
 & & {\footnotesize \textcolor{green!50!black}{+35.3}} & {\footnotesize \textcolor{green!50!black}{+63.1}} & {\footnotesize \textcolor{green!50!black}{+29.9}} & {\footnotesize \textcolor{green!50!black}{+59.6}} & {\footnotesize \textcolor{green!50!black}{+23.3}} & {\footnotesize \textcolor{green!50!black}{+65.5}} & {\footnotesize \textcolor{green!50!black}{+33.8}} & {\footnotesize \textcolor{green!50!black}{+51.7}} & {\footnotesize \textcolor{green!50!black}{+9.9}} & {\footnotesize \textcolor{green!50!black}{+27.8}} & {\footnotesize \textcolor{green!50!black}{+32.8}} & {\footnotesize \textcolor{green!50!black}{+41.7}} & {\footnotesize \textcolor{green!50!black}{-14.1}} & {\footnotesize \textcolor{green!50!black}{+1.1}} & {\footnotesize \textcolor{green!50!black}{+19.6}} & {\footnotesize \textcolor{green!50!black}{+33.0}} & {\footnotesize \textcolor{green!50!black}{+1.6}} & {\footnotesize \textcolor{green!50!black}{+3.7}} \\
 & PS & 23.8 & 67.9 & 9.2 & 15.2 & 16.5 & 15.5 & 38.1 & 64.1 & 32.3 & 43.2 & 39.7 & 68.0 & 53.0 & 51.7 & 46.2 & 51.4 & 39.8 & 43.9 \\
 & & {\footnotesize \textcolor{green!50!black}{+11.6}} & {\footnotesize \textcolor{green!50!black}{+31.3}} & {\footnotesize \textcolor{green!50!black}{+1.3}} & {\footnotesize \textcolor{green!50!black}{-20.7}} & {\footnotesize \textcolor{green!50!black}{+10.8}} & {\footnotesize \textcolor{green!50!black}{-10.4}} & {\footnotesize \textcolor{green!50!black}{+25.4}} & {\footnotesize \textcolor{green!50!black}{+30.3}} & {\footnotesize \textcolor{green!50!black}{+15.8}} & {\footnotesize \textcolor{green!50!black}{+2.9}} & {\footnotesize \textcolor{green!50!black}{+22.4}} & {\footnotesize \textcolor{green!50!black}{+14.2}} & {\footnotesize \textcolor{green!50!black}{+18.1}} & {\footnotesize \textcolor{green!50!black}{+3.3}} & {\footnotesize \textcolor{green!50!black}{+35.6}} & {\footnotesize \textcolor{green!50!black}{+28.2}} & {\footnotesize \textcolor{green!50!black}{+25.2}} & {\footnotesize \textcolor{green!50!black}{+8.4}} \\
 & MM & 45.1 & 63.8 & 53.4 & 89.4 & 42.5 & 82.1 & 58.0 & 83.5 & 80.5 & 91.7 & 58.5 & 73.0 & 73.6 & 85.1 & 80.7 & 92.7 & 72.2 & 86.3 \\
 & & {\footnotesize \textcolor{green!50!black}{+24.9}} & {\footnotesize \textcolor{green!50!black}{+39.7}} & {\footnotesize \textcolor{green!50!black}{+34.9}} & {\footnotesize \textcolor{green!50!black}{+60.2}} & {\footnotesize \textcolor{green!50!black}{+31.6}} & {\footnotesize \textcolor{green!50!black}{+57.3}} & {\footnotesize \textcolor{green!50!black}{+40.8}} & {\footnotesize \textcolor{green!50!black}{+62.6}} & {\footnotesize \textcolor{green!50!black}{+12.8}} & {\footnotesize \textcolor{green!50!black}{+16.4}} & {\footnotesize \textcolor{green!50!black}{+31.8}} & {\footnotesize \textcolor{green!50!black}{+22.8}} & {\footnotesize \textcolor{green!50!black}{+10.0}} & {\footnotesize \textcolor{green!50!black}{-3.6}} & {\footnotesize \textcolor{green!50!black}{+34.3}} & {\footnotesize \textcolor{green!50!black}{+35.5}} & {\footnotesize \textcolor{green!50!black}{+41.7}} & {\footnotesize \textcolor{green!50!black}{+42.7}} \\
 & TF & 77.6 & 92.1 & 76.6 & 94.6 & 55.4 & 92.5 & 78.1 & 98.3 & 89.9 & 96.0 & 80.6 & 95.0 & 84.7 & 92.5 & 90.9 & 98.7 & 88.0 & 96.6 \\
 & & {\footnotesize \textcolor{green!50!black}{+55.1}} & {\footnotesize \textcolor{green!50!black}{+78.1}} & {\footnotesize \textcolor{green!50!black}{+54.9}} & {\footnotesize \textcolor{green!50!black}{+80.3}} & {\footnotesize \textcolor{green!50!black}{+43.3}} & {\footnotesize \textcolor{green!50!black}{+82.6}} & {\footnotesize \textcolor{green!50!black}{+56.8}} & {\footnotesize \textcolor{green!50!black}{+87.7}} & {\footnotesize \textcolor{green!50!black}{+33.2}} & {\footnotesize \textcolor{green!50!black}{+46.2}} & {\footnotesize \textcolor{green!50!black}{+56.9}} & {\footnotesize \textcolor{green!50!black}{+68.4}} & {\footnotesize \textcolor{green!50!black}{+24.0}} & {\footnotesize \textcolor{green!50!black}{+22.4}} & {\footnotesize \textcolor{green!50!black}{+48.4}} & {\footnotesize \textcolor{green!50!black}{+63.0}} & {\footnotesize \textcolor{green!50!black}{+67.1}} & {\footnotesize \textcolor{green!50!black}{+75.6}} \\
\hline
\end{tabular}
}
\caption{Performance comparison of image forensics methods
CAT-Net (CN), MMFusion (MM), PSCC-Net (PS), and TruFor (TF) for both FR (Fully Regenerated) and SP (Splicing) scenarios. For each method, Mean IoU and Accuracy (Acc) scores are shown. Inpainting Models: BN (BrushNet), CN (ControlNet), HDP (HDPainter), PPt (PowerPaint), IA (InpaintAnything), and RA (RemoveAnything). The green values in the second row for each retrained model indicate the improvement over the corresponding original model.}
\label{tab:performance_comparison_inp_mdl}
\end{table*}

\subsection{Analysis of Model Detection Performance}

The varying performance across models can be attributed to their architectural choices and training strategies. TruFor's superior performance likely stems from its extensively pretrained Noiseprint++ component, which was trained using self-supervised methods on images with diverse processing procedures. While MMFusion shares architectural similarities with TruFor, its use of multiple modalities may lead to overfitting, potentially explaining its lower performance compared to TruFor. On the other hand, PSCC-Net's poor localization performance can be attributed to its relatively small model size, suggesting possible underfitting. CAT-Net's performance is particularly affected by our evaluation setup for two reasons. First, its design leverages JPEG double compression artifacts for detection, but our dataset contains PNG images where quantization tables are not preserved. Second, while JPEG compression was intrinsic to CAT-Net's original training data, it lacks explicit augmentations for compression robustness unlike other models. This explains its vulnerability to JPEG compression artifacts compared to models with more robust training strategies.

\subsection{Model Compression Robustness Analysis}

\begin{figure}[!htb]
    \centering
    \includegraphics[width=0.46\textwidth]{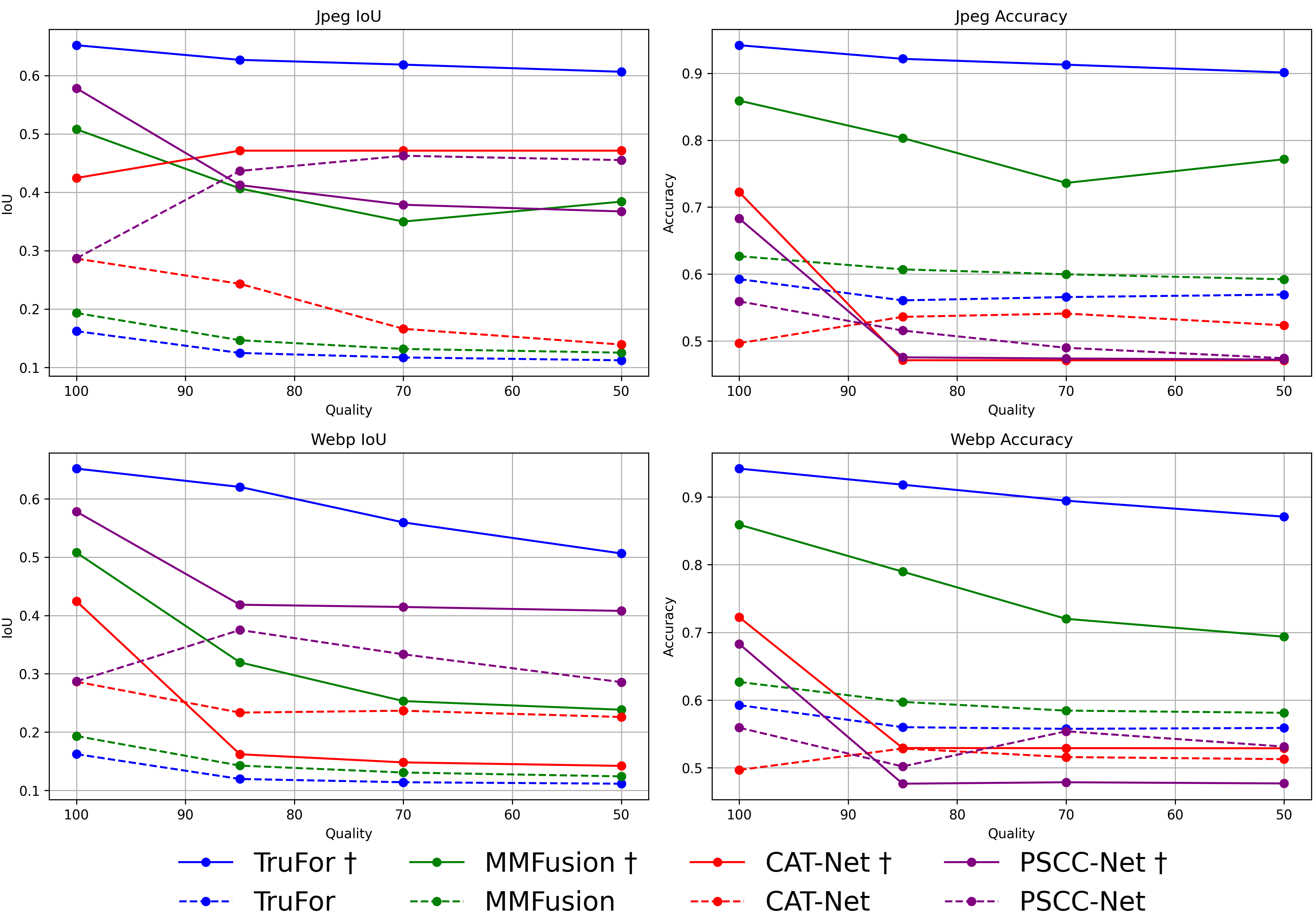}
    \label{fig:compression_metrics}
    \caption{Robustness of model detection performance under compression. Top row shows model detection performance when subjected to JPEG compression at varying quality levels, while bottom row shows detection performance under WEBP compression.} 
    \label{fig:compression_det}
\end{figure}

We evaluate model robustness against JPEG and WEBP image compression at quality levels $0.85$, $0.7$, and $0.5$. Figure \ref{fig:compression_det} presents detection and localization results. Retrained TruFor shows the strongest resilience, maintaining stable performance across quality levels for both compression types. WEBP compression affects performance more than JPEG, particularly for localization tasks. All models show higher degradation in IoU scores compared to accuracy metrics, indicating that manipulation localization is more sensitive to compression artifacts compared with detection. \looseness=-1

\subsection{Comparison with human performance}

In this section, we present extended results of our human evaluation study, analyzing both the participants' detection performance in more cases and the relationship between demographic factors and classification accuracy.

Participants included 26 males, 6 females, and 10 undisclosed. Ages ranged from 18 to 65+, with the largest group being 18-24 (19), followed by 25-34 (8) and 35-44 (6). Users were asked to detect inpainting and draw bounding boxes around suspected manipulated regions. For IoU computation, ground truth masks were converted to bounding boxes.

Table \ref{tab:demographics} shows the results of chi-square tests \cite{pearson1900chi} for independence between demographic variables and classification accuracy (i.e., the proportion of correctly identified images), after filtering out participants with fewer than 20 votes (reducing the sample from 42 to 34 participants). The chi-square tests indicated that the independence hypothesis could not be rejected in any case, and the effect sizes (measured by Cramer's V \cite{cramer1946v}) showed negligible to weak associations, suggesting that these demographic factors have limited practical significance.

\begin{table}[!htb]
\centering
\begin{tabular}{lcc}
\hline
Demographic Factor & Cramer's V & p-value \\
\hline
Gender & 0.0661 & 0.0014 \\
Age & 0.0988 & 0.0003 \\
Education Completed & 0.0929 & 0.0004 \\
Current Education & 0.0986 & 0.0001 \\
AI Familiarity & 0.0811 & 0.0013 \\
DIP Familiarity & 0.0978 & 0.0002 \\
\hline
\end{tabular}
\caption{Chi-square test results for demographic factors vs verification accuracy}
\label{tab:demographics}
\end{table}

Table \ref{tab:human_comparison} presents the comparison between human evaluators and automated models. Human performance reached 67.4\% accuracy and 15.2 IoU, significantly lower than retrained models like TruFor (95.3\% accuracy, 68.3\%) and MMFusion (87.9\% accuracy, 50.9\% IoU). Results are broken down into four categories: \textit{All} represents performance on the complete test set, while \textit{Deceiving}, \textit{Non-Deceiving} and \textit{Intermediate} correspond to UGDA's classification of images based on their potential to fool human perception. The \textit{Intermediate} category includes images that passed the initial realism check but not the second. Users particularly struggled with deceiving images (35.2\% accuracy, 12.9\% IoU) compared to non-deceiving ones (73.7\% accuracy, 40.4\% IoU), validating UGDA's effectiveness in identifying manipulations that are challenging for human perception. Also, the performance of humans on the intermediate category (59.7\% accuracy, 28.3\% IoU) confirms that the second stage is indeed effective in discarding images that are not truly deceiving. In contrast, retrained models maintain high performance even on these challenging cases, with TruFor achieving 98.9\% accuracy and 87.4\% IoU on deceiving images. The performance gap between humans and models emphasizes the importance of automated detection methods, particularly for high-quality inpainting that can bypass human perception.

\begin{table}[!htb]
\centering
\setlength{\tabcolsep}{0.9mm}
\begin{tabular}{lcccccccc}
\hline
\multirow{2}{*}{Model} & \multicolumn{4}{c}{Accuracy} & \multicolumn{4}{c}{Mean IoU} \\
& All & Dec. & Int. & ND. & All & Dec. & Int. & ND. \\
\hline
Human & 67.4 & 35.2 & 59.7 & 73.7 & 15.2 & 12.9 & 28.3 & 40.4 \\
PSCC & 32.1 & 29.6 & 37.5 & 29.2 & 14.4 & 15.2 & 14.1 & 13.8 \\
CAT-Net & 62.5 & 70.4 & 59.4 & 57.6 & 19.5 & 29.4 & 14.2 & 14.9 \\
PSCC† & 51.3 & 49.6 & 50.0 & 54.4 & 36.3 & 35.2 & 30.4 & 43.3 \\
TruFor & 27.3 & 35.2 & 25.0 & 21.6 & 29.0 & 35.4 & 24.7 & 27.0 \\
MMFus & 39.9 & 48.8 & 34.4 & 36.4 & 32.1 & 38.1 & 30.9 & 27.3 \\
CAT-Net† & \textbf{100} & \textbf{100} & \textbf{100} & \textbf{100} & 47.6 & 45.8 & 44.5 & 52.6 \\
MMFus† & 90.5 & 89.2 & 90.6 & 91.6 & 69.6 & 66.5 & 70.3 & 72.1 \\
TruFor† & 99.7 & 99.2 & \textbf{100} & \textbf{100} & \textbf{87.6} & \textbf{86.9} & \textbf{86.6} & \textbf{89.4} \\
\hline
\end{tabular}
\caption{Human vs. model performance comparison on inpainting detection. Results show accuracy and IoU for full test set (All) and images classified by UGDA as Deceiving (Dec.) or Non-Deceiving (ND.). † indicates models retrained on our dataset. Bold values indicate the best performance per column.}
\label{tab:human_comparison}
\end{table}

In Table \ref{tab:comparison_dec_und} we see the performance comparison between spliced (SP) and fully regenerated (FR) images, across human evaluators and the forensic models. The results reveal that human performance remains consistent across both manipulation types, showing no significant advantage in detecting either SP (0.34 for Deceiving, 0.76 for Non-Deceiving) or FR manipulations (0.37 for Deceiving, 0.74 for Non-Deceiving) in contrast to the forensic models.

\begin{table}[!htb]
\centering
\setlength{\tabcolsep}{1mm}
\begin{tabular}{lcccccccc}
\hline
\multirow{3}{*}{Model} & \multicolumn{4}{c}{Accuracy} & \multicolumn{4}{c}{IoU} \\
\cmidrule(lr){2-5} \cmidrule(lr){6-9}
& \multicolumn{2}{c}{Dec.} & \multicolumn{2}{c}{Non-Dec.} & \multicolumn{2}{c}{Dec.} & \multicolumn{2}{c}{Non-Dec.} \\
\cmidrule(lr){2-3} \cmidrule(lr){4-5} \cmidrule(lr){6-7} \cmidrule(lr){8-9}
& SP & FR & SP & FR & SP & FR & SP & FR \\
\cmidrule(lr){1-1} \cmidrule(lr){2-9}
Human         & 34.1 & 37.2 & 75.8 & 74.4 & 12.0 & 14.0 & 41.3 & 41.9 \\
TruFor        & 47.4 & 11.8 & 33.7 & 10.7 & 43.5 & 20.2 & 38.0 & 20.0 \\
MMFus         & 61.5 & 25.0 & 50.0 & 23.2 & 49.3 & 18.7 & 42.1 & 17.5 \\
PSCC-Net          & 31.4 & 25.0 & 31.6 & 28.6 & 17.3 & 11.5 & 17.2 & 11.5 \\
PSCC-Net†         & 51.9 & 39.5 & 72.4 & 41.1 & 44.8 & 16.5 & 61.1 & 31.7 \\
CAT-Net       & 84.6 & 44.7 & 77.6 & 41.1 & 44.2 & 4.5 & 30.4 & 4.5 \\
MMFus†        & 92.9 & 84.2 & 91.8 & 92.9 & 78.0 & 49.3 & 82.0 & 67.8 \\
TruFor†       & 99.4 & 98.7 & \textbf{100} & \textbf{100} & 91.7 & 78.9 & 95.3 & 88.0 \\
CAT-Net†      & \textbf{100} & \textbf{100} & \textbf{100} & \textbf{100} & 48.8 & 41.0 & 53.4 & 53.5 \\
\hline
\end{tabular}
\caption{Human vs. model performance comparison on inpainting detection. Results show accuracy and IoU for full test set images classified by UGDA as Deceiving (Dec.) or Non-Deceiving (Non-Dec.), SP and FR. † indicates models retrained on our dataset. Bold values
indicate the best performance per column.}
\label{tab:comparison_dec_und}
\end{table}

\subsection{Qualitative Analysis} 

\begin{figure*}[t]
    \centering
    \setlength{\tabcolsep}{1pt}
    \begin{tabular}{ccccccccc}
        \multicolumn{1}{c}{Inpainted} & 
        \multicolumn{2}{c}{CatNet} & 
        \multicolumn{2}{c}{MMFusion} & 
        \multicolumn{2}{c}{PSCC} & 
        \multicolumn{2}{c}{TruFor} \\
        
        & Orig & Retr & Orig & Retr & Orig & Retr & Orig & Retr \\

        \includegraphics[width=0.105\textwidth]{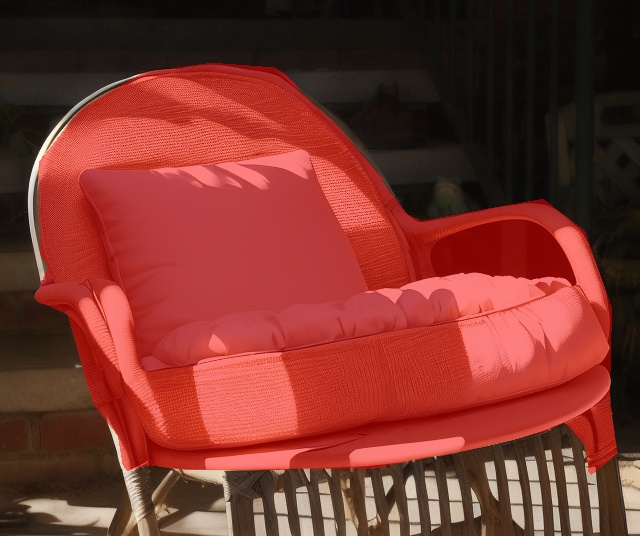} &
        \includegraphics[width=0.105\textwidth]{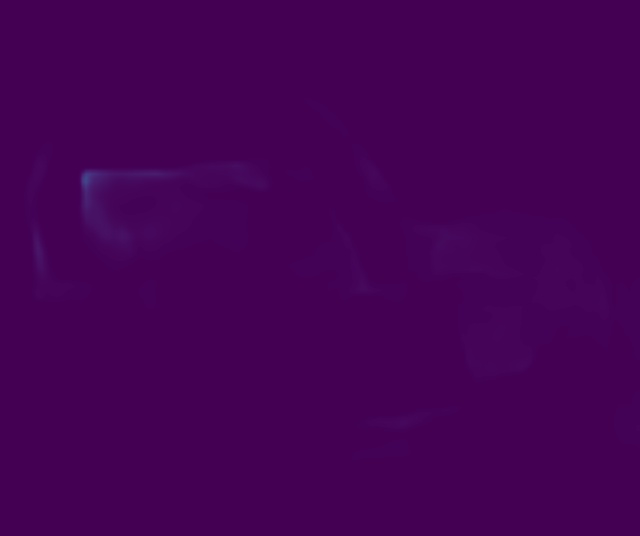} &
        \includegraphics[width=0.105\textwidth]{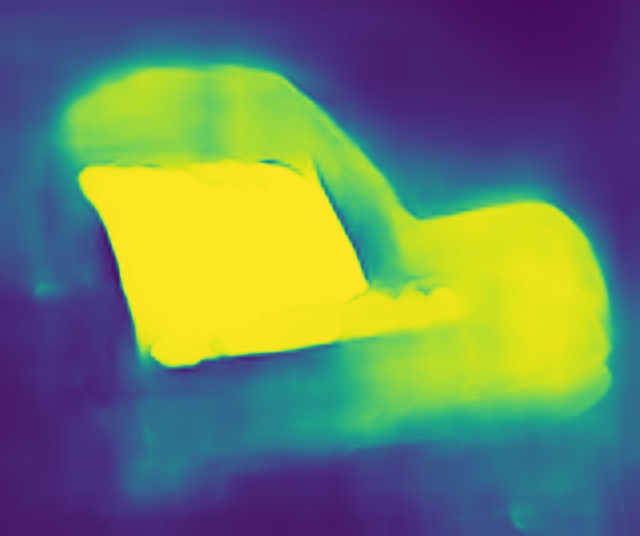} &
        \includegraphics[width=0.105\textwidth]{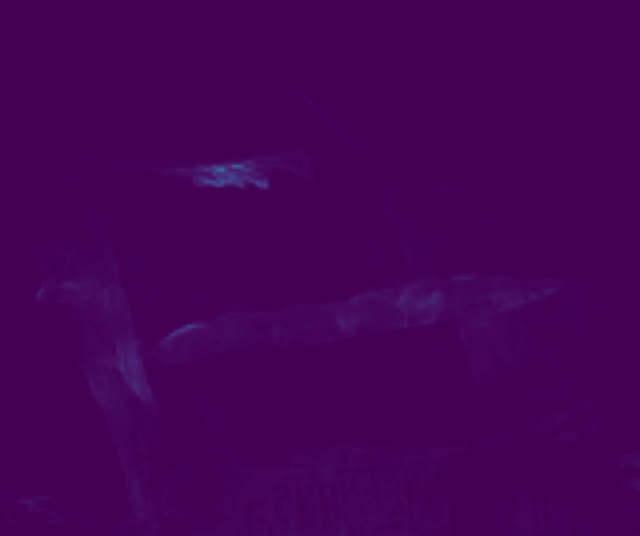} &
        \includegraphics[width=0.105\textwidth]{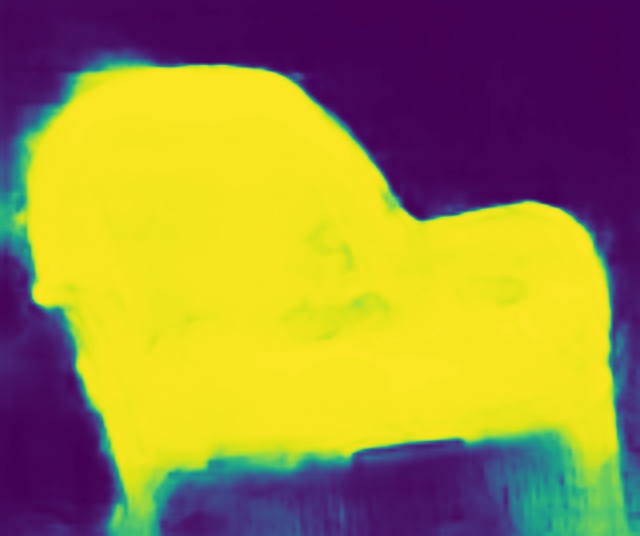} &
        \includegraphics[width=0.105\textwidth]{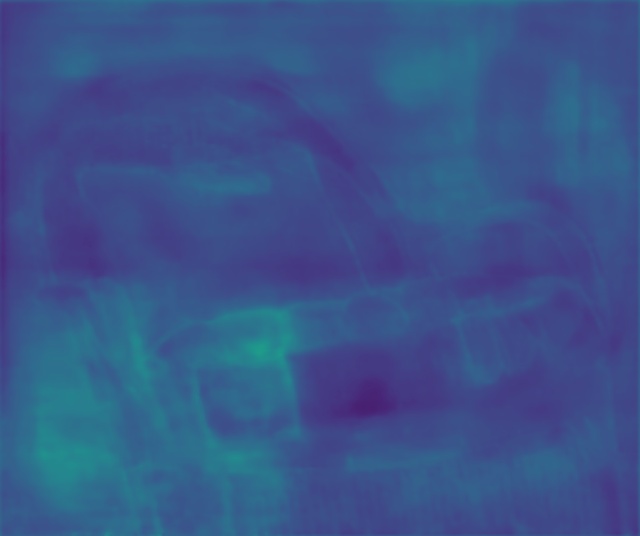} &
        \includegraphics[width=0.105\textwidth]{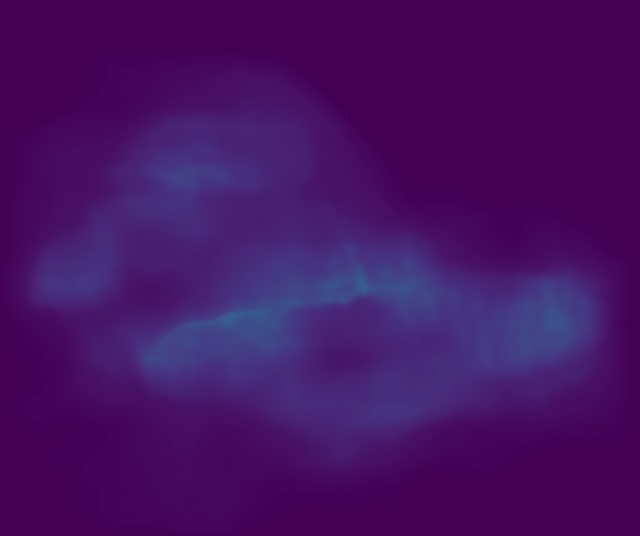} &
        \includegraphics[width=0.105\textwidth]{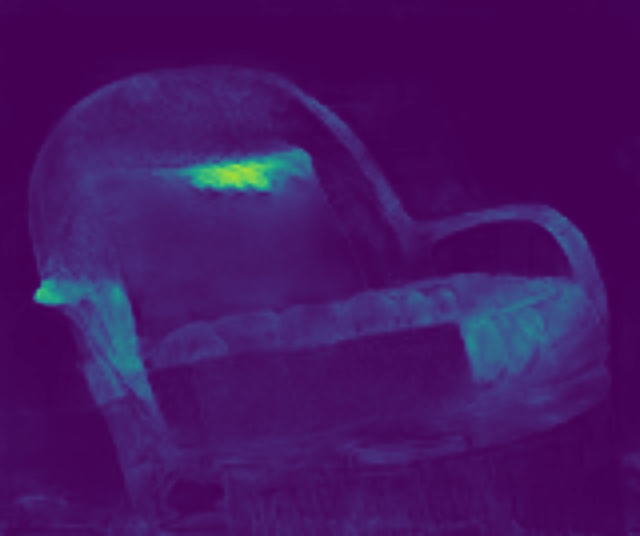} &
        \includegraphics[width=0.105\textwidth]{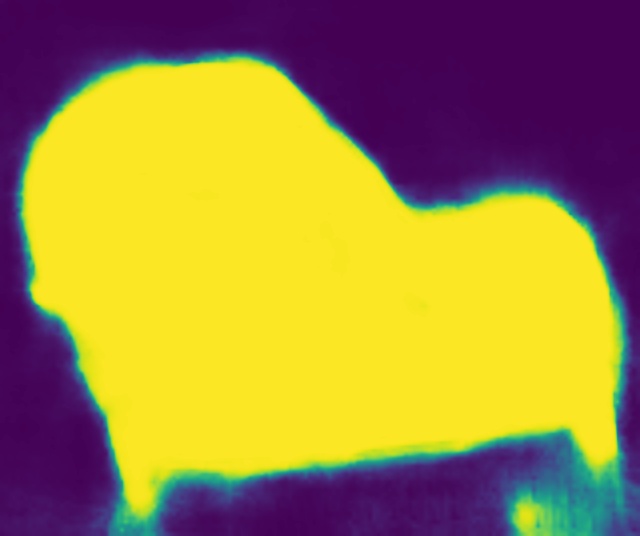} \\

        \includegraphics[width=0.105\textwidth]{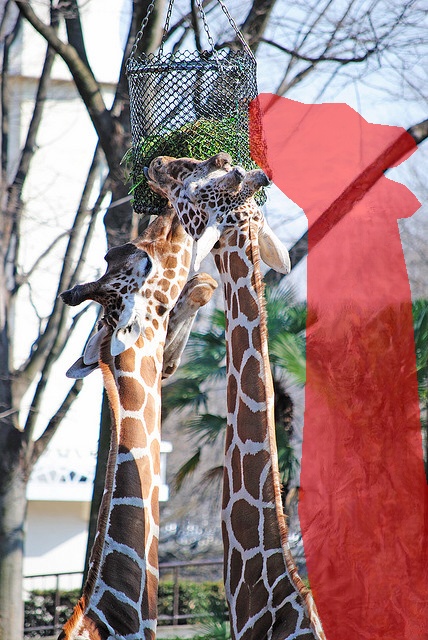} &
        \includegraphics[width=0.105\textwidth]{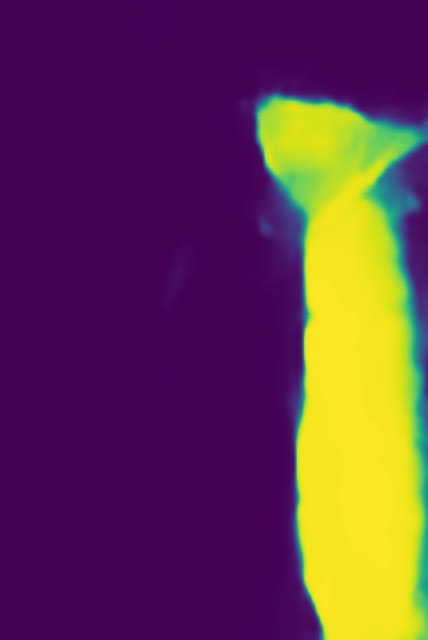} &
        \includegraphics[width=0.105\textwidth]{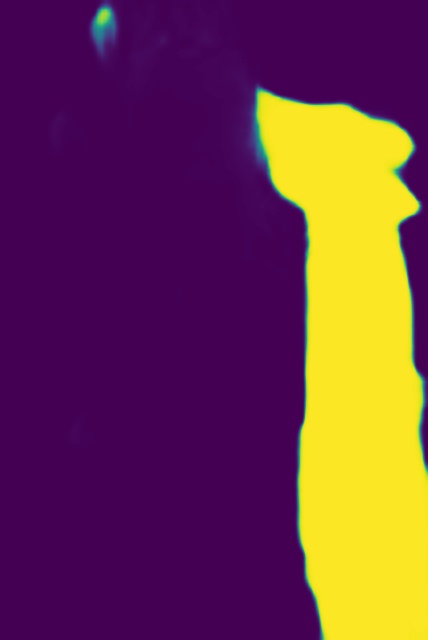} &
        \includegraphics[width=0.105\textwidth]{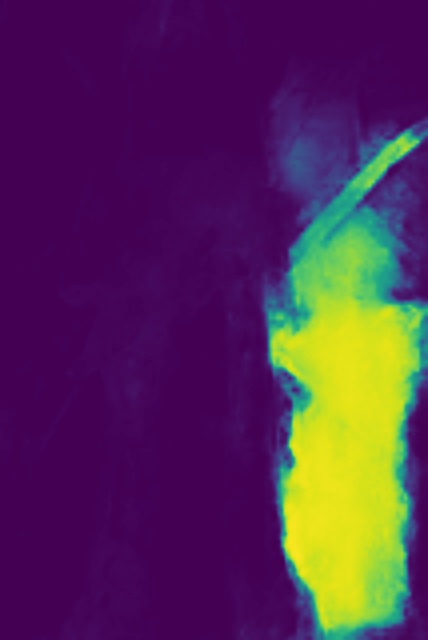} &
        \includegraphics[width=0.105\textwidth]{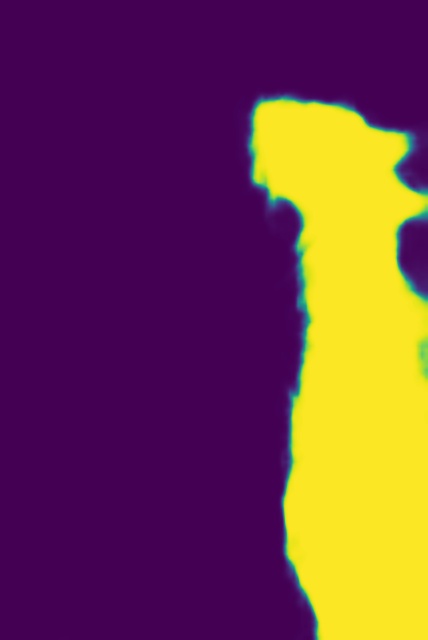} &
        \includegraphics[width=0.105\textwidth]{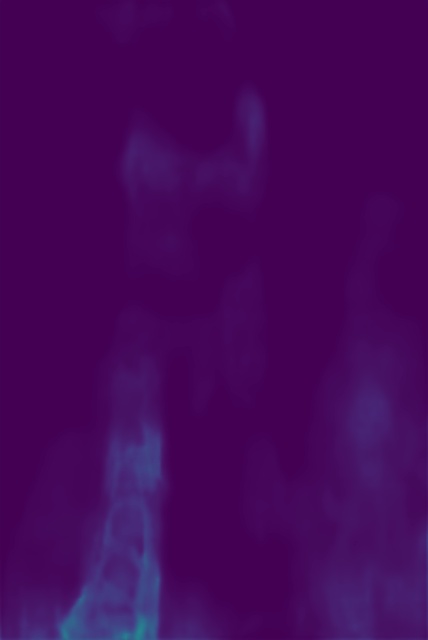} &
        \includegraphics[width=0.105\textwidth]{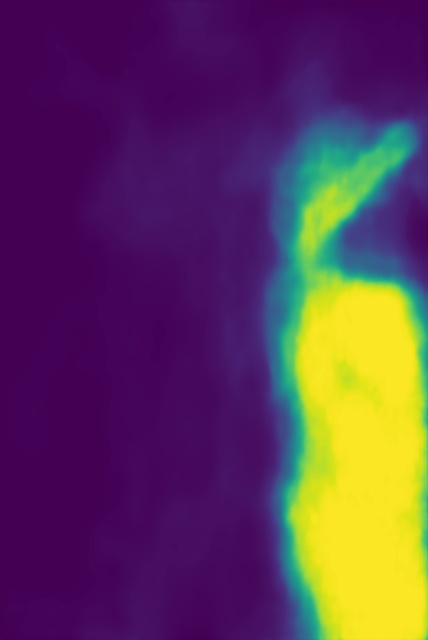} &
        \includegraphics[width=0.105\textwidth]{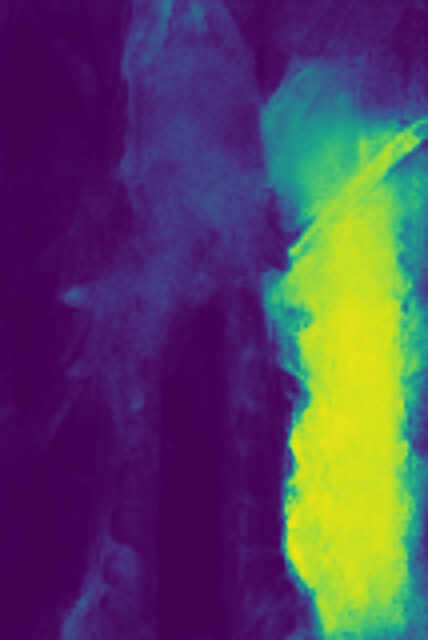} &
        \includegraphics[width=0.105\textwidth]{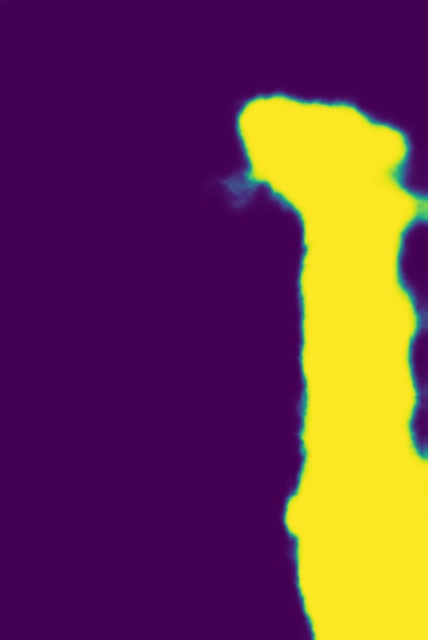} \\

        \includegraphics[width=0.105\textwidth]{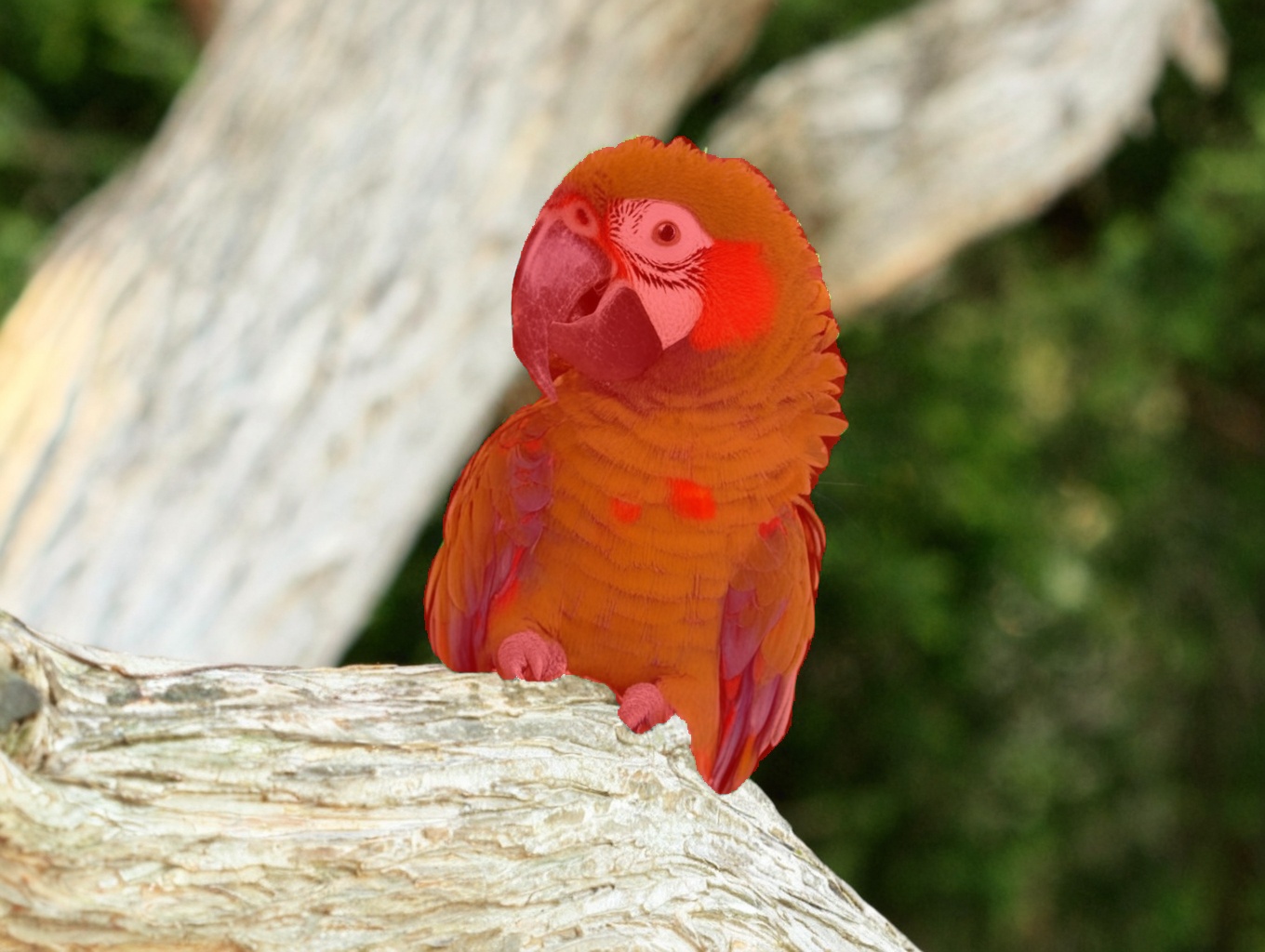} &
        \includegraphics[width=0.105\textwidth]{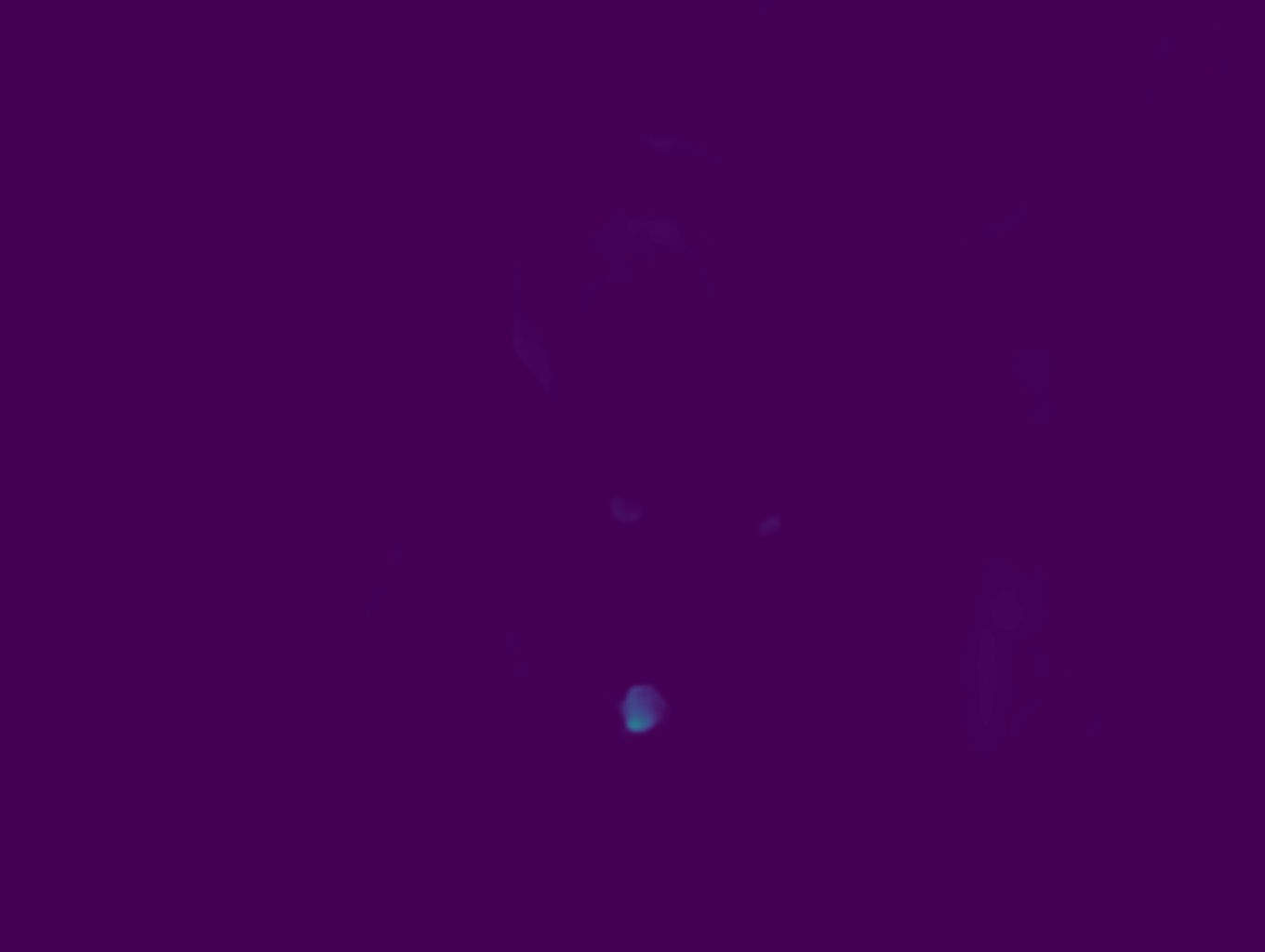} &
        \includegraphics[width=0.105\textwidth]{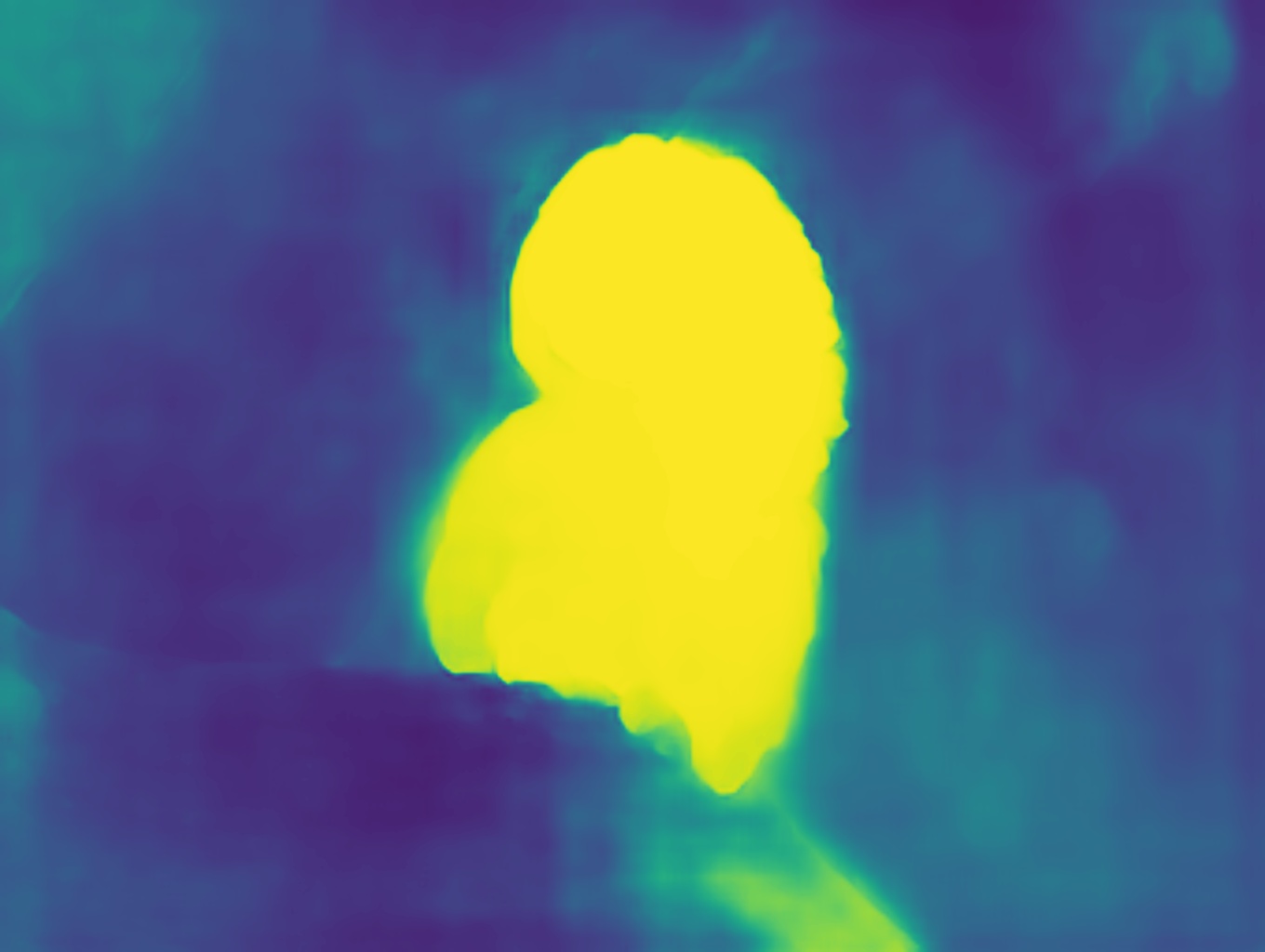} &
        \includegraphics[width=0.105\textwidth]{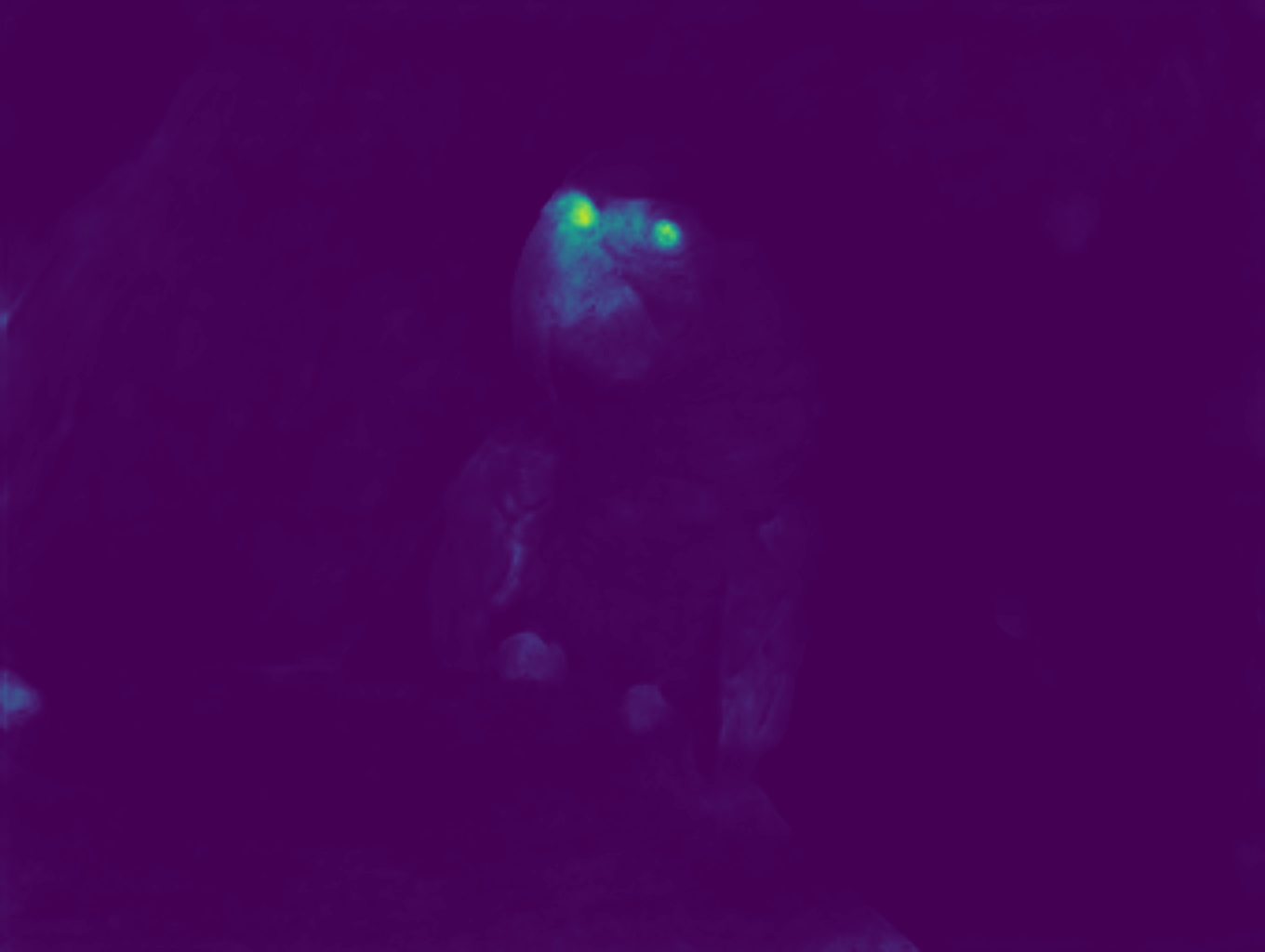} &
        \includegraphics[width=0.105\textwidth]{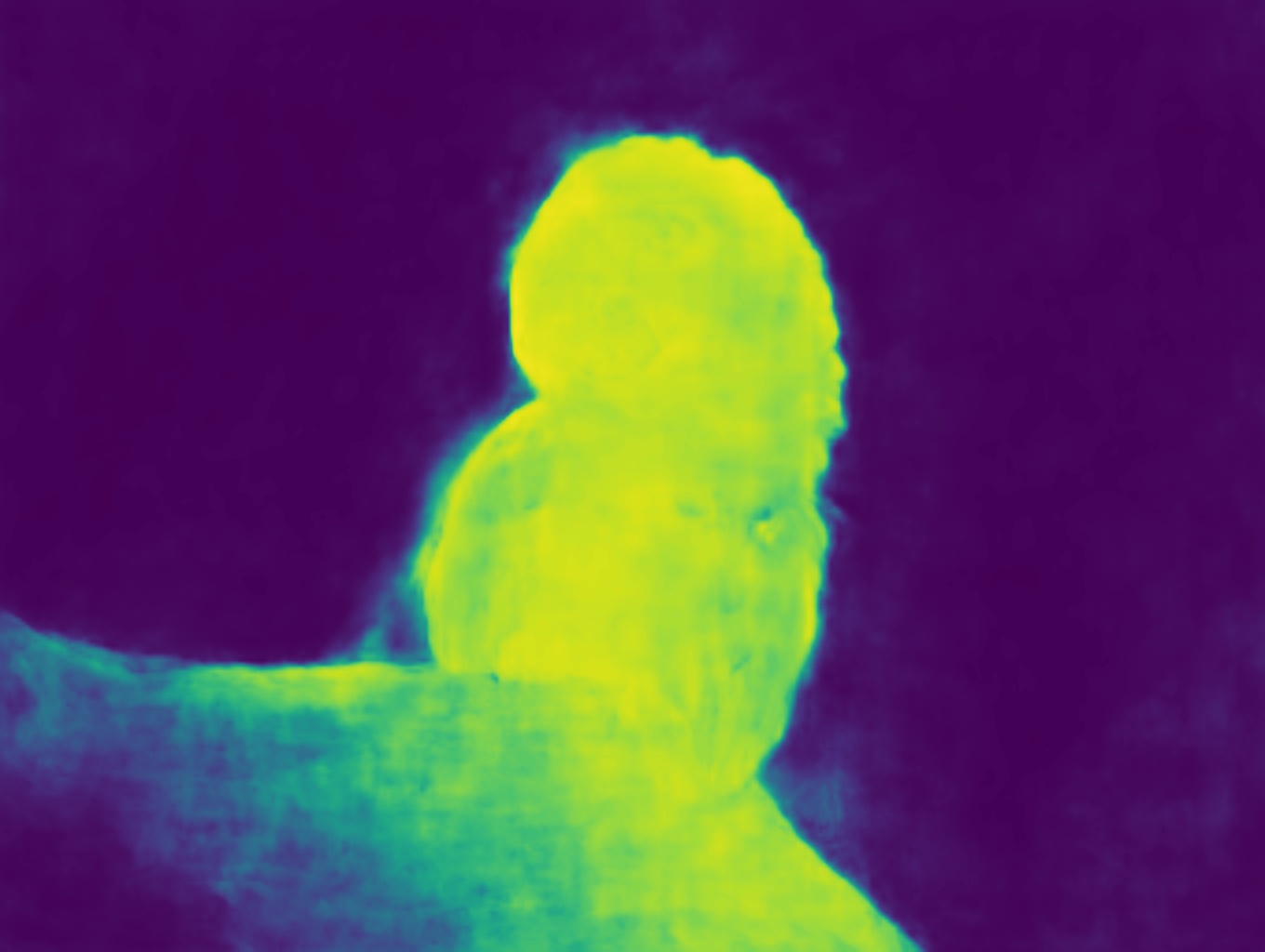} &
        \includegraphics[width=0.105\textwidth]{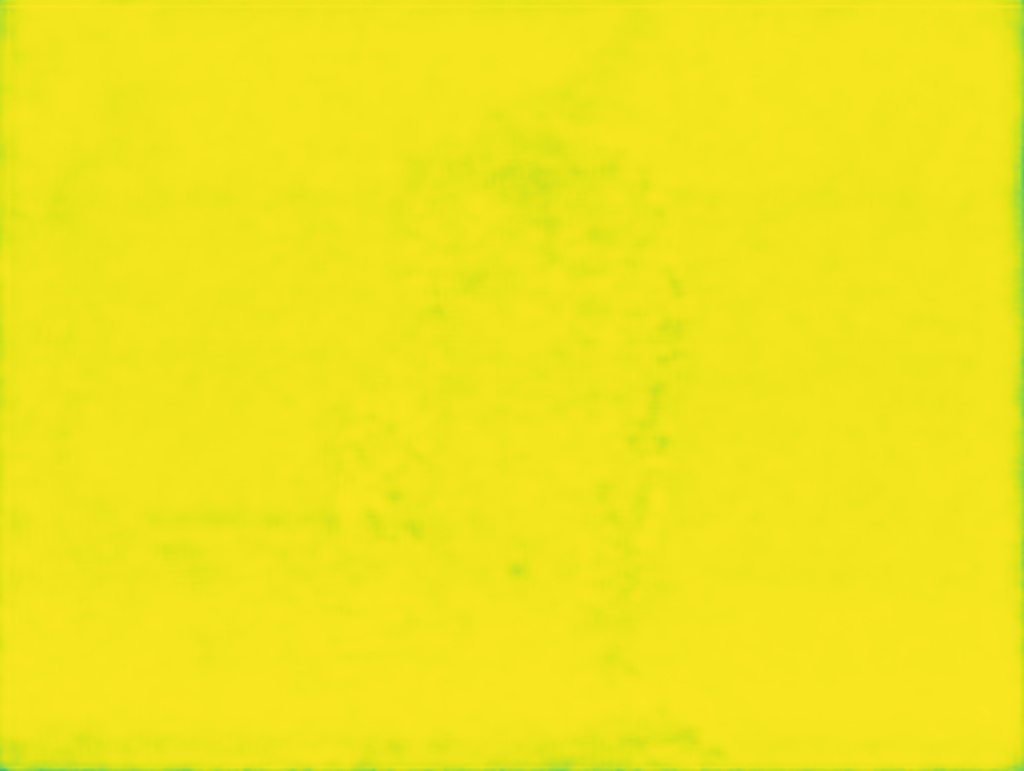} &
        \includegraphics[width=0.105\textwidth]{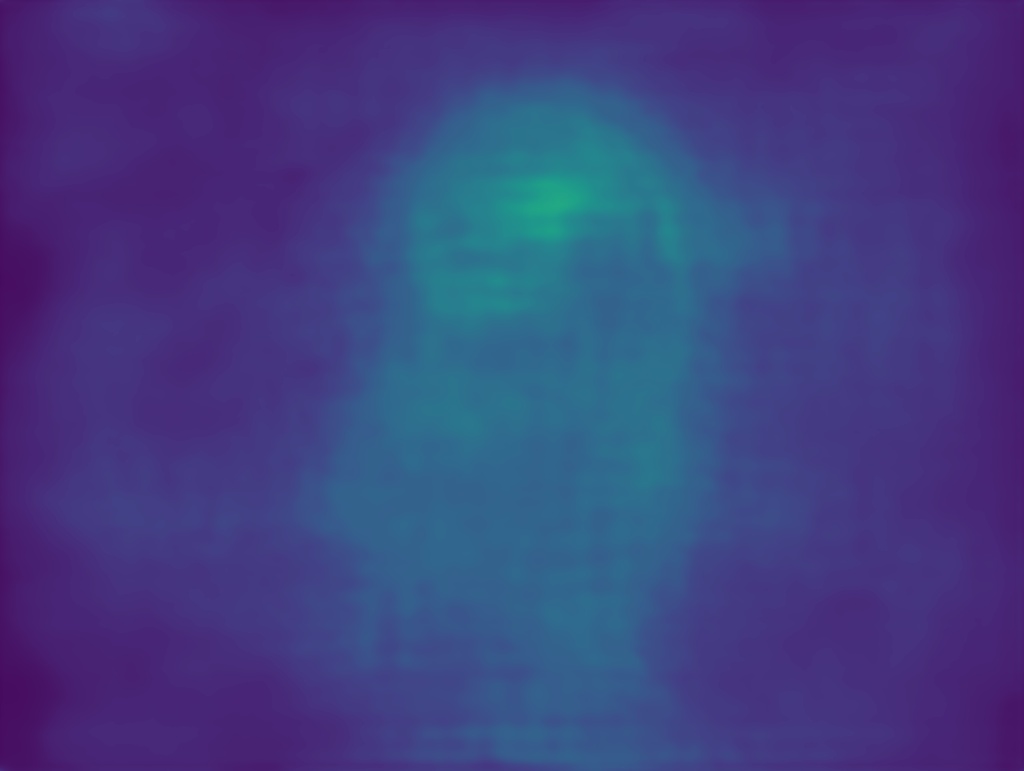} &
        \includegraphics[width=0.105\textwidth]{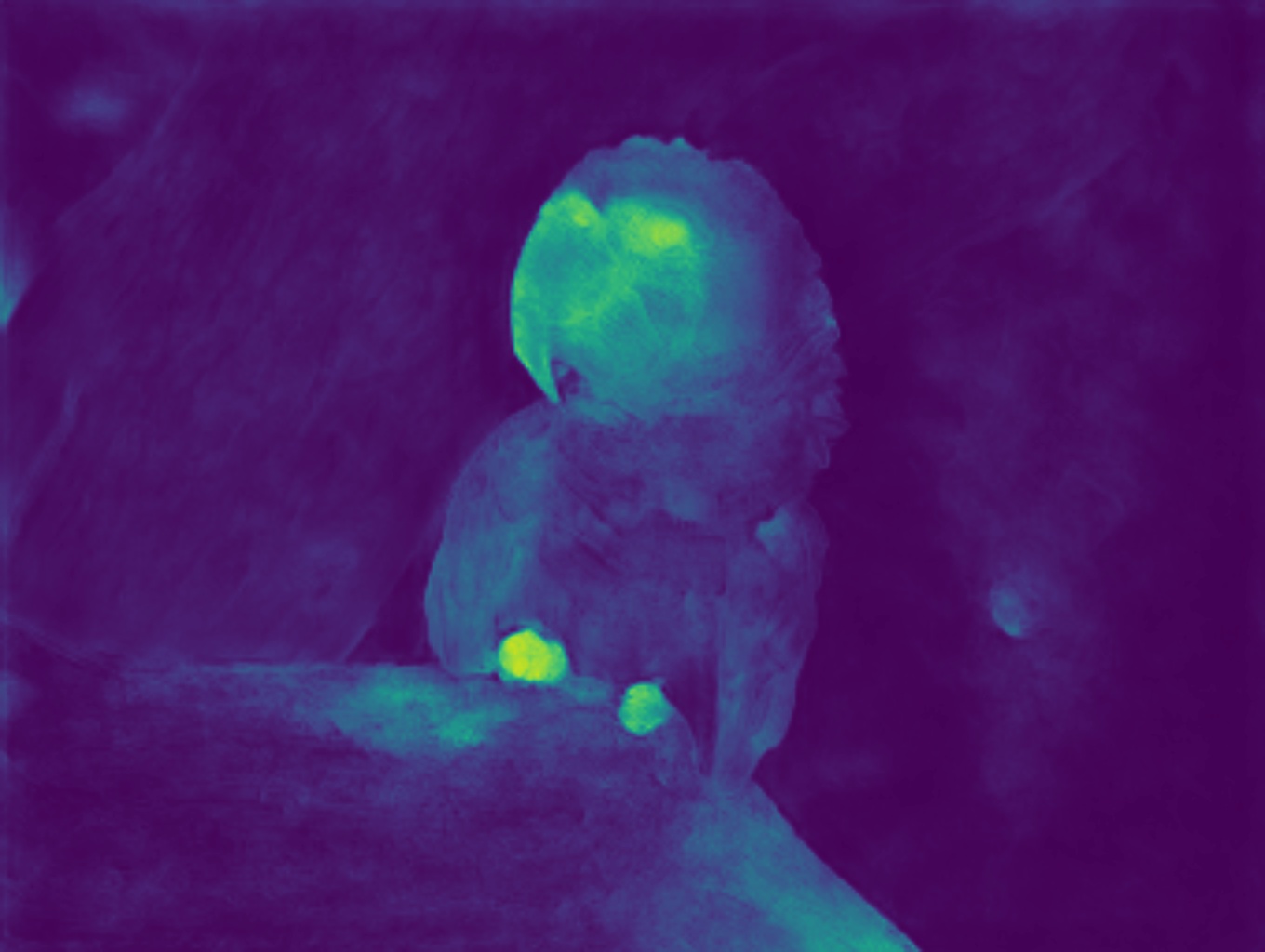} &
        \includegraphics[width=0.105\textwidth]{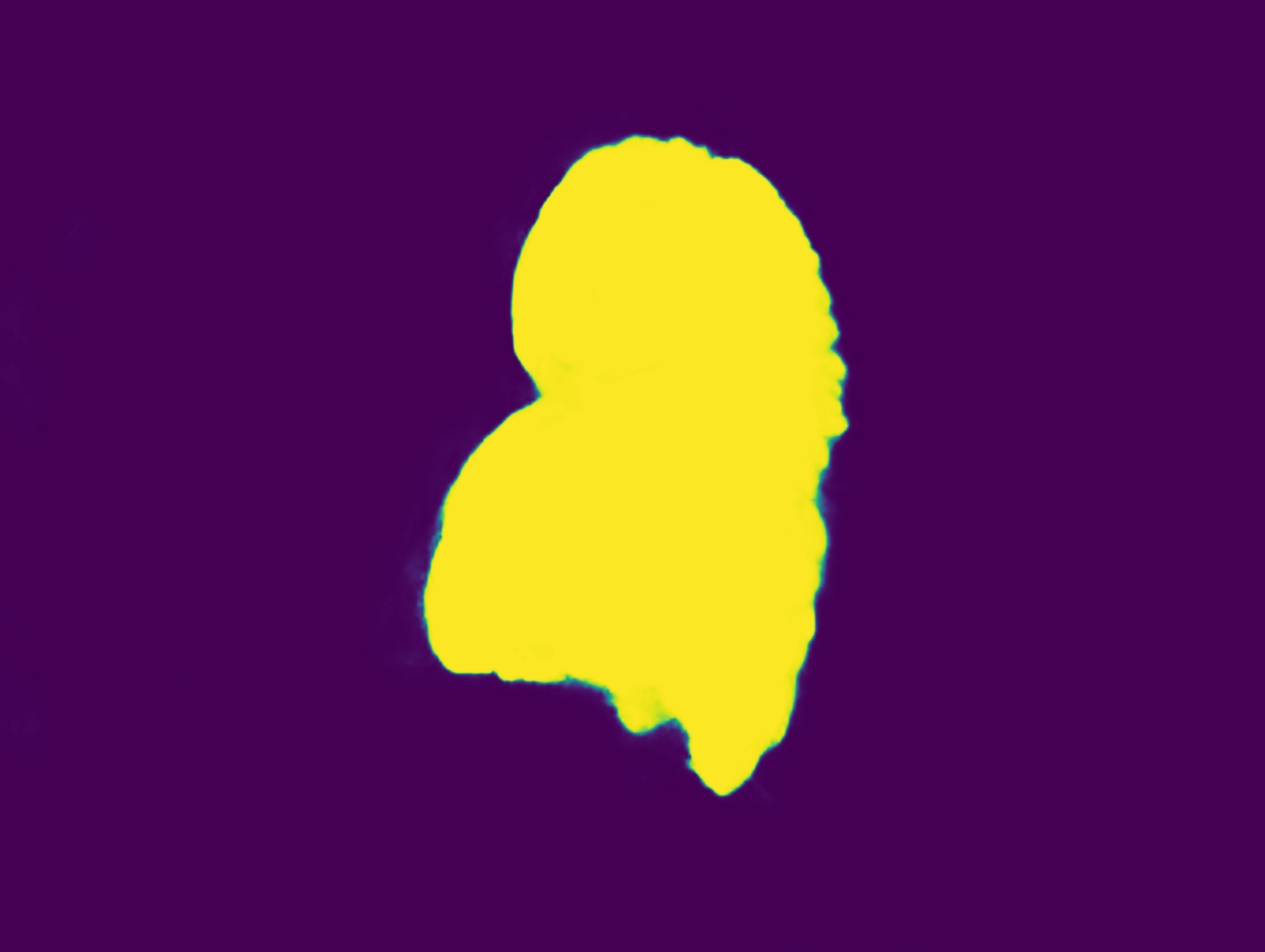} \\

        \includegraphics[width=0.105\textwidth]{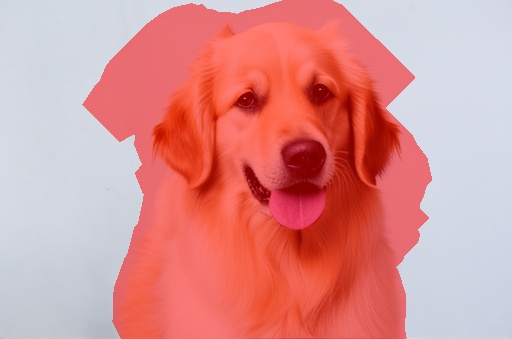} &
        \includegraphics[width=0.105\textwidth]{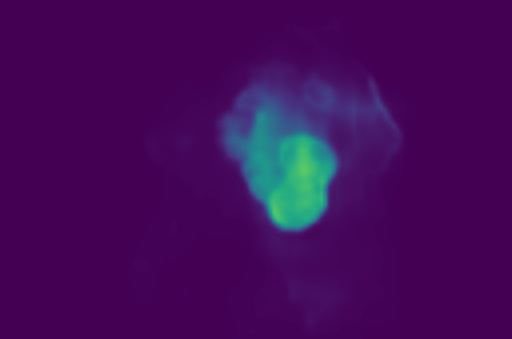} &
        \includegraphics[width=0.105\textwidth]{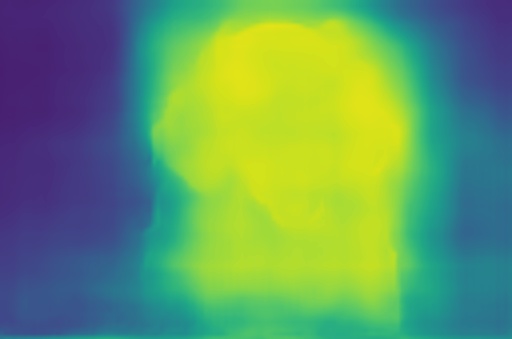} &
        \includegraphics[width=0.105\textwidth]{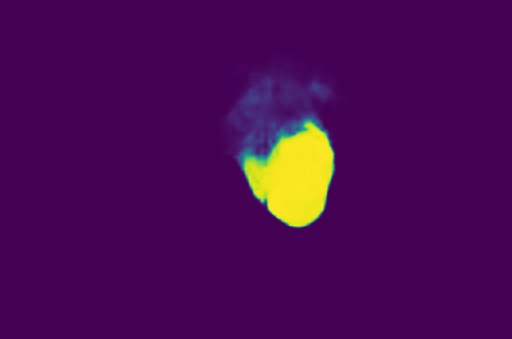} &
        \includegraphics[width=0.105\textwidth]{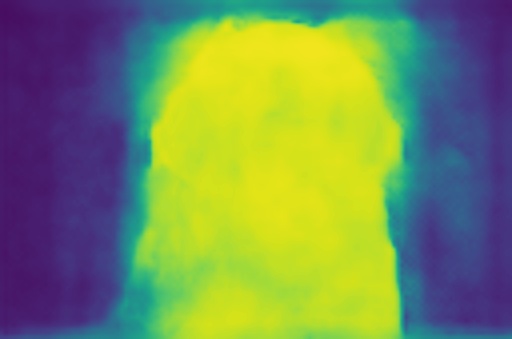} &
        \includegraphics[width=0.105\textwidth]{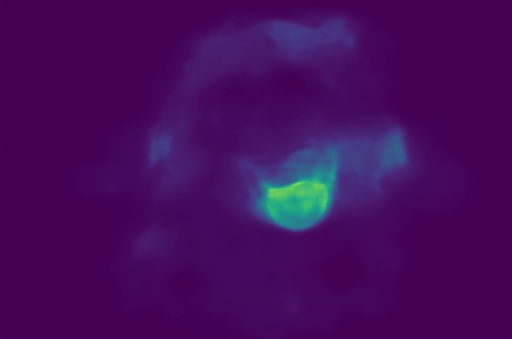} &
        \includegraphics[width=0.105\textwidth]{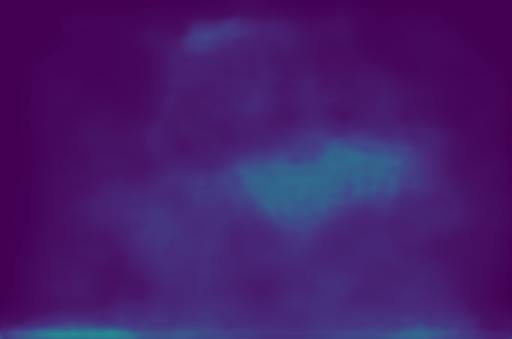} &
        \includegraphics[width=0.105\textwidth]{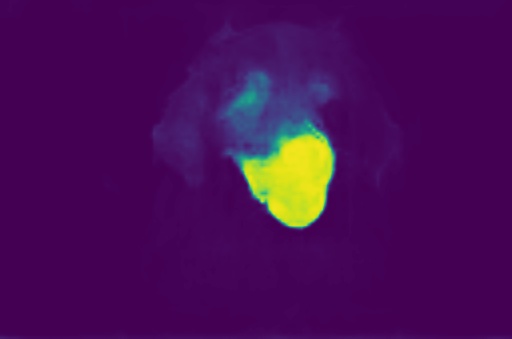} &
        \includegraphics[width=0.105\textwidth]{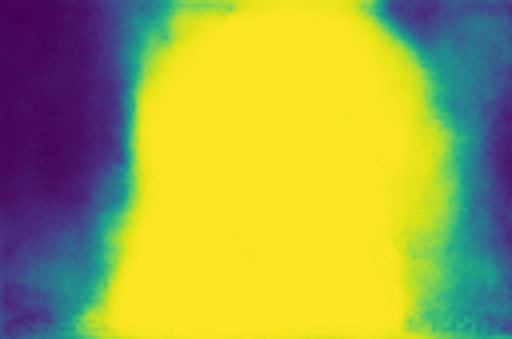} \\

        \includegraphics[width=0.105\textwidth]{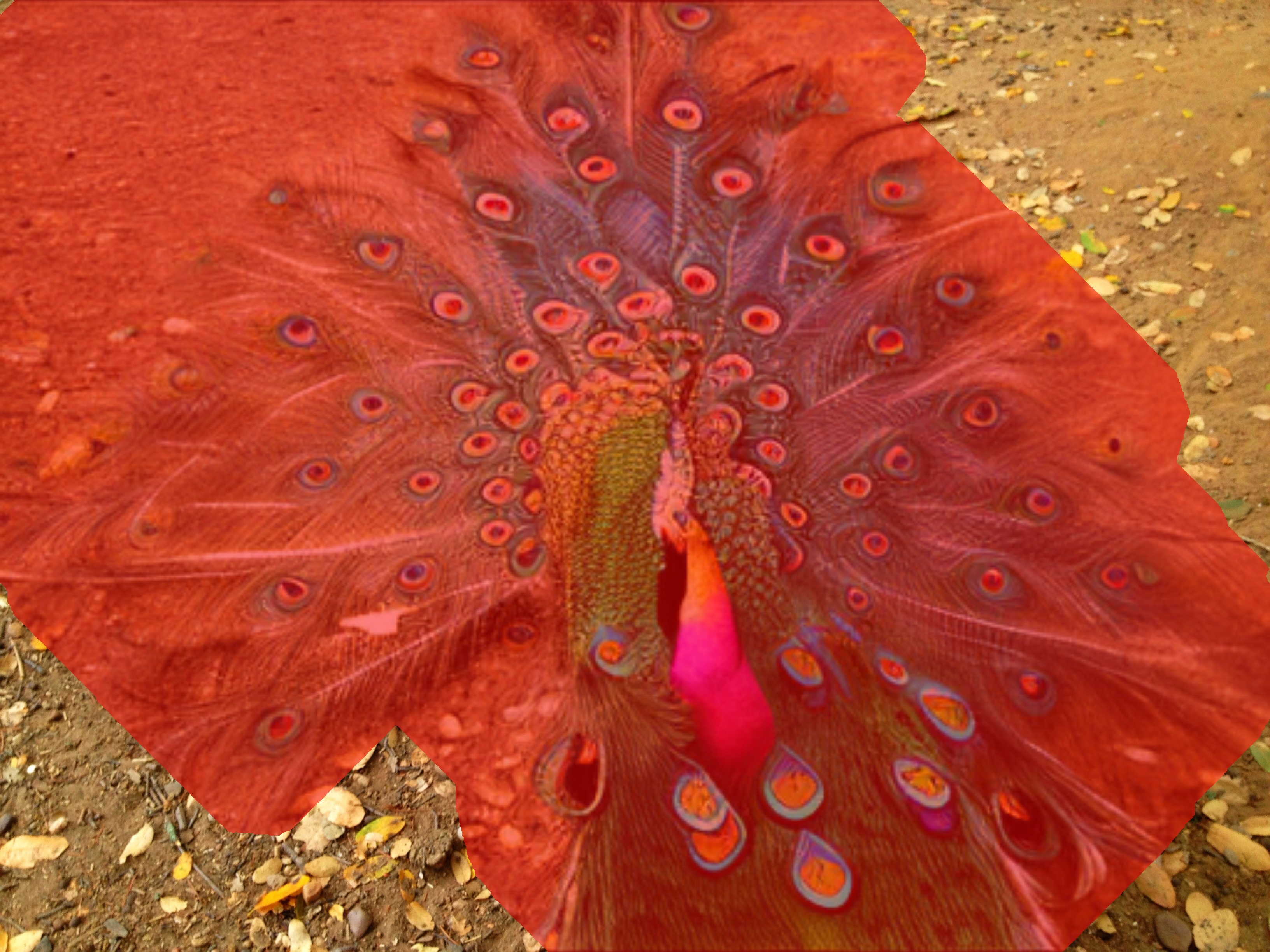} &
        \includegraphics[width=0.105\textwidth]{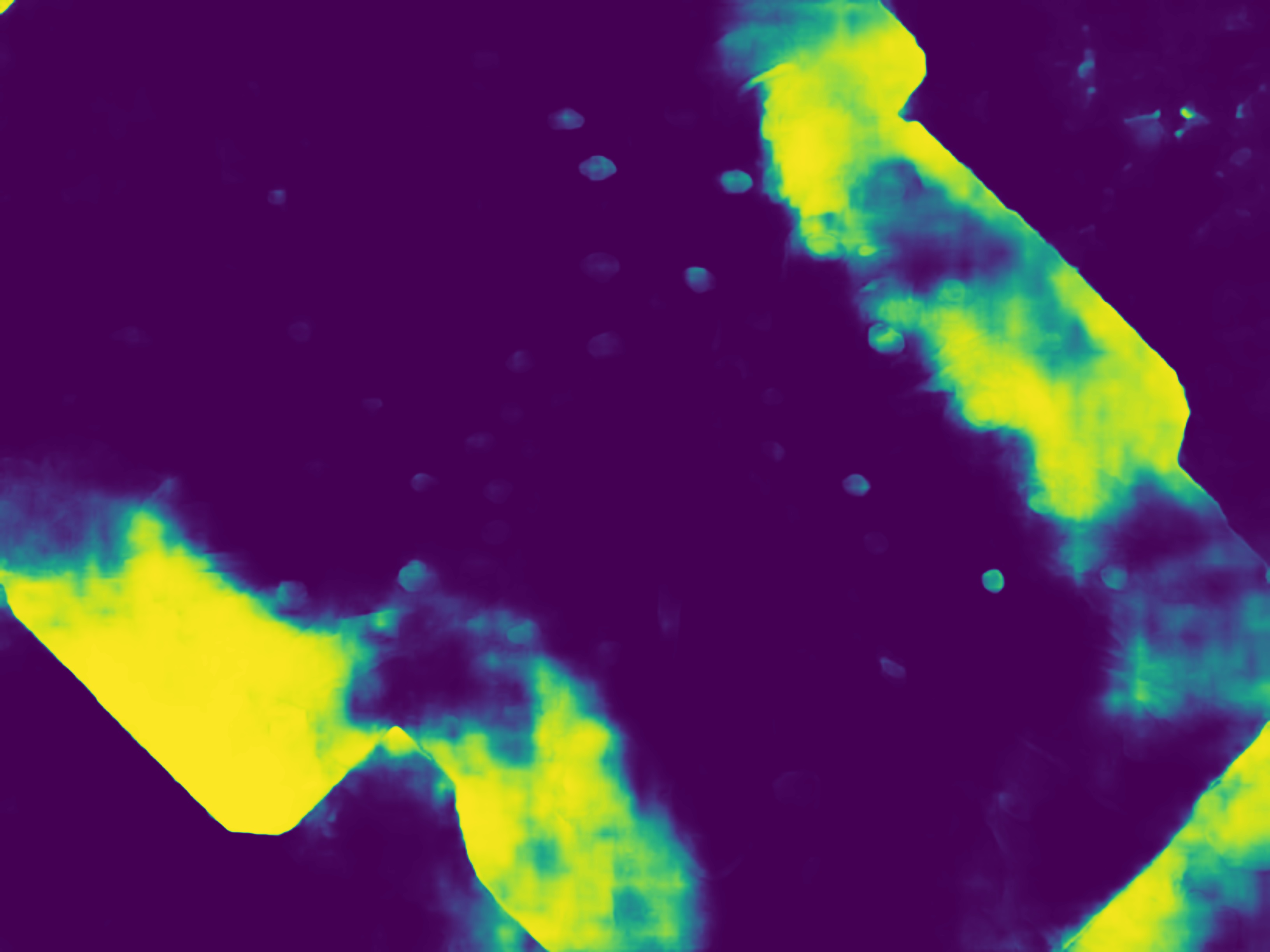} &
        \includegraphics[width=0.105\textwidth]{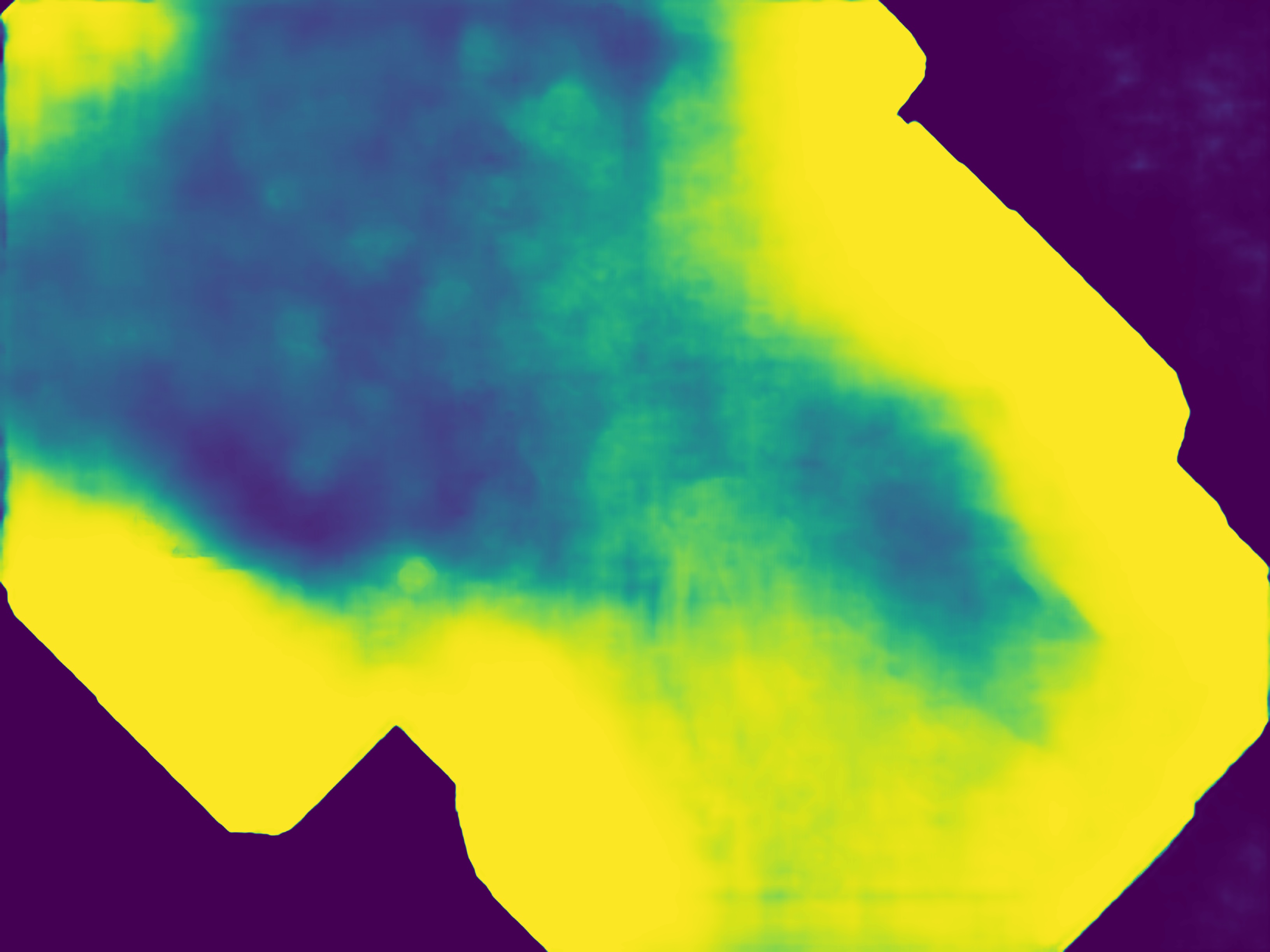} &
        \includegraphics[width=0.105\textwidth]{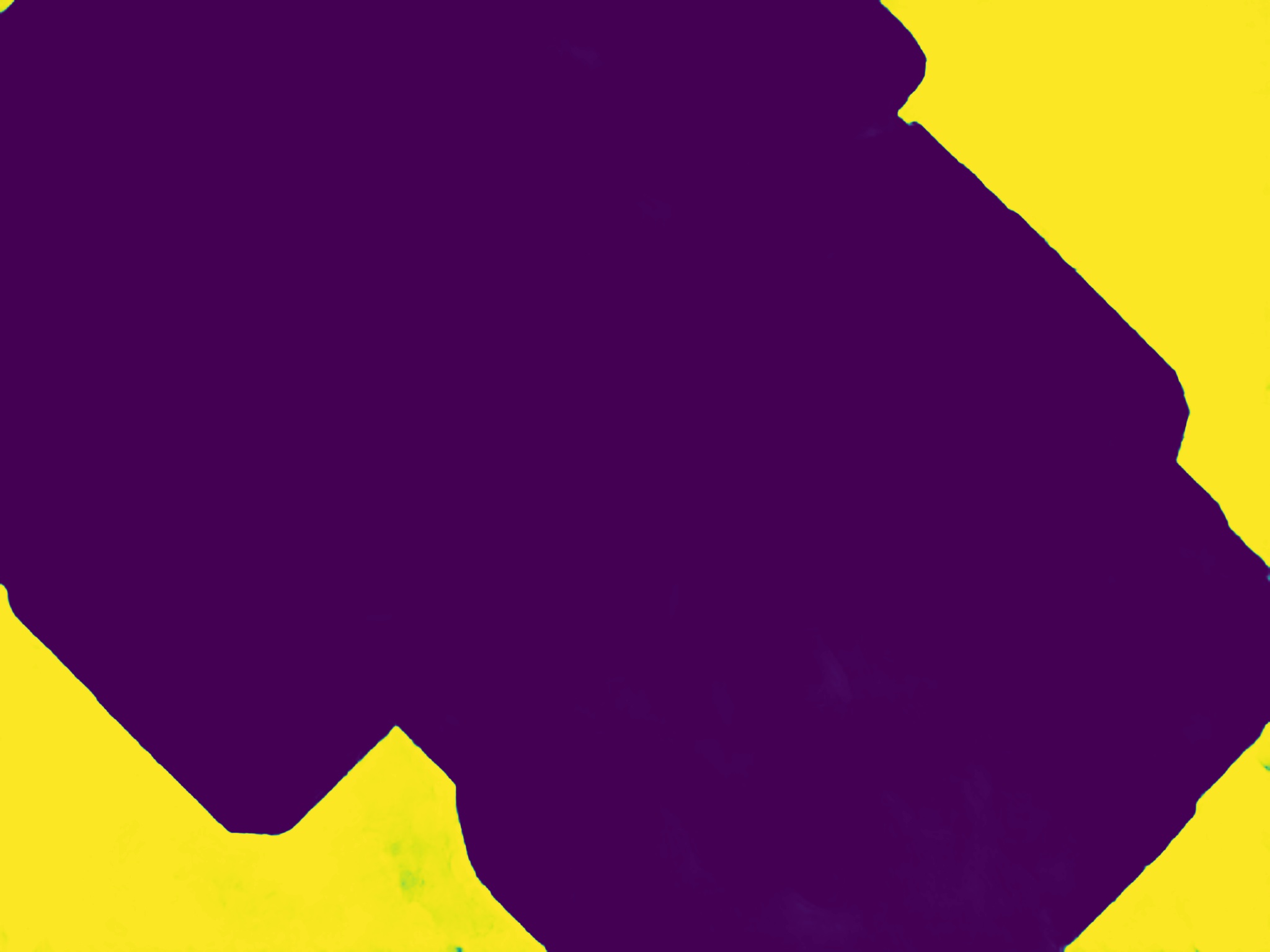} &
        \includegraphics[width=0.105\textwidth]{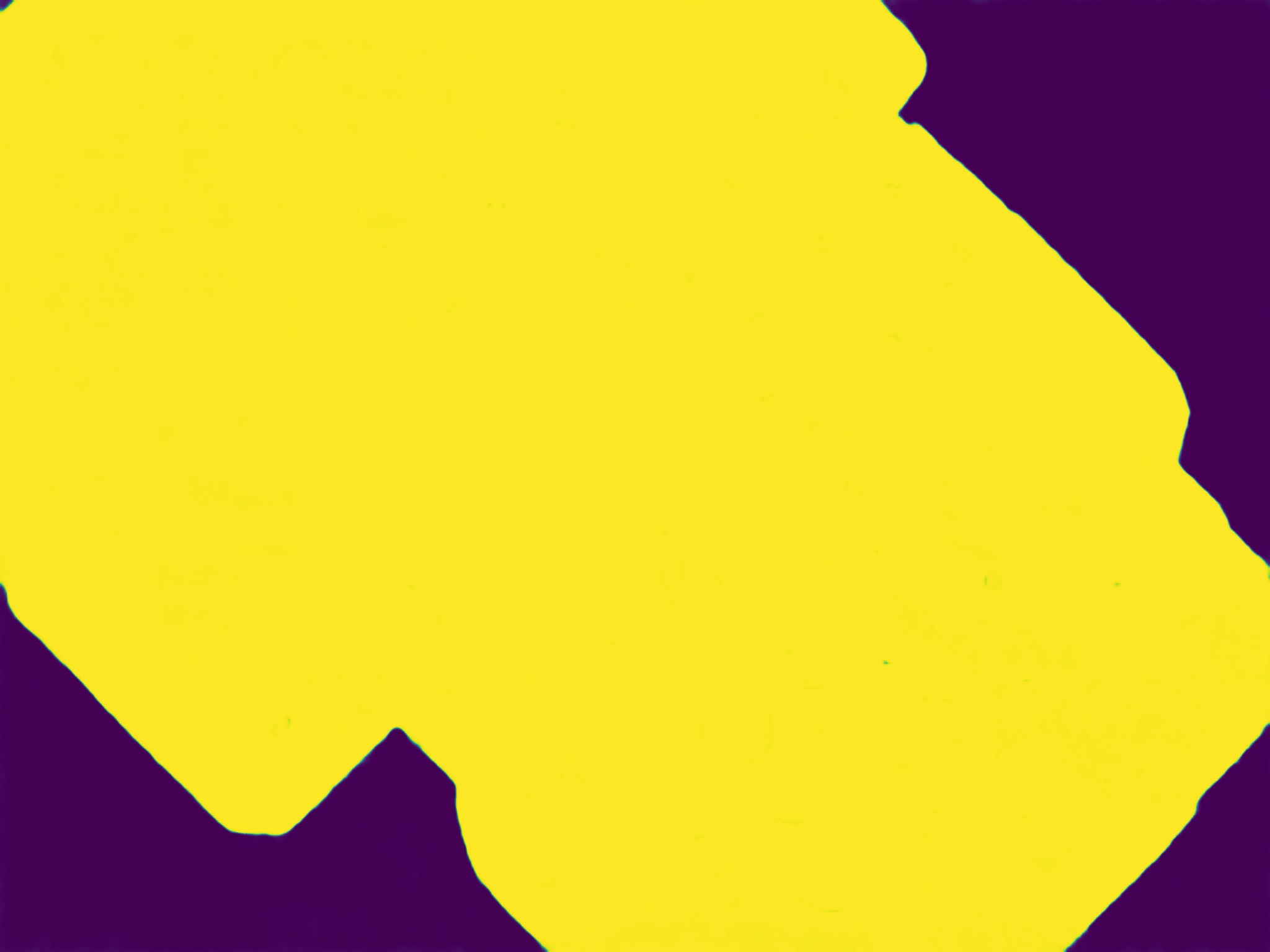} &
        \includegraphics[width=0.105\textwidth]{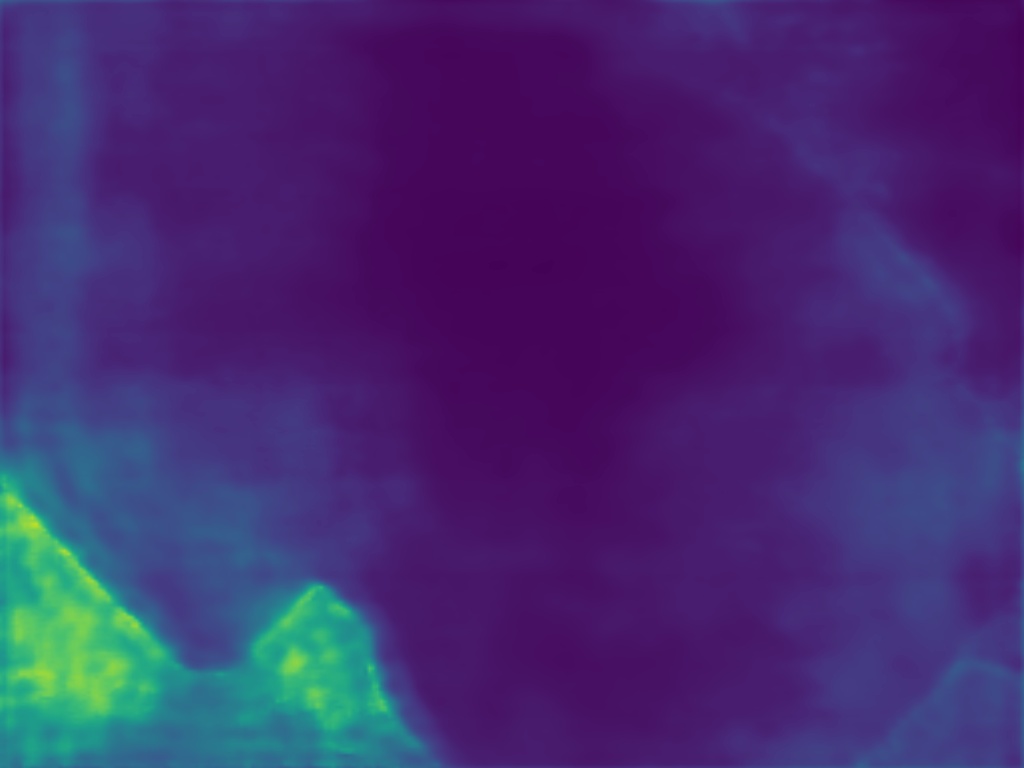} &
        \includegraphics[width=0.105\textwidth]{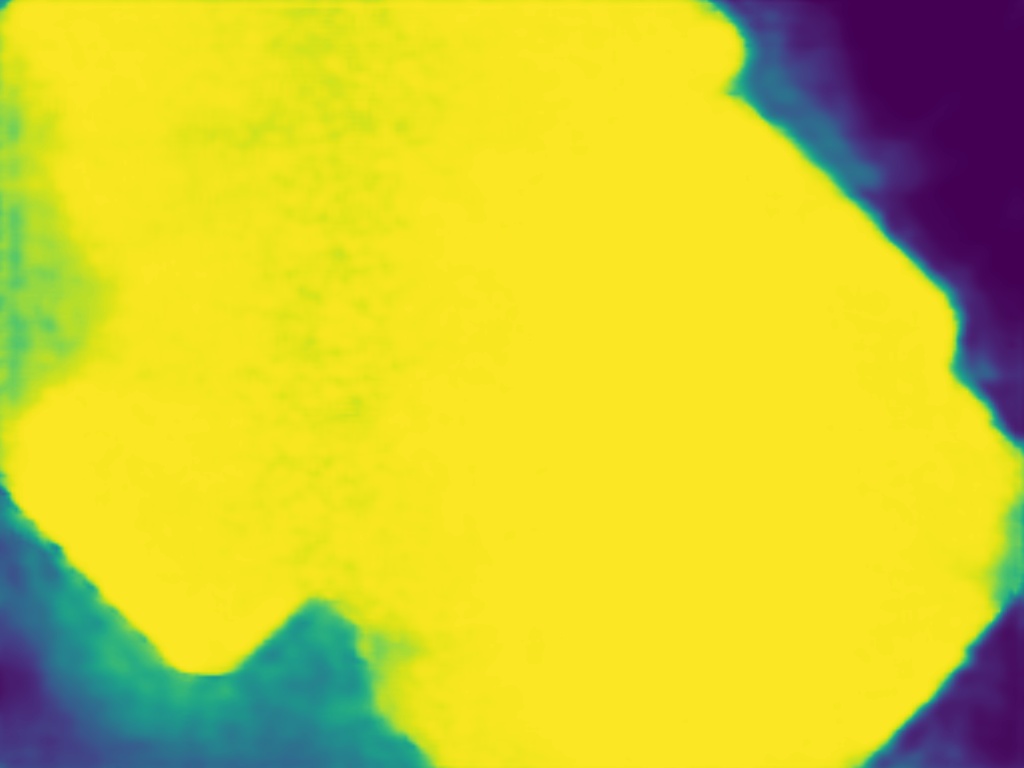} &
        \includegraphics[width=0.105\textwidth]{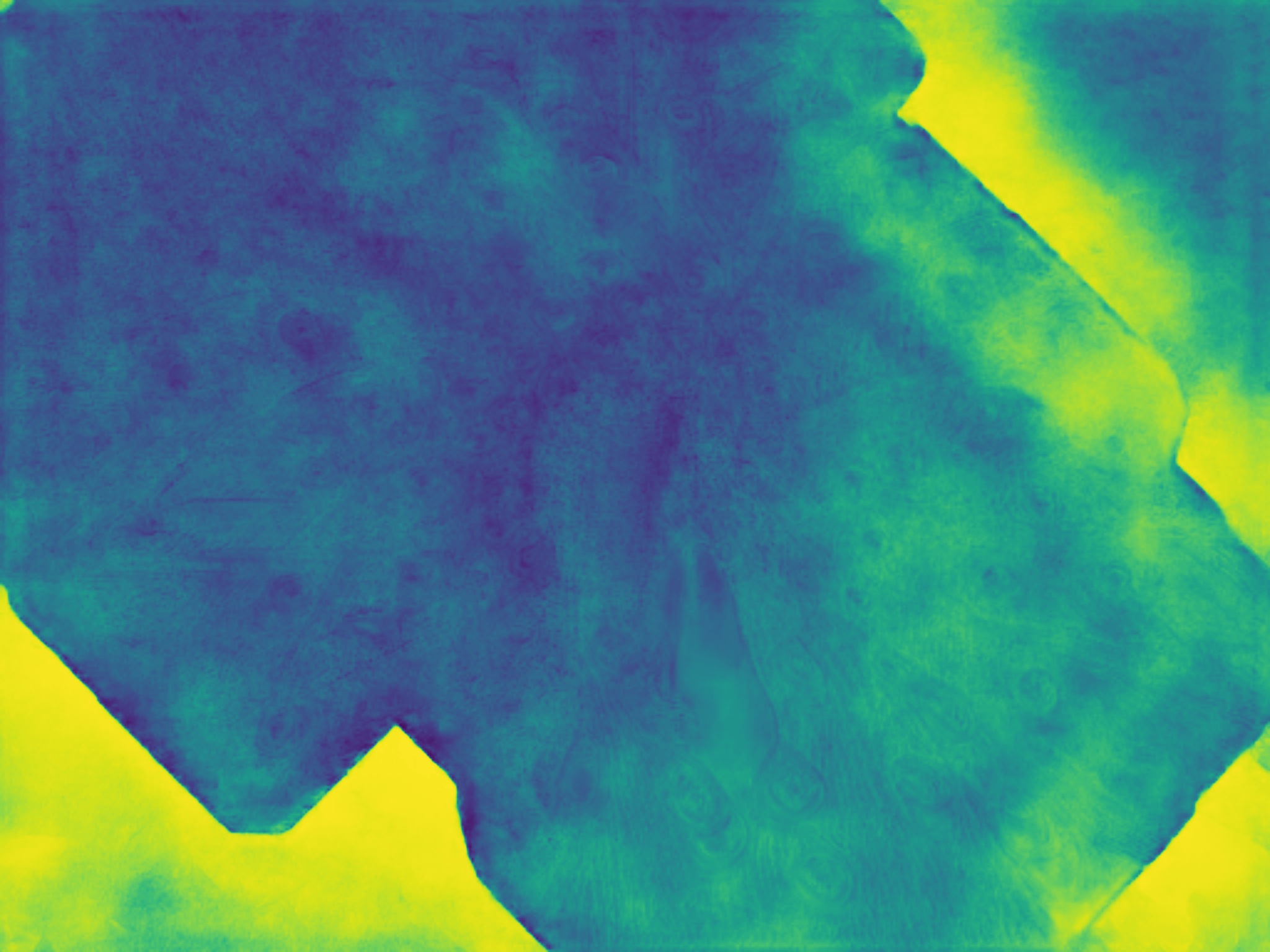} &
        \includegraphics[width=0.105\textwidth]{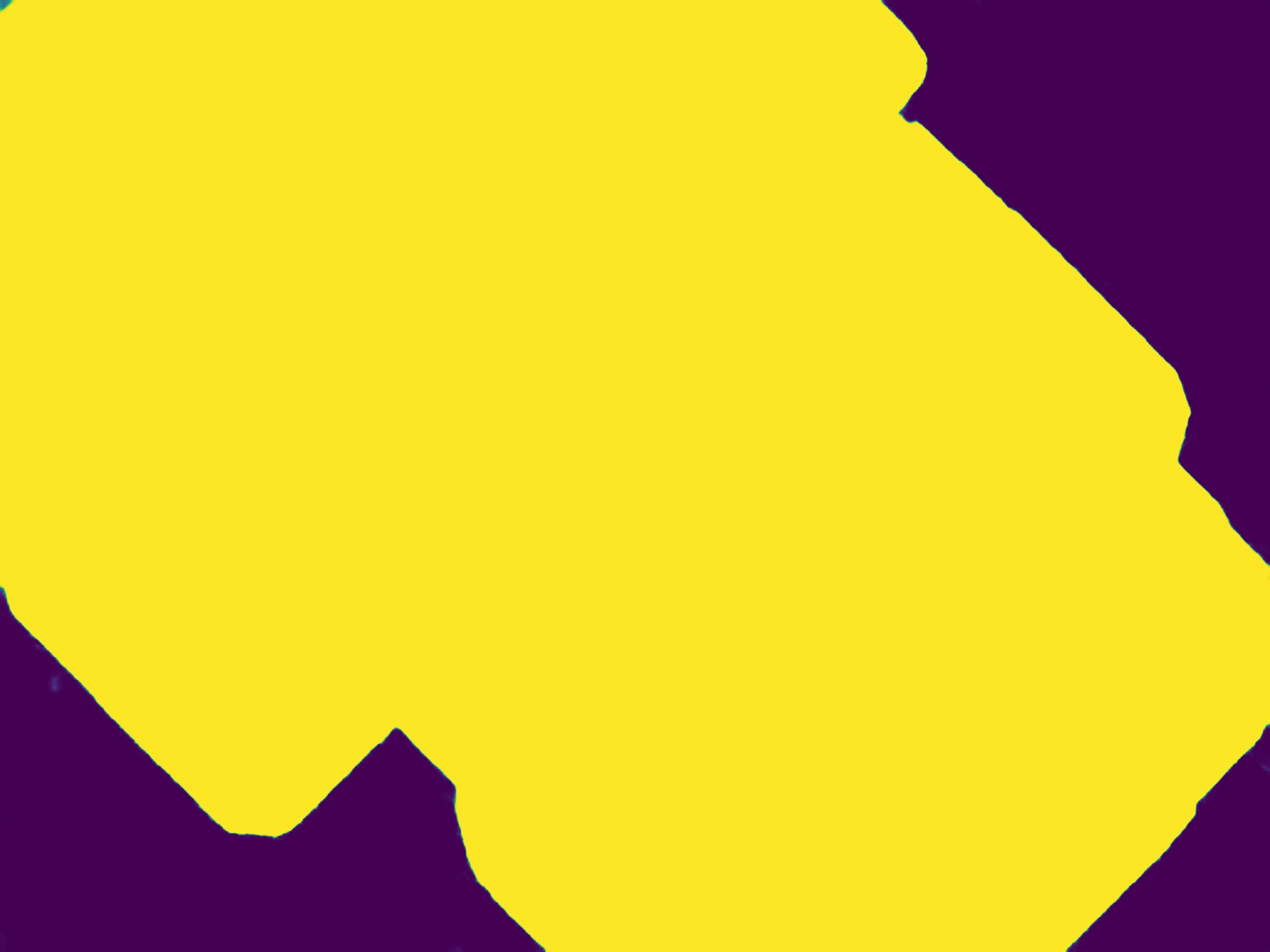} \\

        \includegraphics[width=0.105\textwidth]{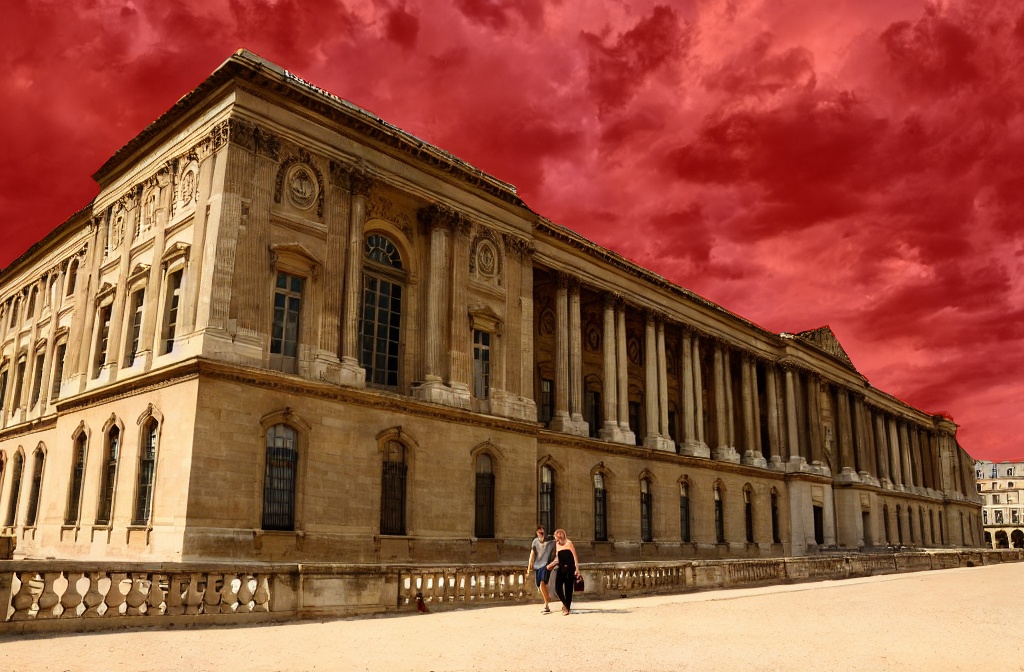} &
        \includegraphics[width=0.105\textwidth]{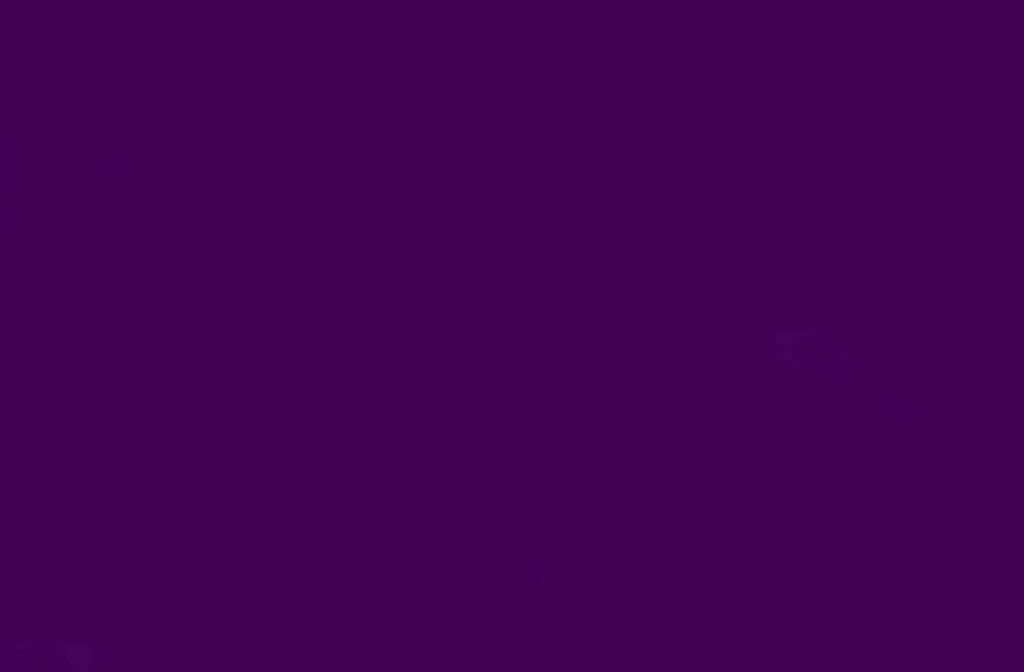} &
        \includegraphics[width=0.105\textwidth]{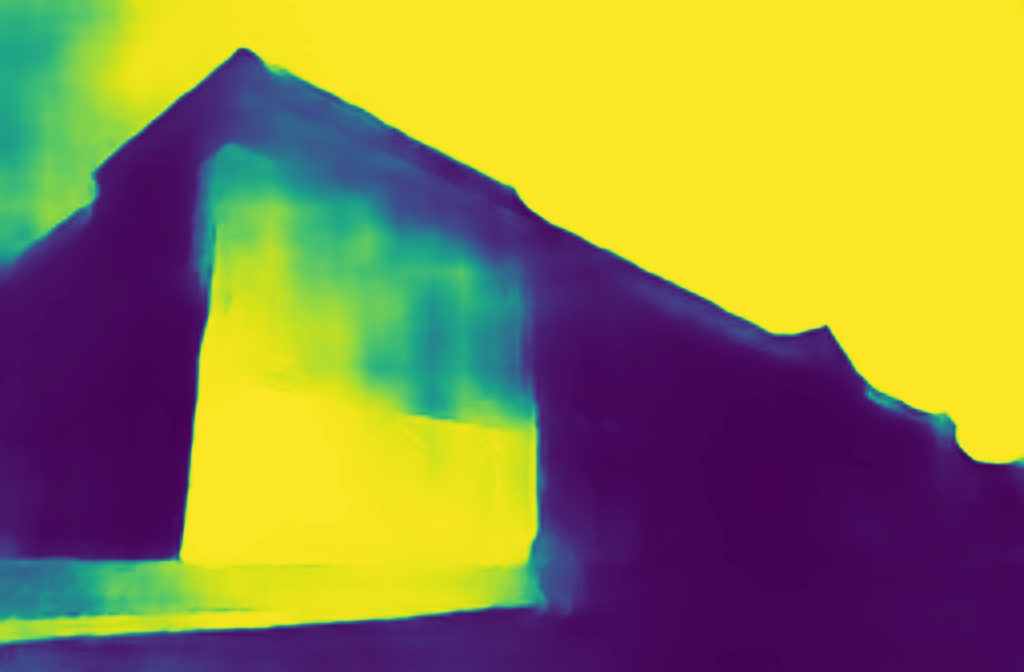} &
        \includegraphics[width=0.105\textwidth]{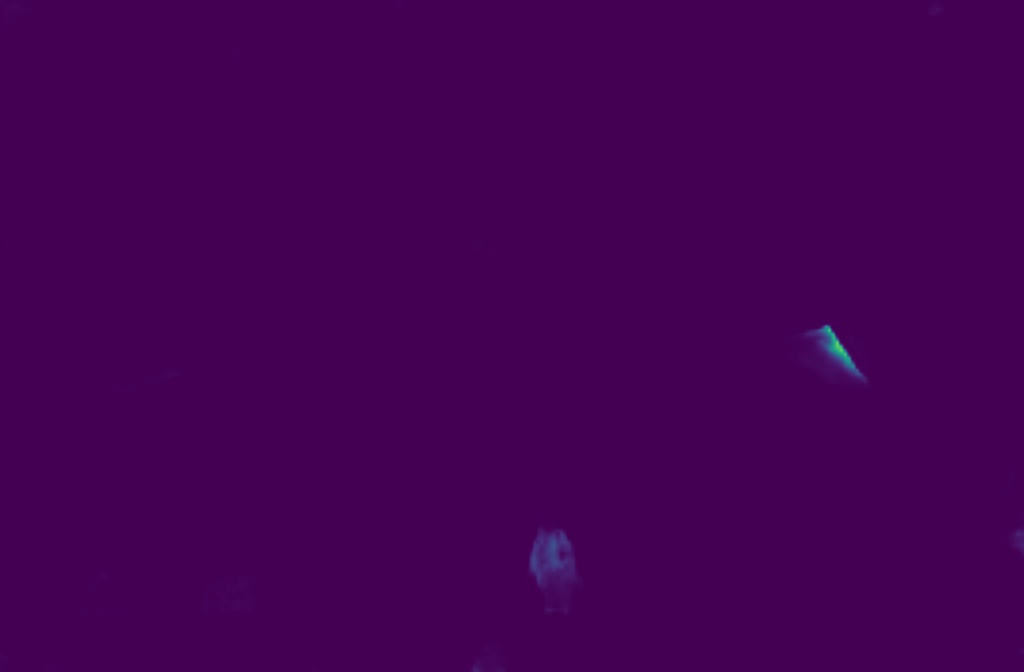} &
        \includegraphics[width=0.105\textwidth]{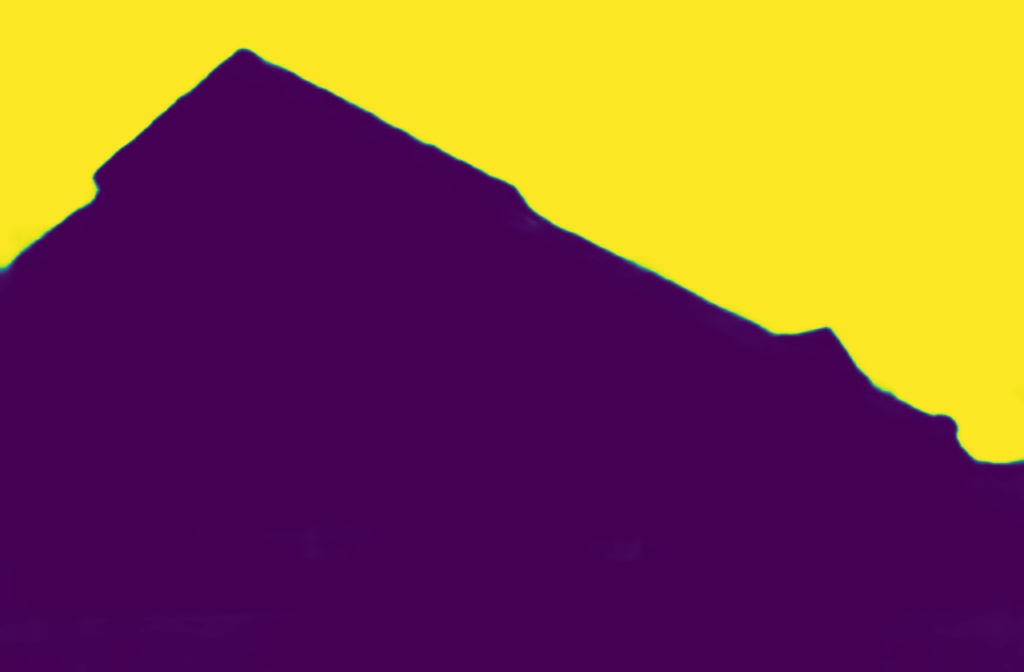} &
        \includegraphics[width=0.105\textwidth]{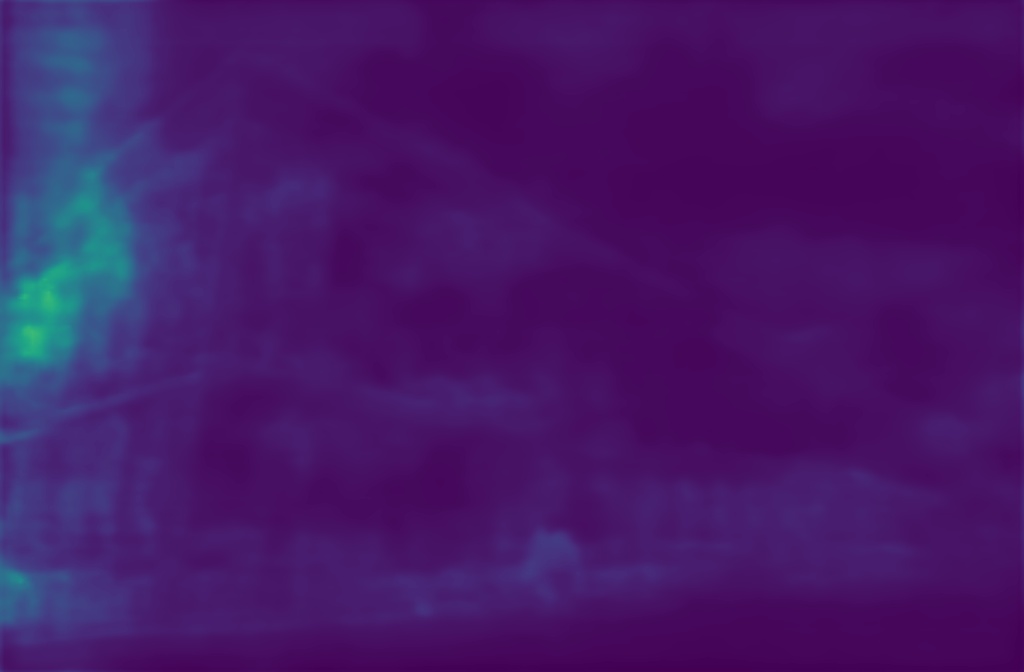} &
        \includegraphics[width=0.105\textwidth]{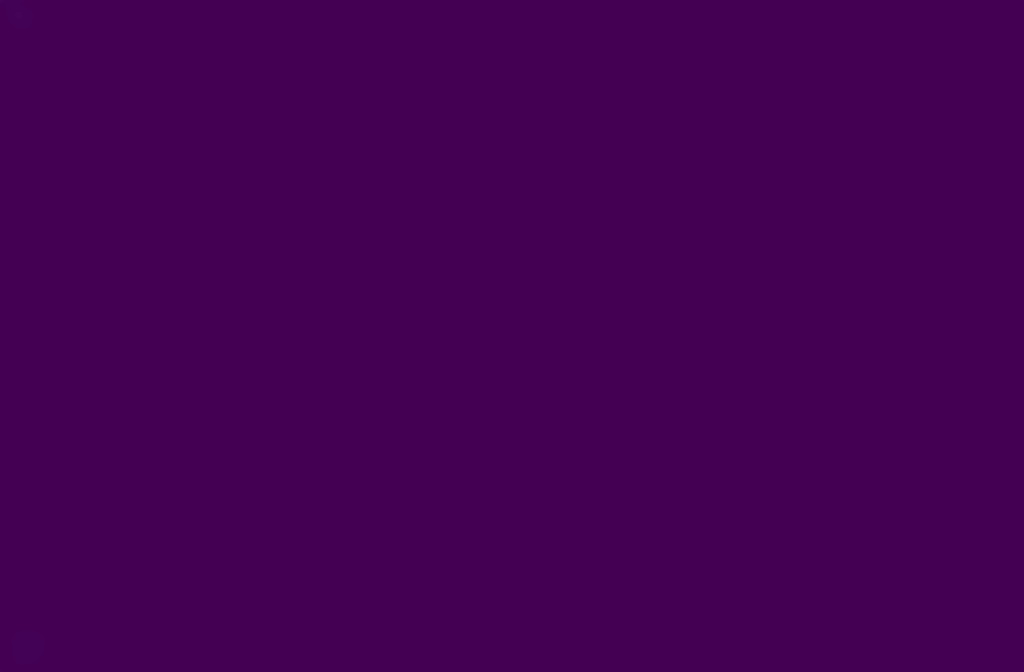} &
        \includegraphics[width=0.105\textwidth]{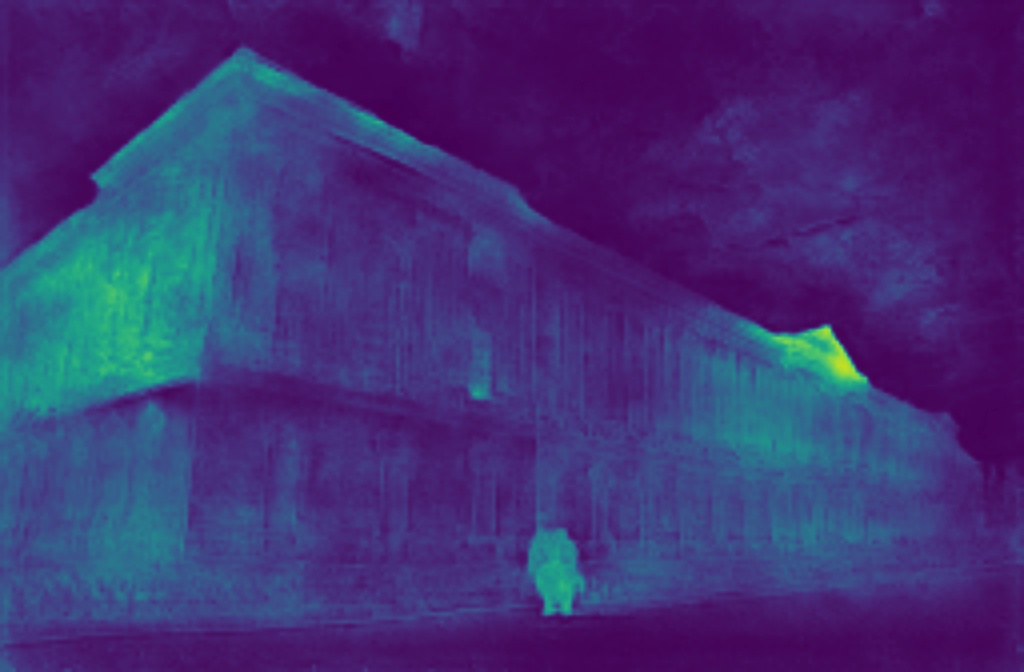} &
        \includegraphics[width=0.105\textwidth]{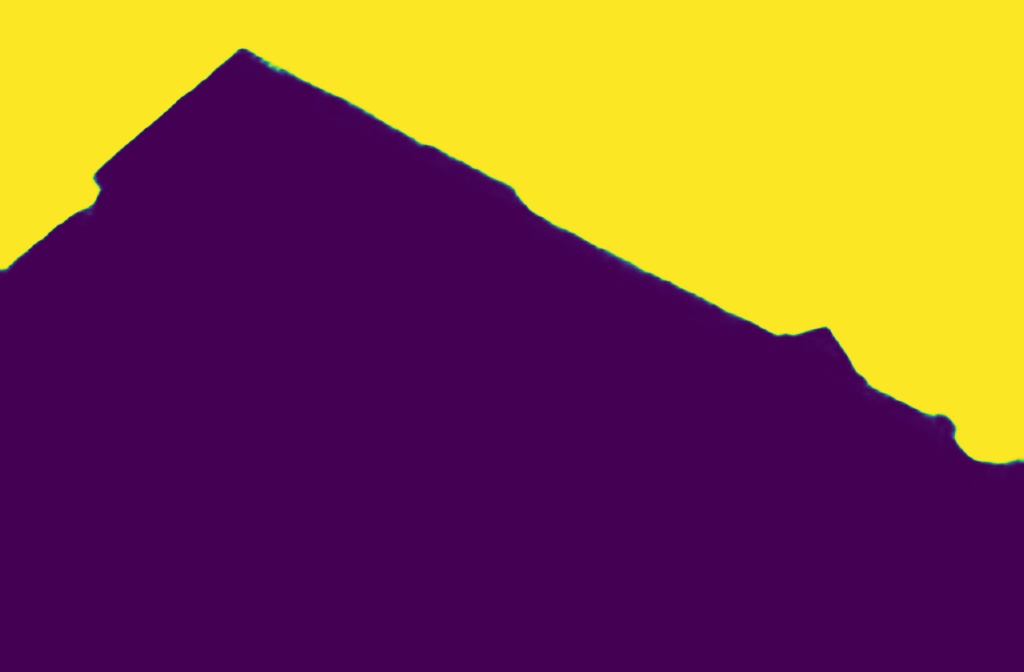}
    \end{tabular}
    \caption{Comparison of forgery localization results. For each row, from left to right: inpainted image, followed by localization maps from CatNet, MMFusion, PSCC, and TruFor models, showing both original (Orig) and retrained (Retr) versions.}
    \label{fig:loc_comparison_qual}
\end{figure*}

In Figure \ref{fig:loc_comparison_qual}, we present a comparison of the localization maps before and after fine-tuning. The results demonstrate that fine-tuning can significantly improve localization performance. For PSCC-Net, while improvements are observed in the second and fifth rows, poor localization results persist in other cases. Regarding the remaining models, localization improvements are evident in all cases, with TruFor consistently demonstrating the most accurate localization maps. The second row showcases an example where the original CAT-Net, MMFusion, and TruFor successfully identified the inpainted area, while the fourth row presents a case where the original model could only partially detect the inpainted region.
The fifth row presents an interesting case involving the original MMFusion model. If the predicted mask were inverted, it would have successfully identified the inpainted area. This can be attributed to the fact that in splicing it can be ambiguous which area is spliced and which is original. In AI inpainting cases, however, there is no ambiguity about which region has been modified.

\section{Example Outputs}

\begin{figure*}[p]
    \centering
    
    \begin{subfigure}{\textwidth}
        \begin{subfigure}{0.32\textwidth}
            \includegraphics[width=0.49\linewidth]{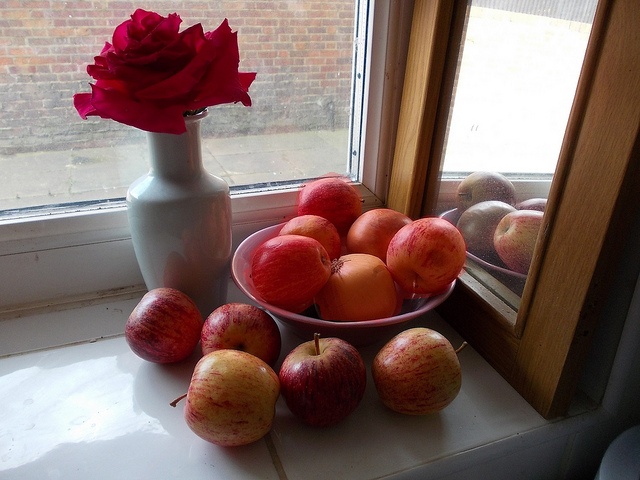}
            \includegraphics[width=0.49\linewidth]{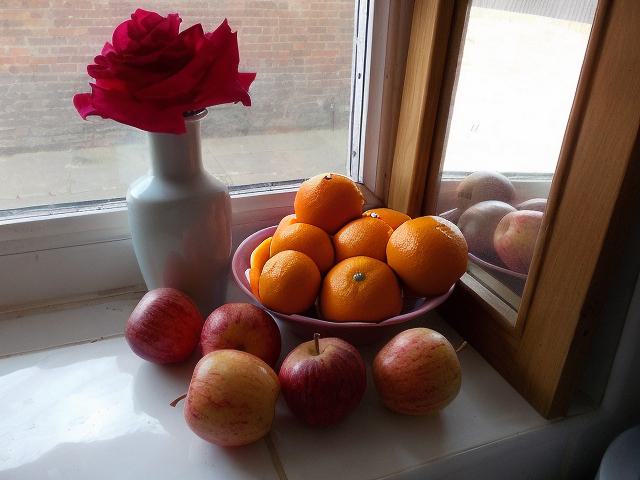}
            \caption{``a juicy orange to add a vibrant pop of color to the composition''}
        \end{subfigure}
        \hfill
        \begin{subfigure}{0.32\textwidth}
            \includegraphics[width=0.49\linewidth]{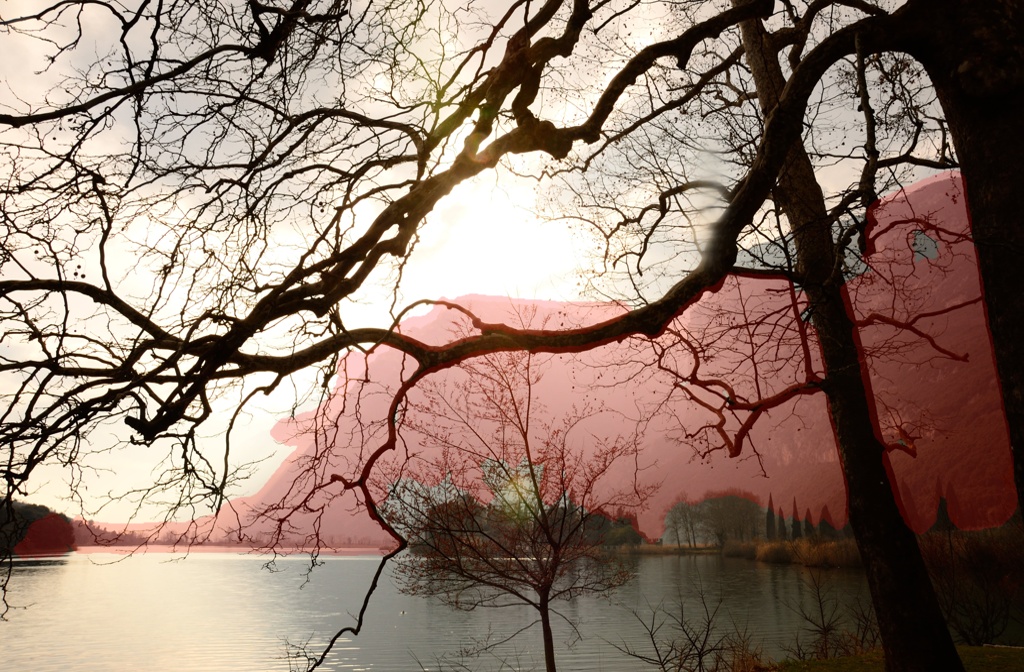}
            \includegraphics[width=0.49\linewidth]{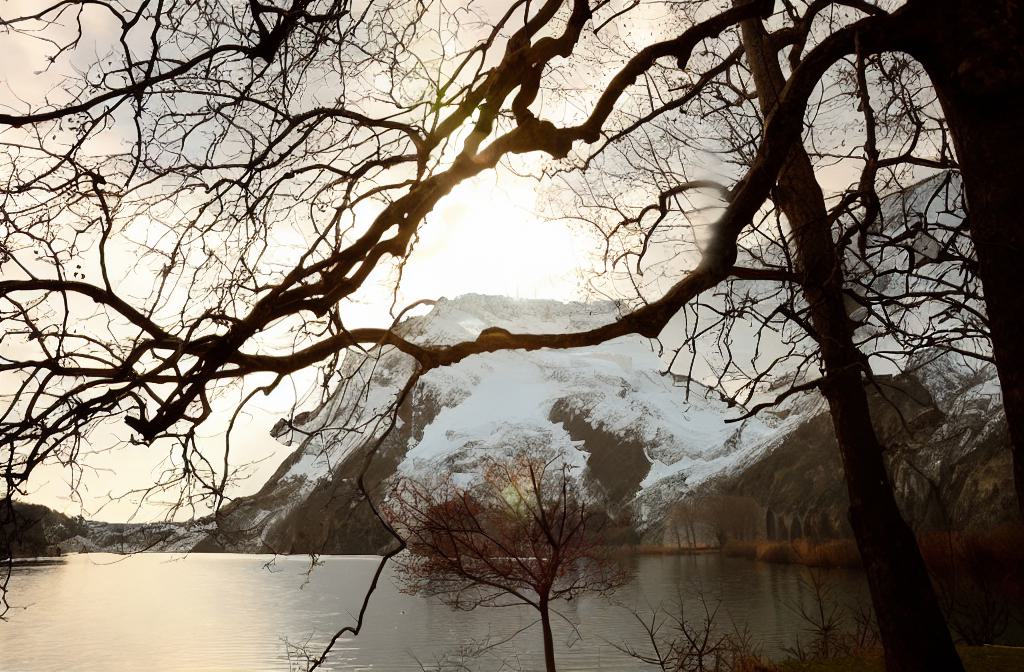}
            \caption{``a majestic snow-capped mountain to create a scenic landscape''}
        \end{subfigure}
        \hfill
        \begin{subfigure}{0.32\textwidth}
            \includegraphics[width=0.49\linewidth]{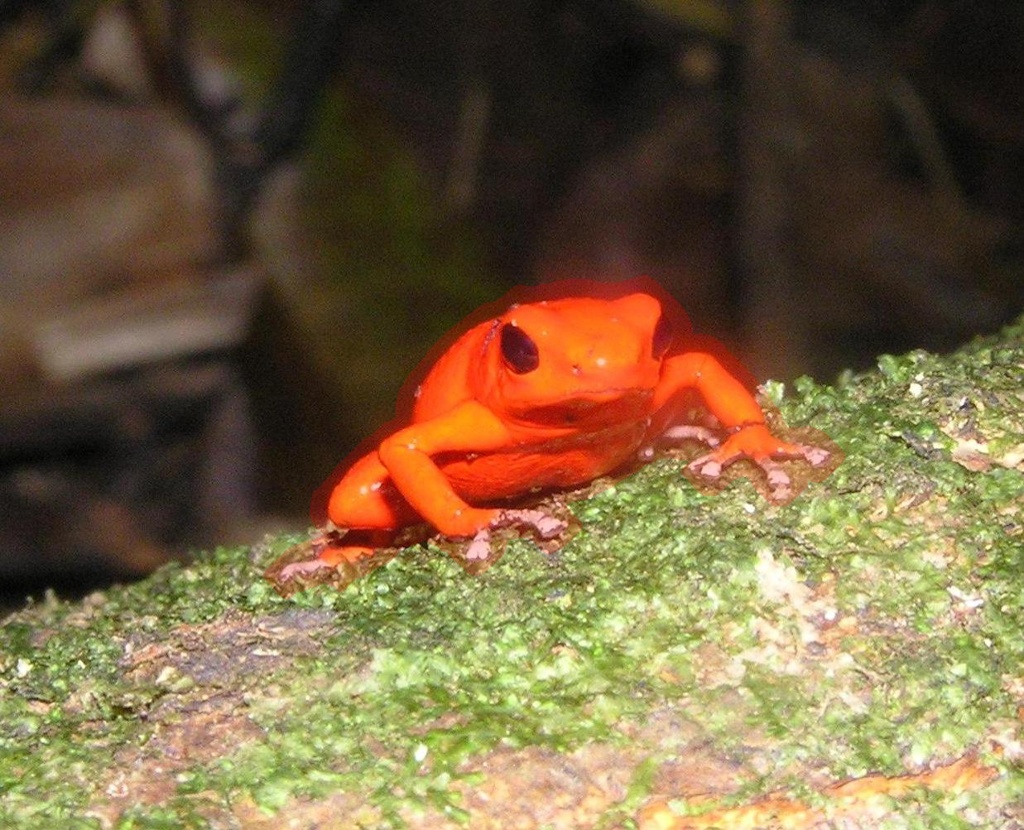}
            \includegraphics[width=0.49\linewidth]{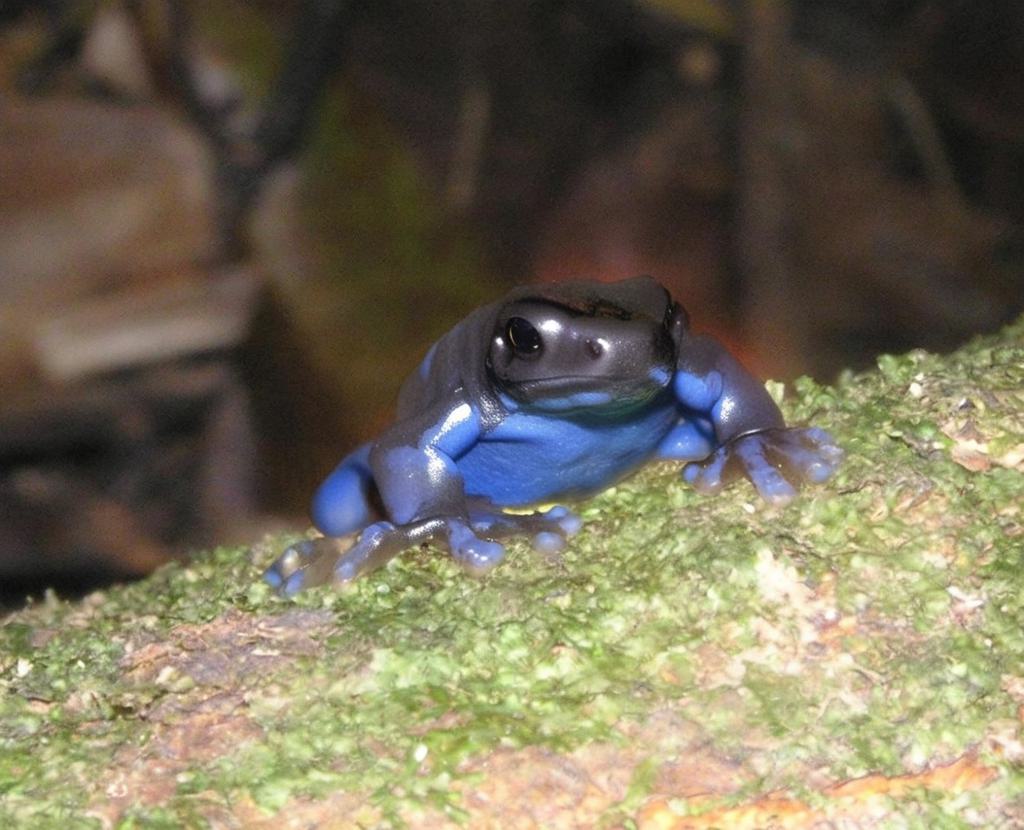}
            \caption{``a vibrant blue poison dart frog''}
        \end{subfigure}
    \end{subfigure}
    
    \vspace{1em}
    
    \begin{subfigure}{\textwidth}
        \begin{subfigure}{0.32\textwidth}
            \includegraphics[width=0.49\linewidth]{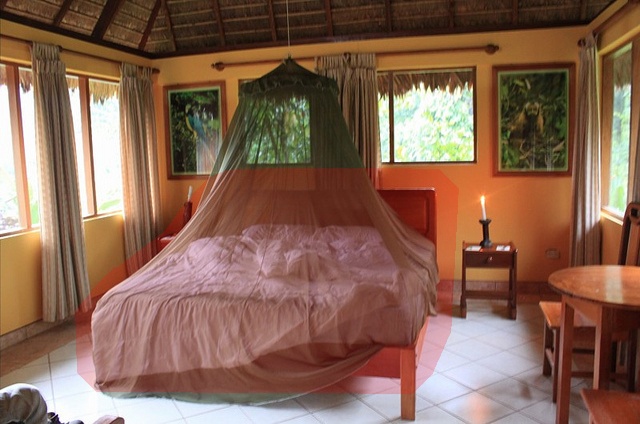}
            \includegraphics[width=0.49\linewidth]{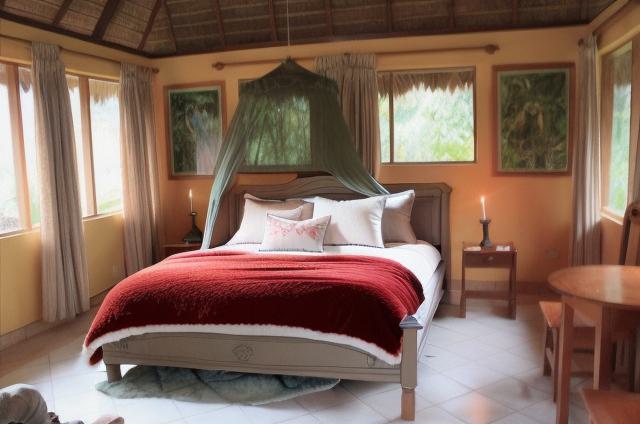}
            \caption{``a cozy blanket and fluffy pillows to complete the bedroom scene''}
        \end{subfigure}
        \hfill
        \begin{subfigure}{0.32\textwidth}
            \includegraphics[width=0.49\linewidth]{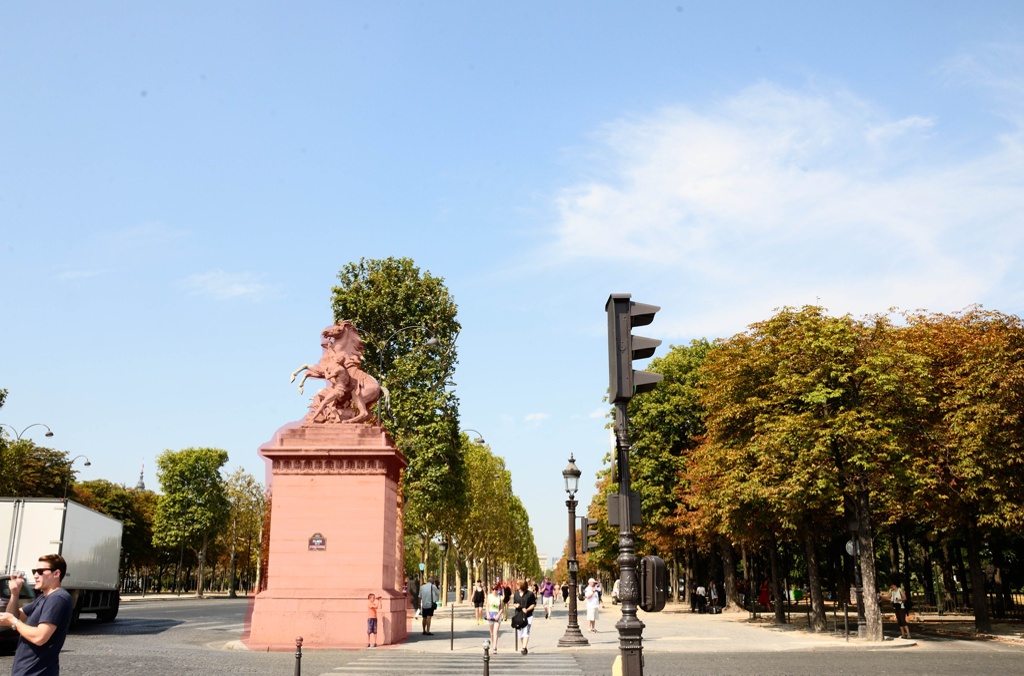}
            \includegraphics[width=0.49\linewidth]{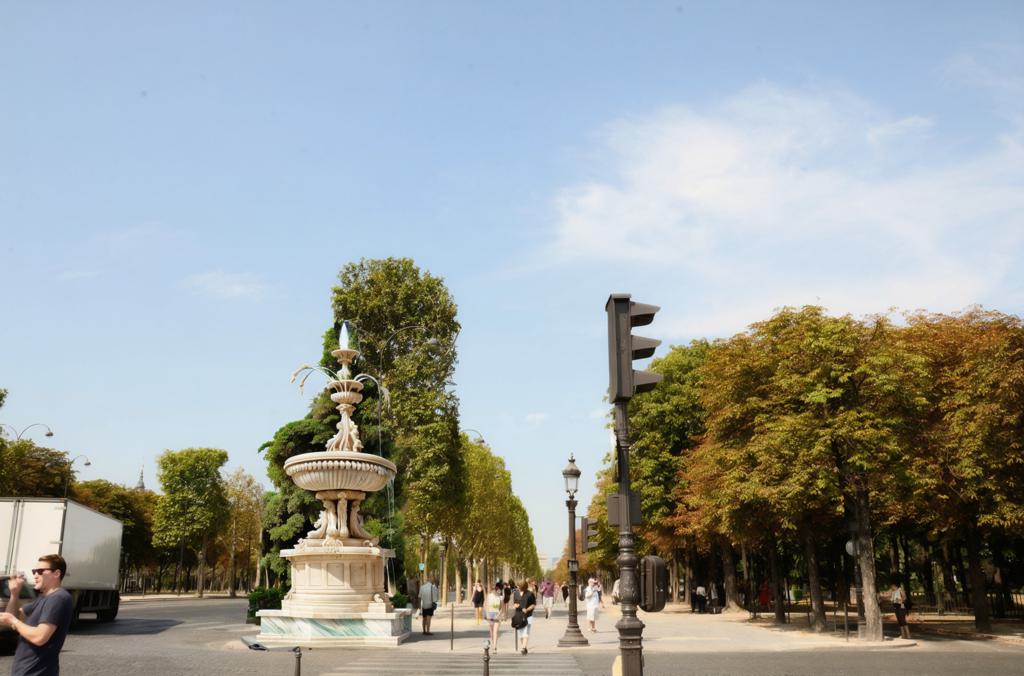}
            \caption{``a grand marble fountain surrounded by lush greenery''}
        \end{subfigure}
        \hfill
        \begin{subfigure}{0.32\textwidth}
            \includegraphics[width=0.49\linewidth]{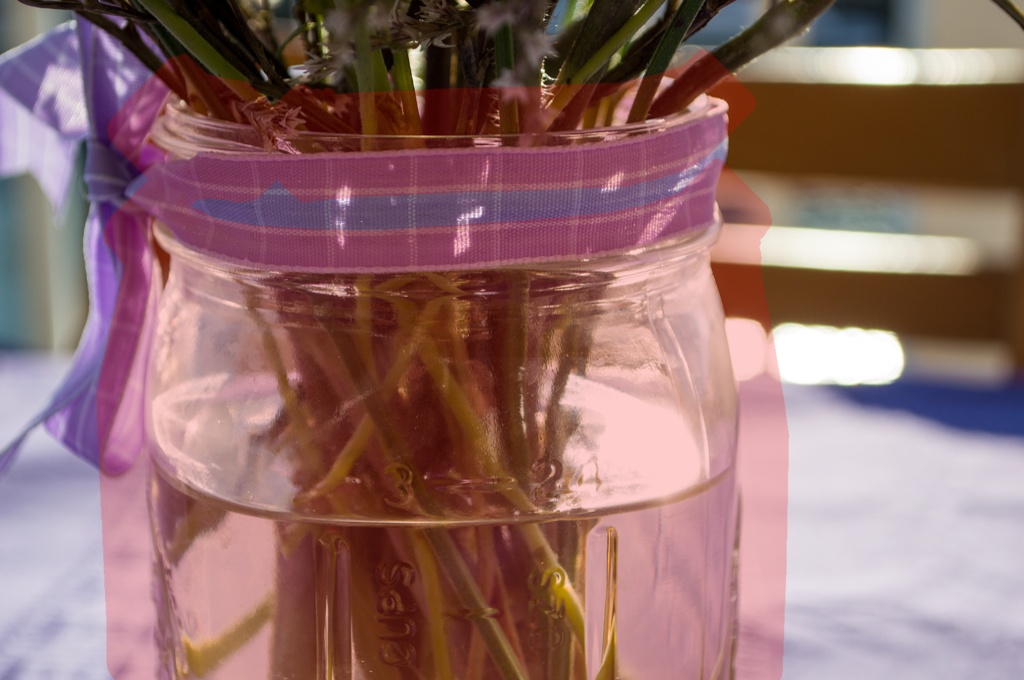}
            \includegraphics[width=0.49\linewidth]{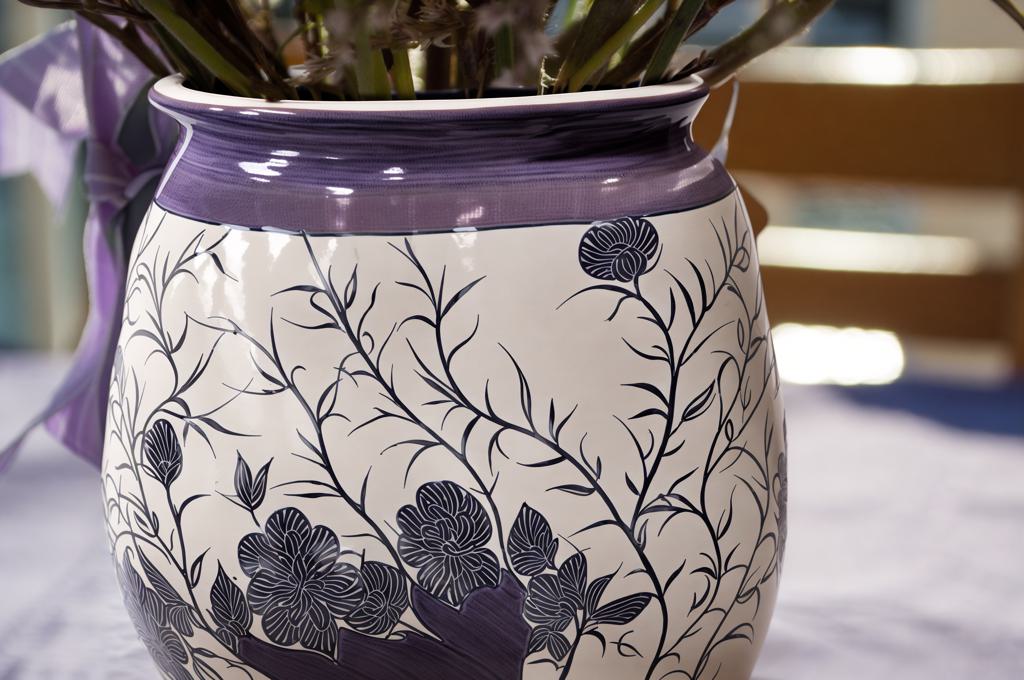}
            \caption{``a decorative ceramic vase''}
        \end{subfigure}
    \end{subfigure}
    
    \vspace{1em}
    
    \begin{subfigure}{\textwidth}
        \begin{subfigure}{0.32\textwidth}
            \includegraphics[width=0.49\linewidth]{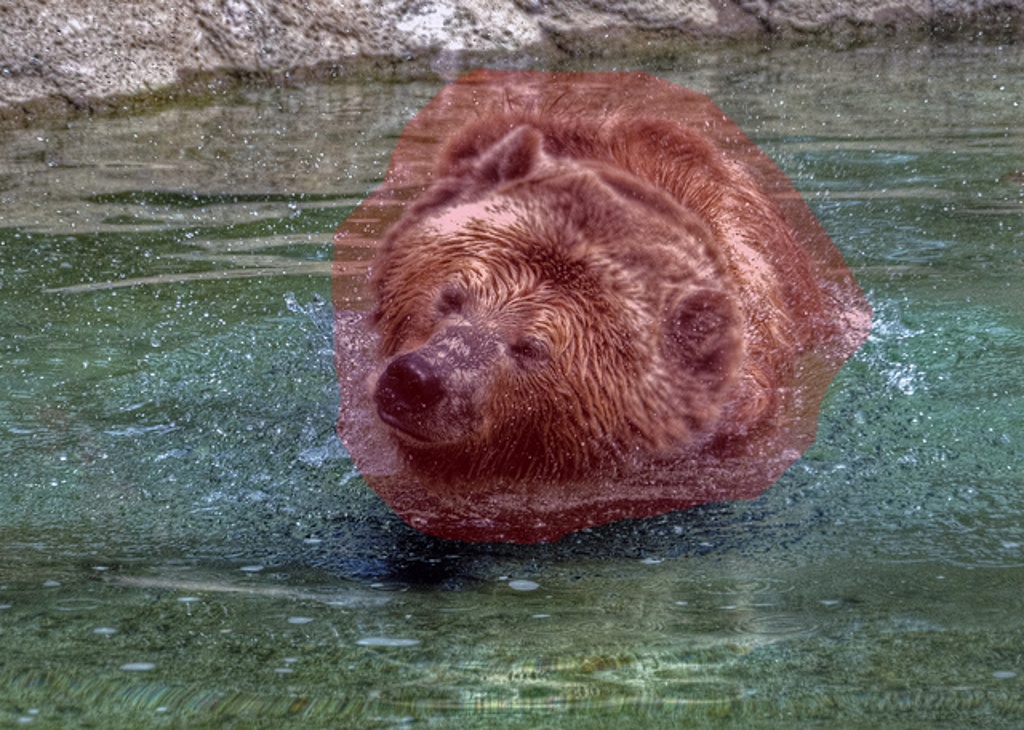}
            \includegraphics[width=0.49\linewidth]{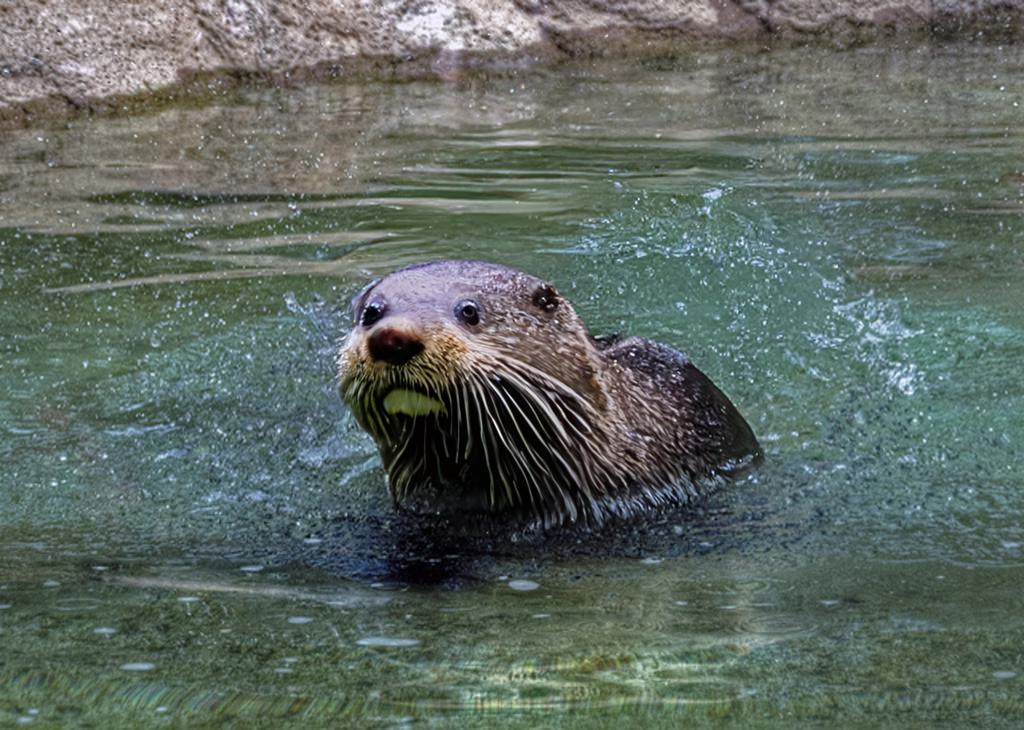}
            \caption{``a playful otter swimming in the river stream''}
        \end{subfigure}
        \hfill
        \begin{subfigure}{0.32\textwidth}
            \includegraphics[width=0.49\linewidth]{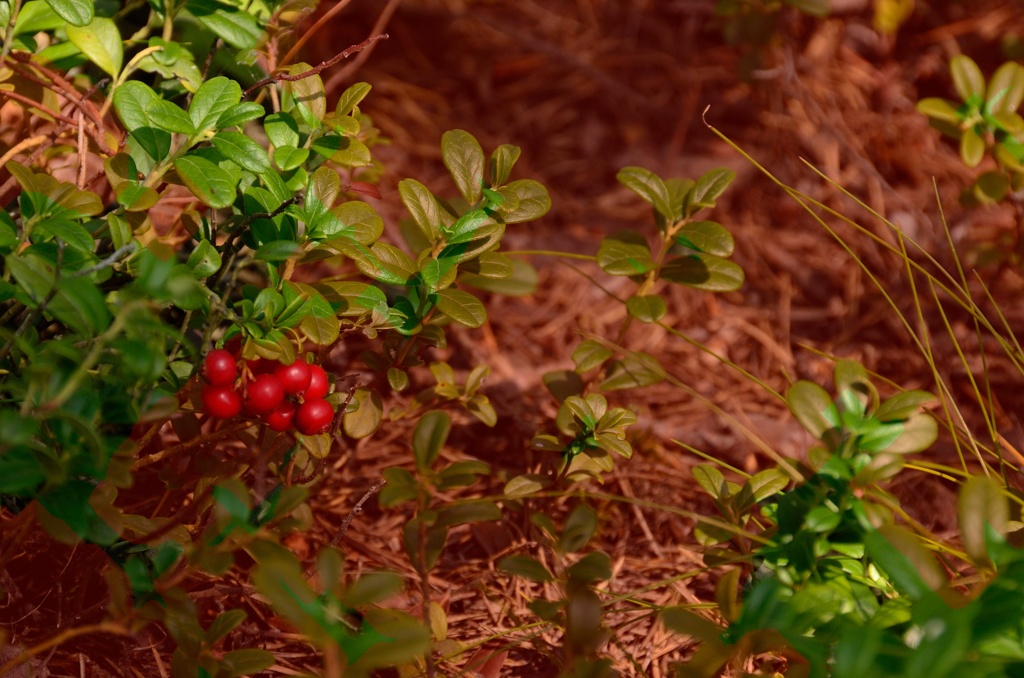}
            \includegraphics[width=0.49\linewidth]{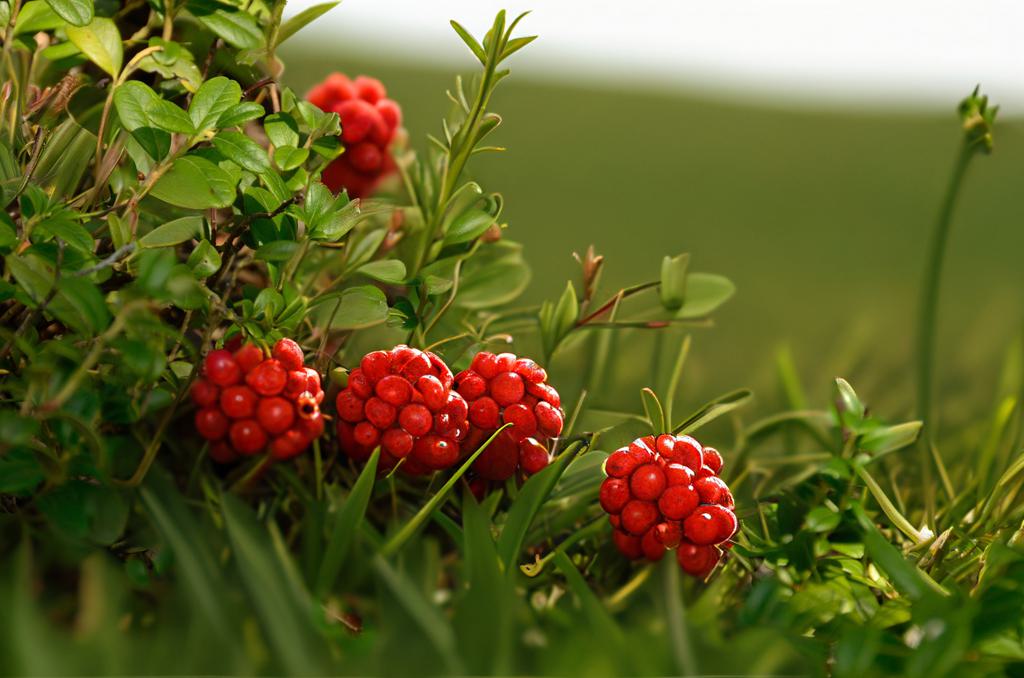}
            \caption{``a cluster of small red berries growing in the grass''}
        \end{subfigure}
        \hfill
        \begin{subfigure}{0.32\textwidth}
            \includegraphics[width=0.49\linewidth]{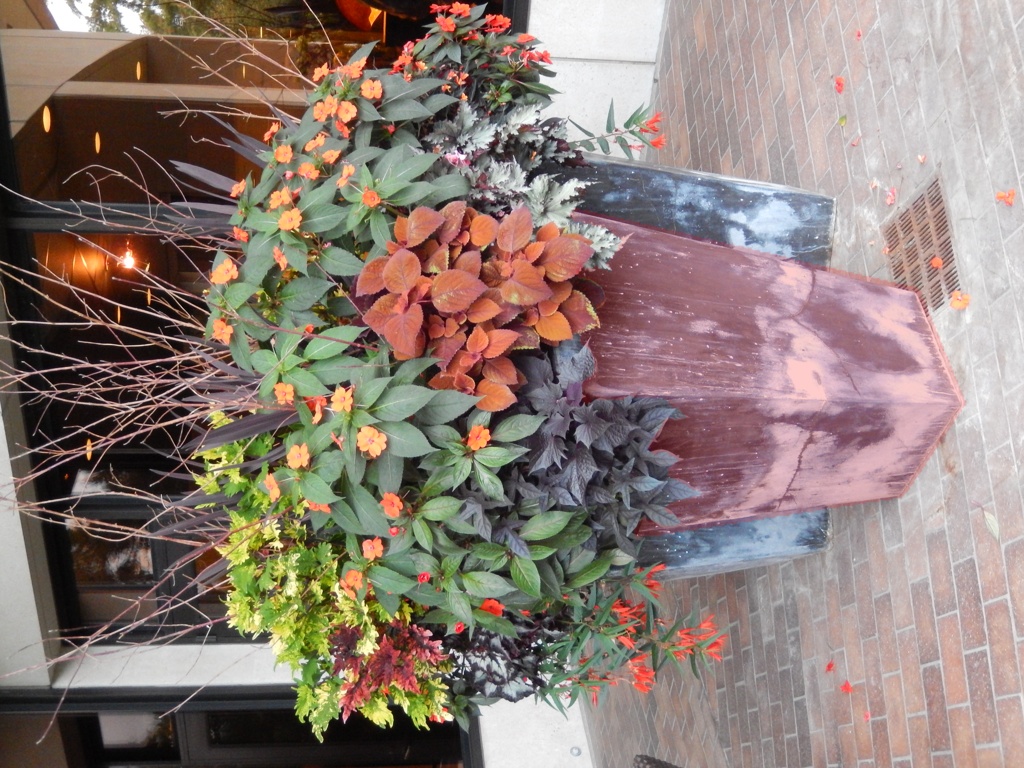}
            \includegraphics[width=0.49\linewidth]{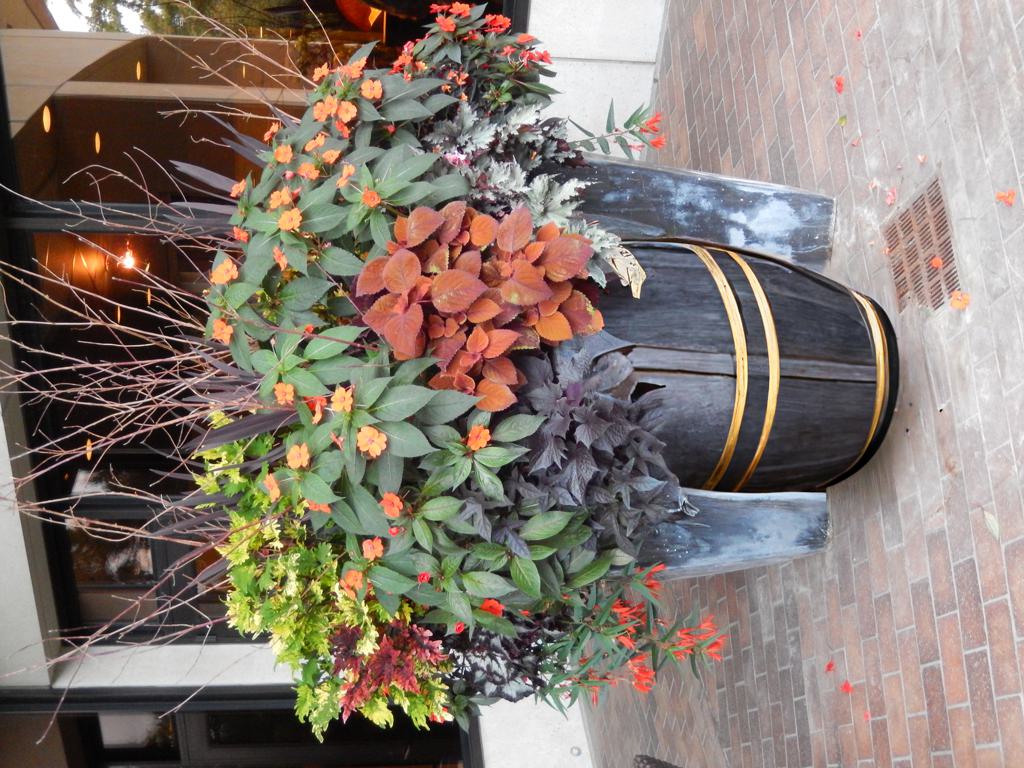}
            \caption{``a rustic wooden barrel planter''}
        \end{subfigure}
    \end{subfigure}
    
    \vspace{1em}
    
    \begin{subfigure}{\textwidth}
        \begin{subfigure}{0.32\textwidth}
            \includegraphics[width=0.49\linewidth]{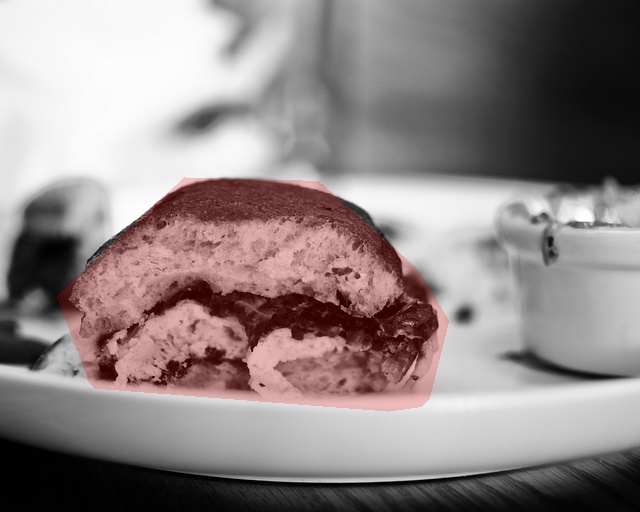}
            \includegraphics[width=0.49\linewidth]{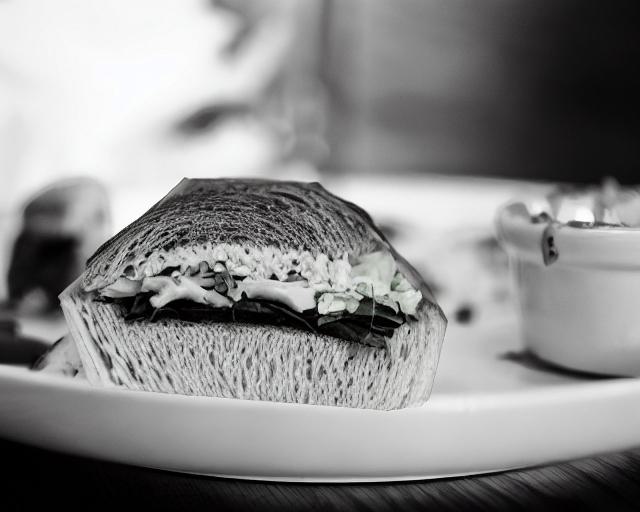}
            \caption{``a fresh, delicious sandwich to complete the meal''}
        \end{subfigure}
        \hfill
        \begin{subfigure}{0.32\textwidth}
            \includegraphics[width=0.49\linewidth]{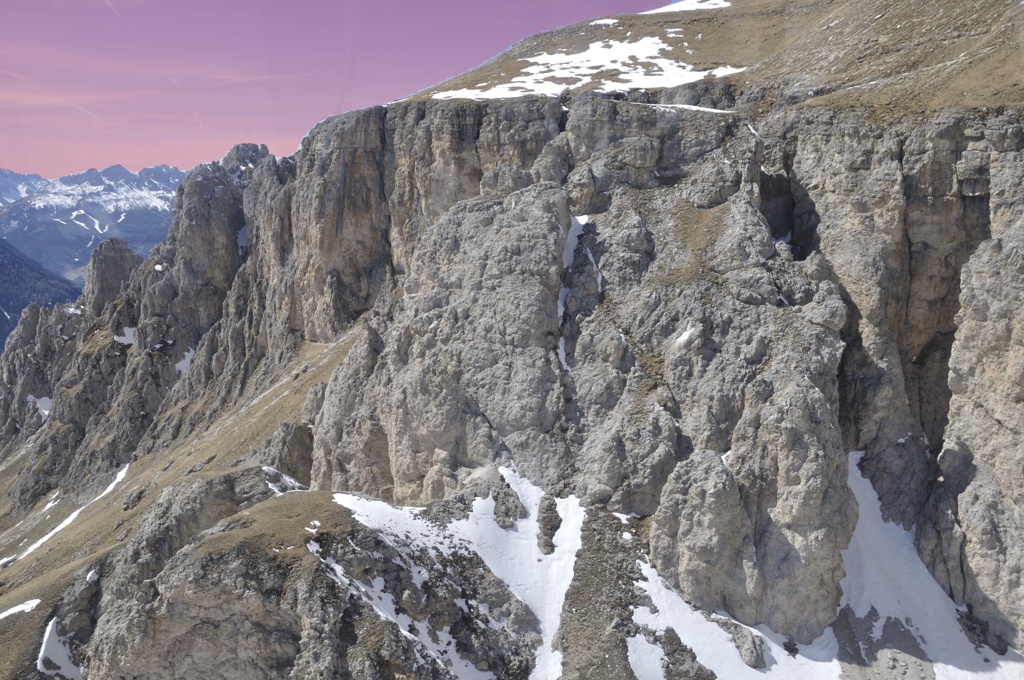}
            \includegraphics[width=0.49\linewidth]{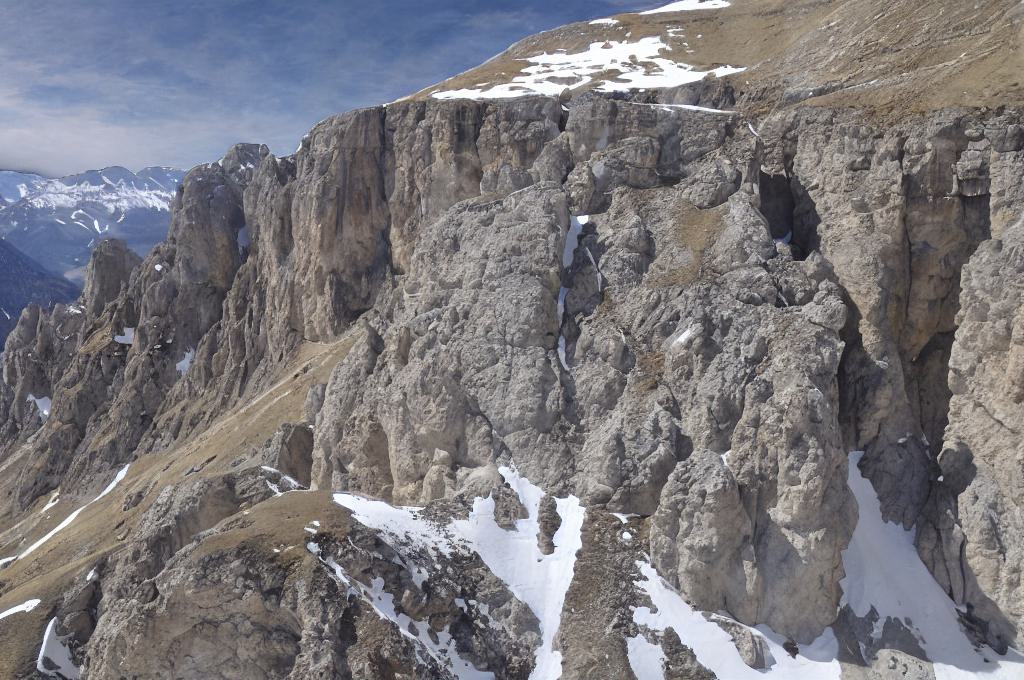}
            \caption{``a clear blue sky to enhance the mountain landscape''}
        \end{subfigure}
        \hfill
        \begin{subfigure}{0.32\textwidth}
            \includegraphics[width=0.49\linewidth]{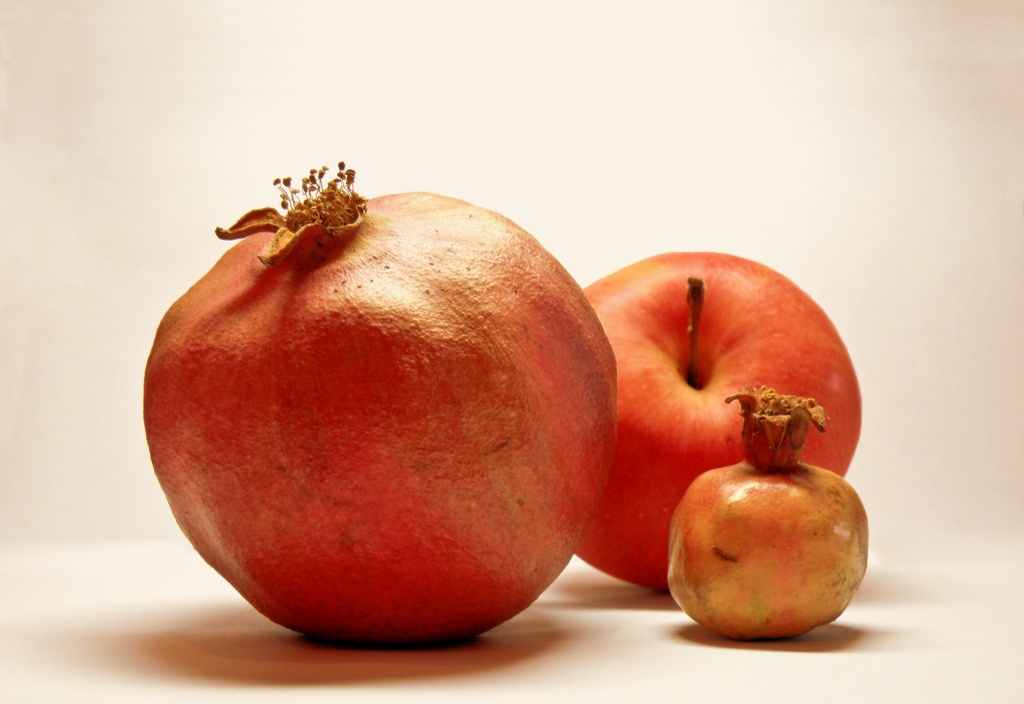}
            \includegraphics[width=0.49\linewidth]{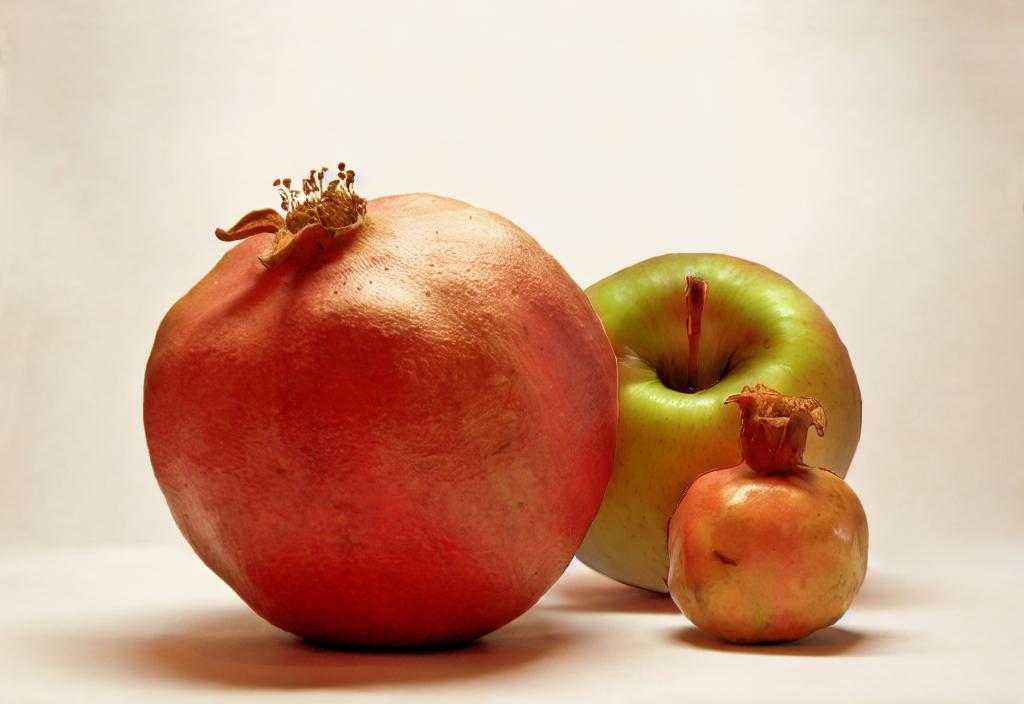}
            \caption{``a ripe golden delicious apple''}
        \end{subfigure}
    \end{subfigure}
    
    \vspace{1em}
    
    \begin{subfigure}{\textwidth}
        \begin{subfigure}{0.32\textwidth}
            \includegraphics[width=0.49\linewidth]{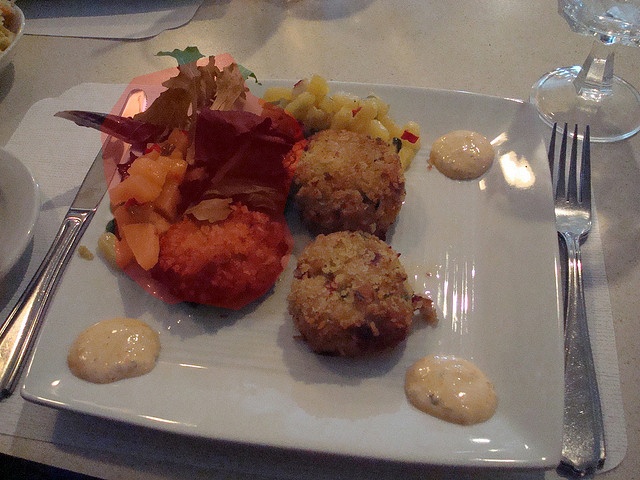}
            \includegraphics[width=0.49\linewidth]{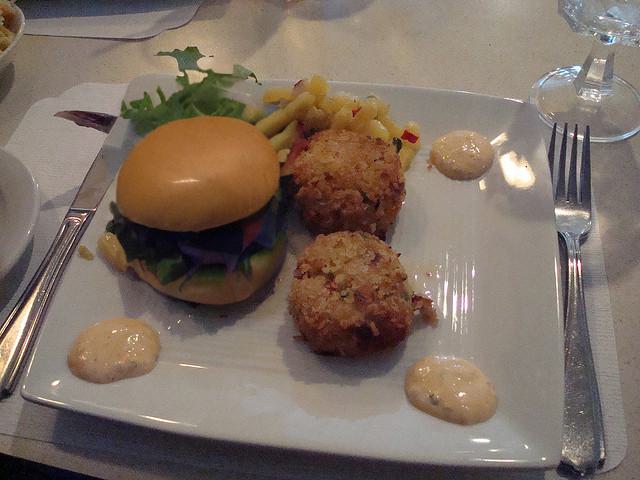}
            \caption{``a delicious cheeseburger to make the meal even more tempting''}
        \end{subfigure}
        \hfill
        \begin{subfigure}{0.32\textwidth}
            \includegraphics[width=0.49\linewidth]{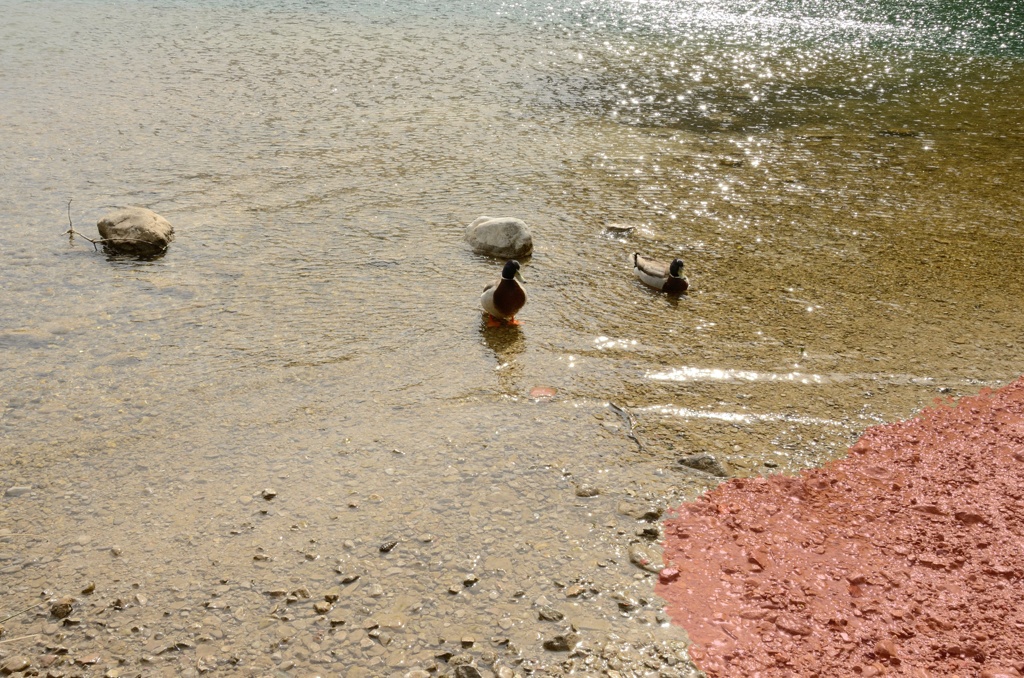}
            \includegraphics[width=0.49\linewidth]{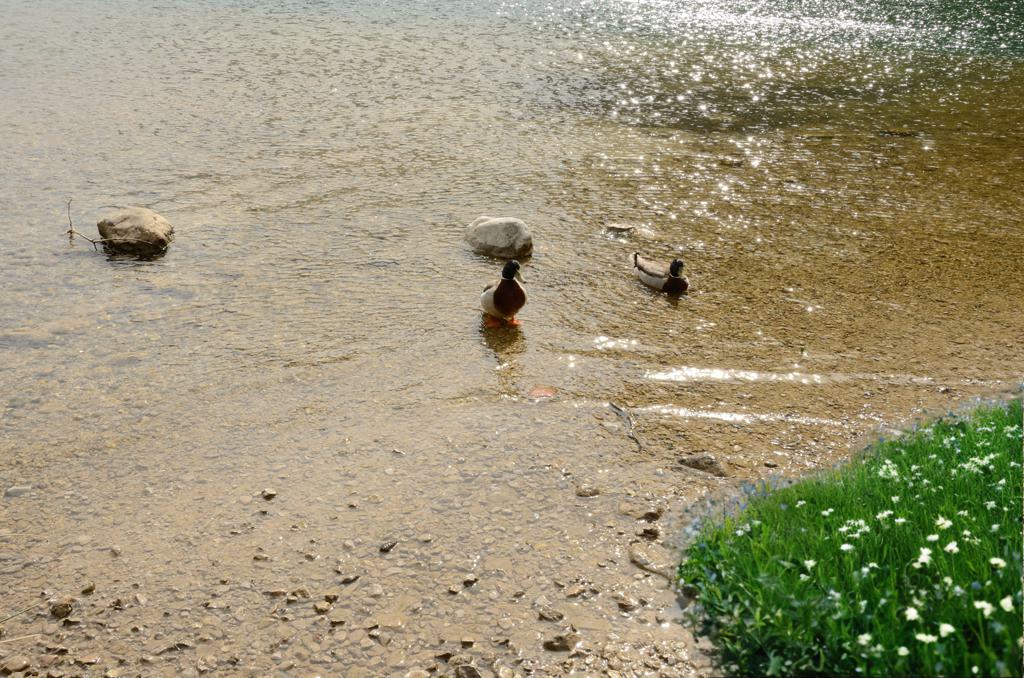}
            \caption{``a lush green meadow, adding a touch of nature to the serene landscape''}
        \end{subfigure}
        \hfill
        \begin{subfigure}{0.32\textwidth}
            \includegraphics[width=0.49\linewidth]{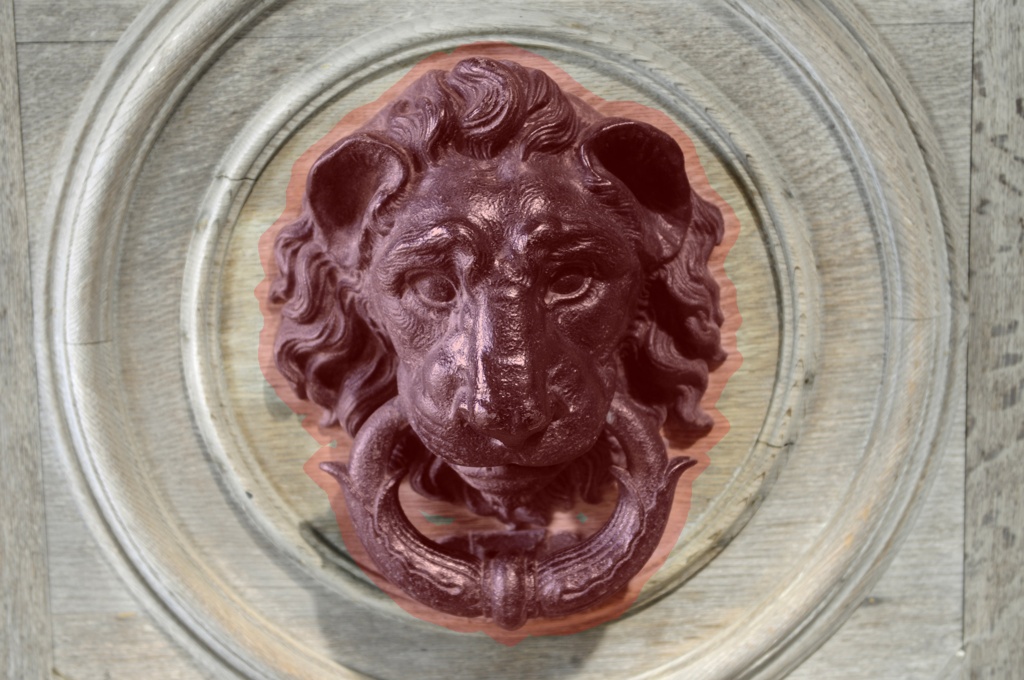}
            \includegraphics[width=0.49\linewidth]{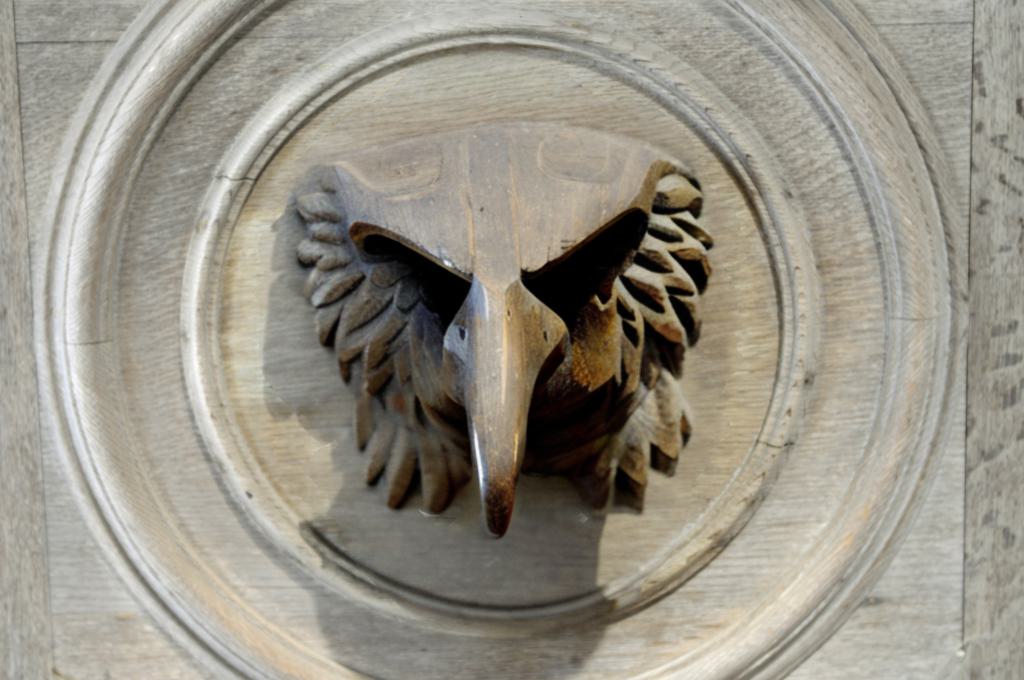}
            \caption{``an intricately carved wooden eagle head''}
        \end{subfigure}
    \end{subfigure}
    
    \vspace{1em}
    
    \begin{subfigure}{\textwidth}
        \begin{subfigure}{0.32\textwidth}
            \includegraphics[width=0.49\linewidth]{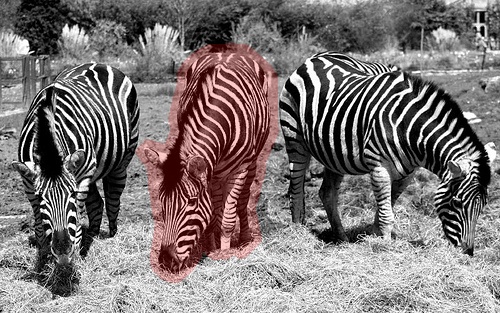}
            \includegraphics[width=0.49\linewidth]{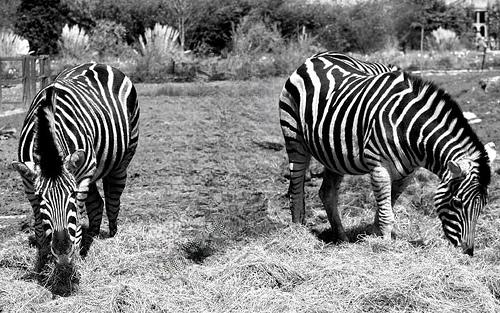}
            \caption{No prompt}
        \end{subfigure}
        \hfill
        \begin{subfigure}{0.32\textwidth}
            \includegraphics[width=0.49\linewidth]{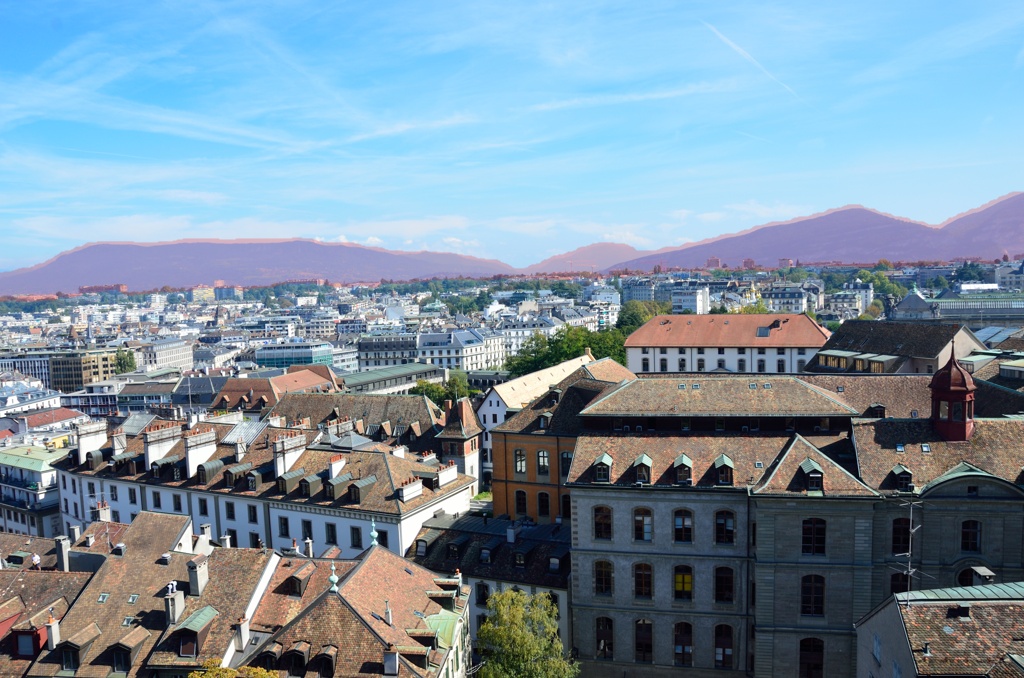}
            \includegraphics[width=0.49\linewidth]{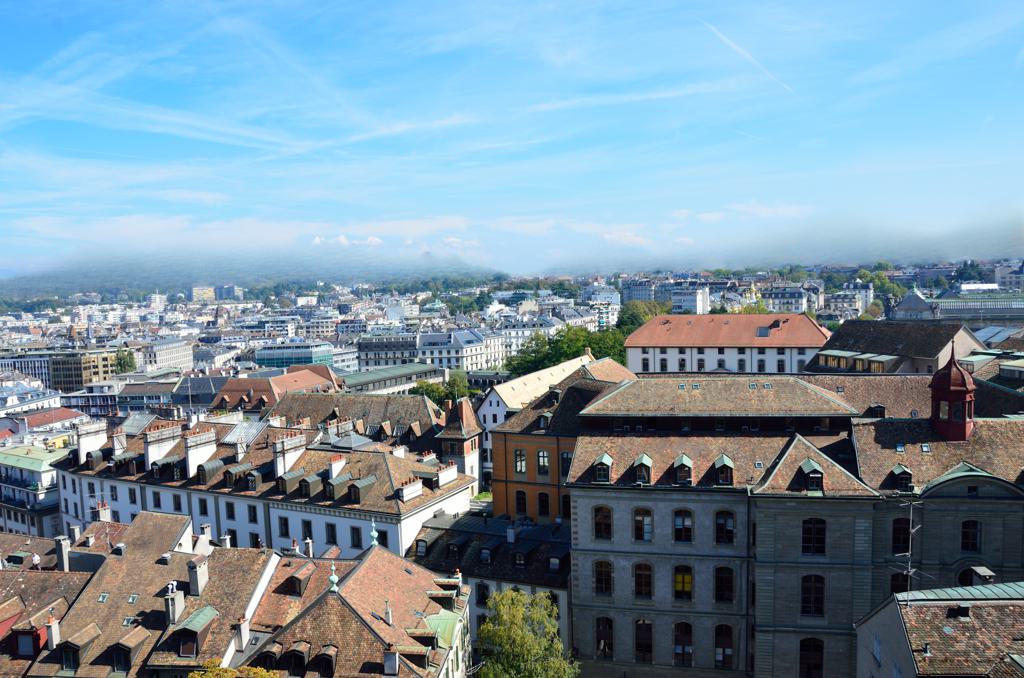}
            \caption{No prompt}
        \end{subfigure}
        \hfill
        \begin{subfigure}{0.32\textwidth}
            \includegraphics[width=0.49\linewidth]{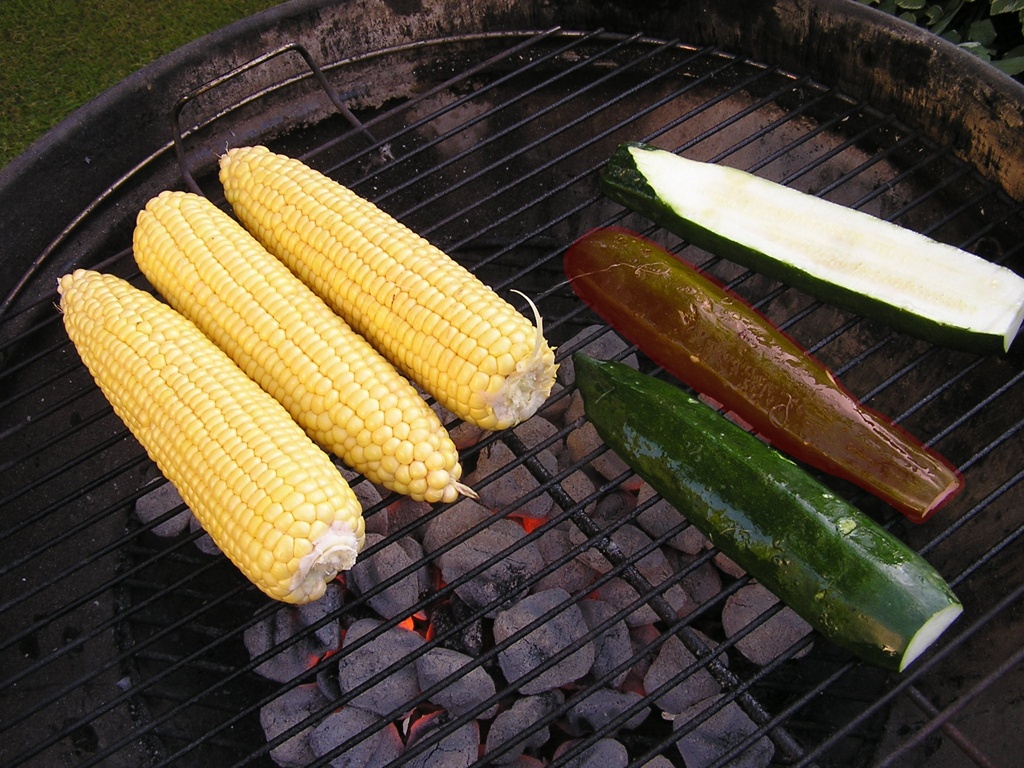}
            \includegraphics[width=0.49\linewidth]{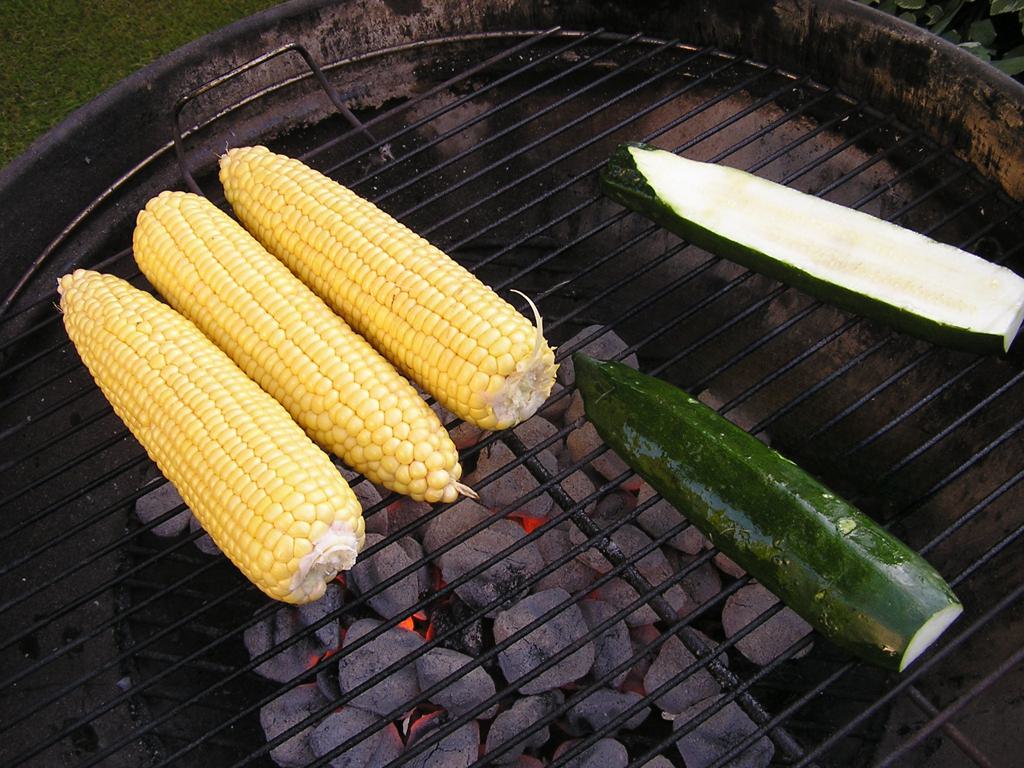}
            \caption{No prompt}
        \end{subfigure}
    \end{subfigure}
    
    \caption{Example pairs of original images (with inpainting mask overlaid in semi-transparent red) and their corresponding inpainted results across three datasets: COCO (first column), RAISE (second column), and OpenImages (third column). Each row showcases results from a different inpainting model: BrushNet, PowerPaint, HD-Painter, ControlNet, Inpaint-Anything, and Remove-Anything. The text below each pair shows the prompt used for text-guided models.}
    \label{fig:model_comparison_examples}
\end{figure*}

\begin{figure*}[p]
    \centering
    
    \begin{subfigure}{\textwidth}
        \begin{subfigure}{0.32\textwidth}
            \includegraphics[width=0.49\linewidth]{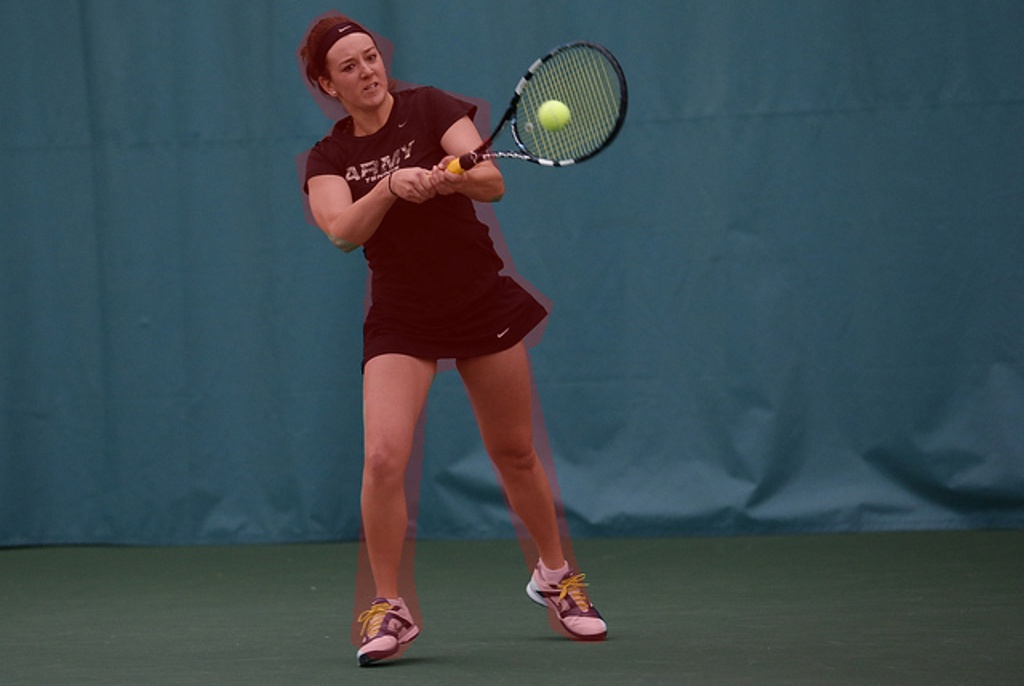}
            \includegraphics[width=0.49\linewidth]{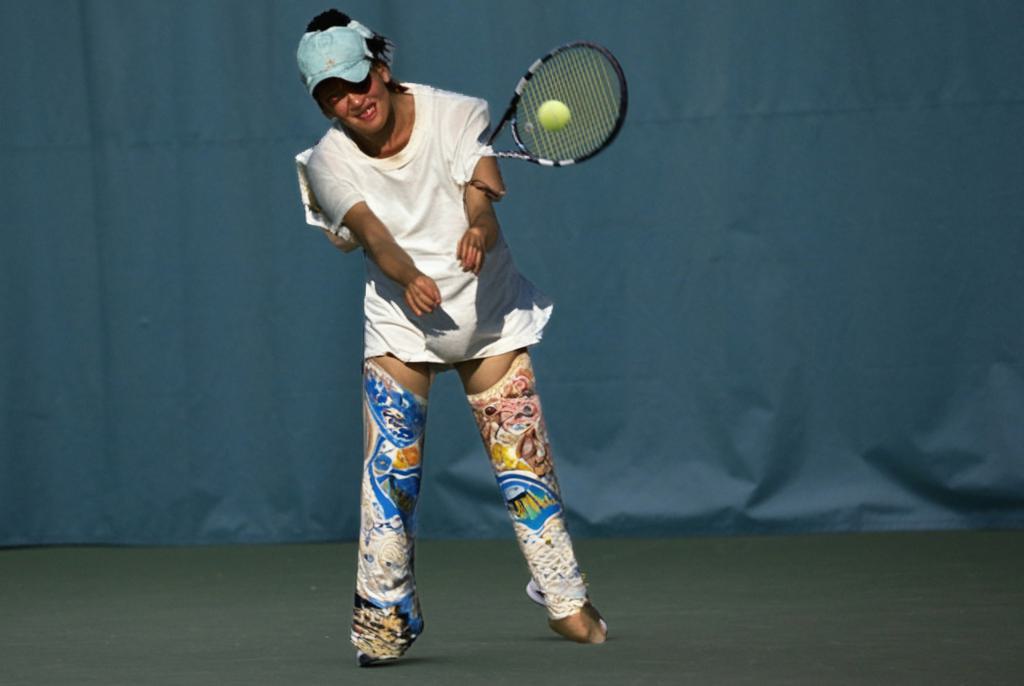}
            \caption{``a person playing volleyball on the beach''}
        \end{subfigure}
        \hfill
        \begin{subfigure}{0.32\textwidth}
            \includegraphics[width=0.49\linewidth]{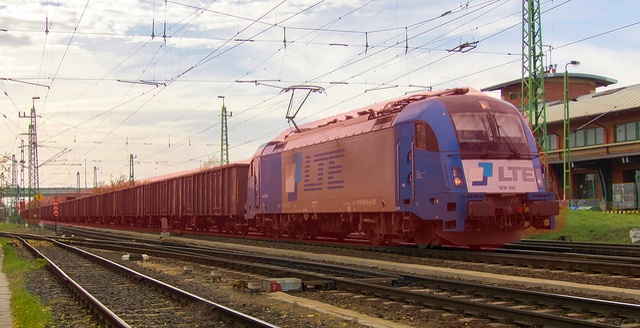}
            \includegraphics[width=0.49\linewidth]{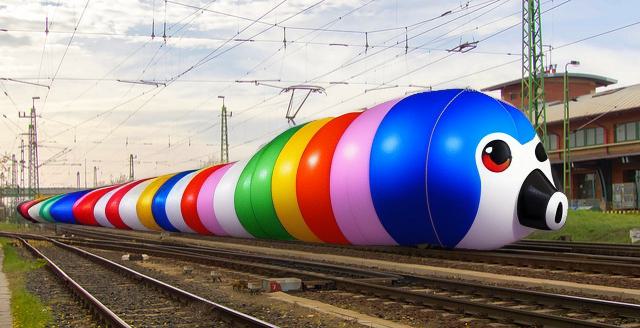}
            \caption{``a colorful hot air balloon floating in the sky''}
        \end{subfigure}
        \hfill
        \begin{subfigure}{0.32\textwidth}
            \includegraphics[width=0.49\linewidth]{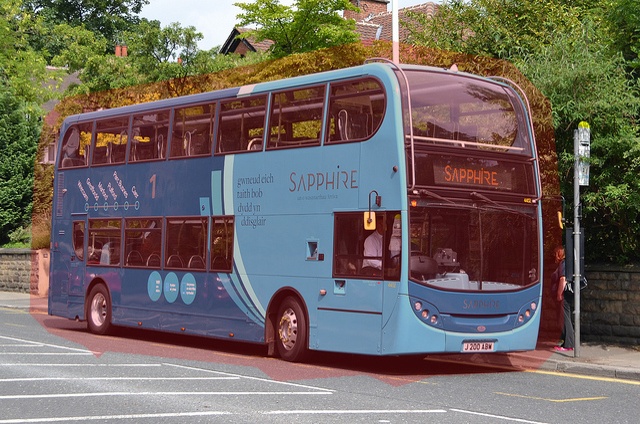}
            \includegraphics[width=0.49\linewidth]{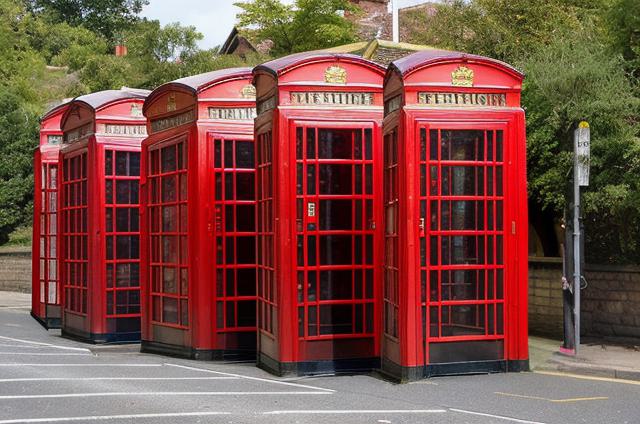}
            \caption{``a red London phone booth''}
        \end{subfigure}
    \end{subfigure}
    
    \vspace{1em}
    
    \begin{subfigure}{\textwidth}
        \begin{subfigure}{0.32\textwidth}
            \includegraphics[width=0.49\linewidth]{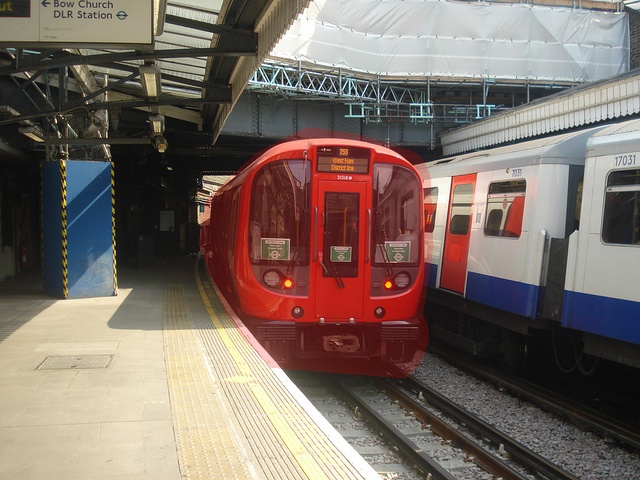}
            \includegraphics[width=0.49\linewidth]{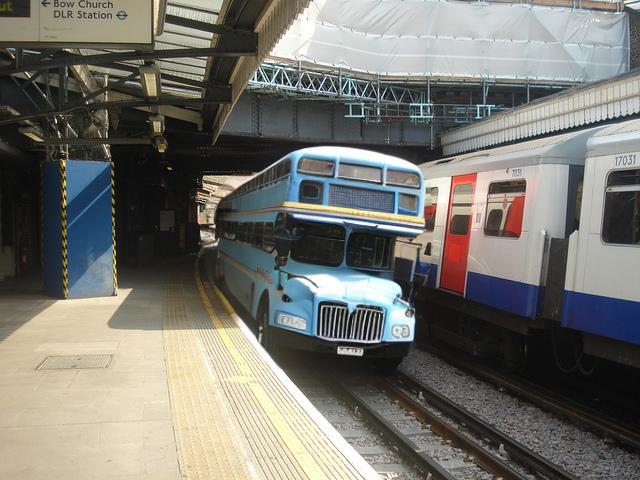}
            \caption{``a blue bus traveling down the tracks''}
        \end{subfigure}
        \hfill
        \begin{subfigure}{0.32\textwidth}
            \includegraphics[width=0.49\linewidth]{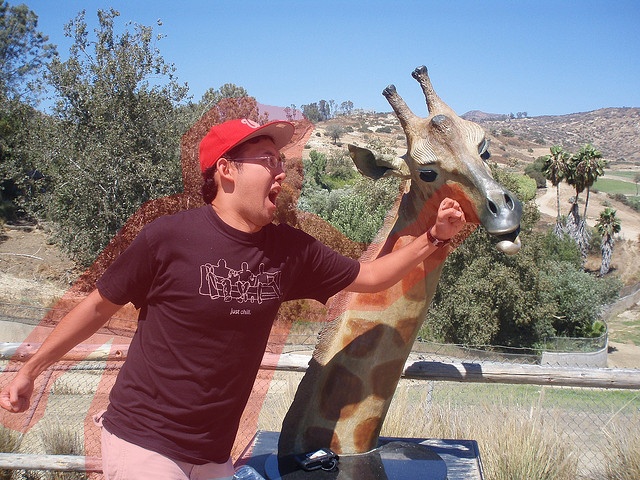}
            \includegraphics[width=0.49\linewidth]{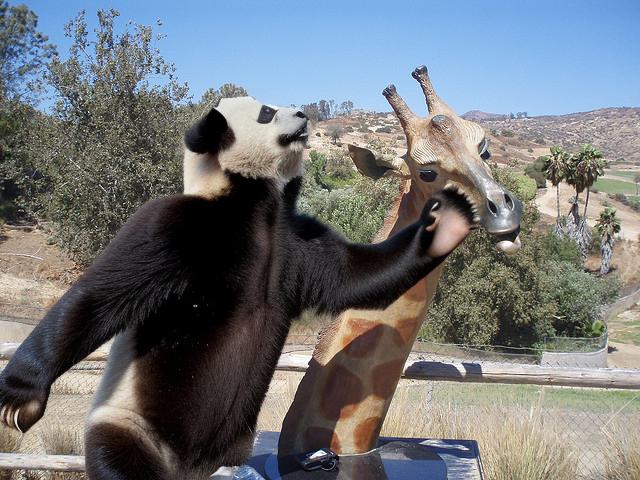}
            \caption{``a playful panda bear imitating a martial arts move''}
        \end{subfigure}
        \hfill
        \begin{subfigure}{0.32\textwidth}
            \includegraphics[width=0.49\linewidth]{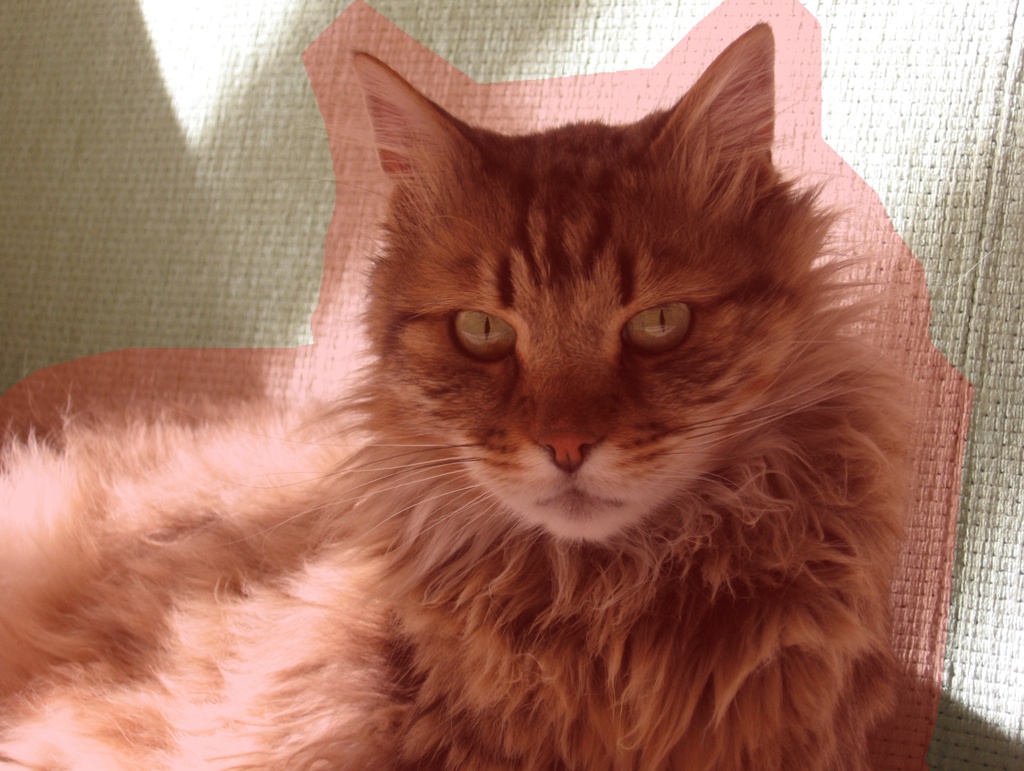}
            \includegraphics[width=0.49\linewidth]{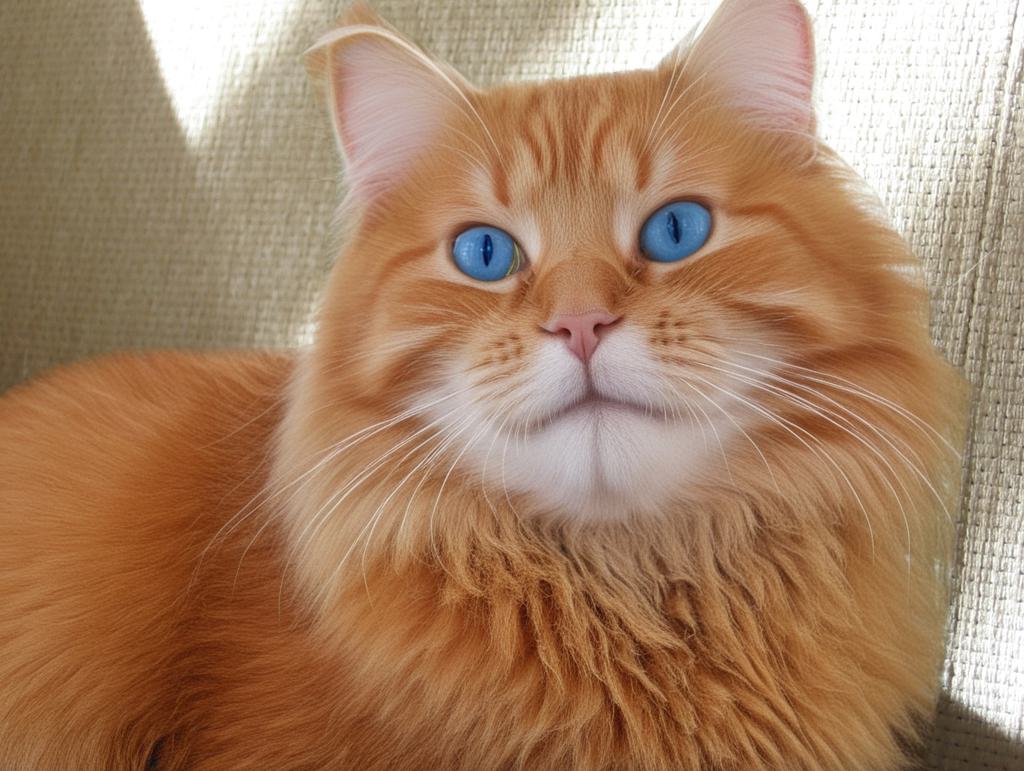}
            \caption{``a fluffy orange tabby cat with bright blue''}
        \end{subfigure}
    \end{subfigure}
    
    \vspace{1em}
    
    \begin{subfigure}{\textwidth}
        \begin{subfigure}{0.32\textwidth}
            \includegraphics[width=0.49\linewidth]{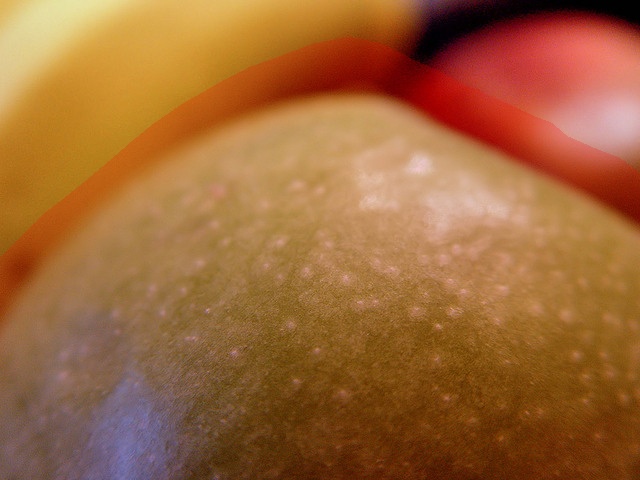}
            \includegraphics[width=0.49\linewidth]{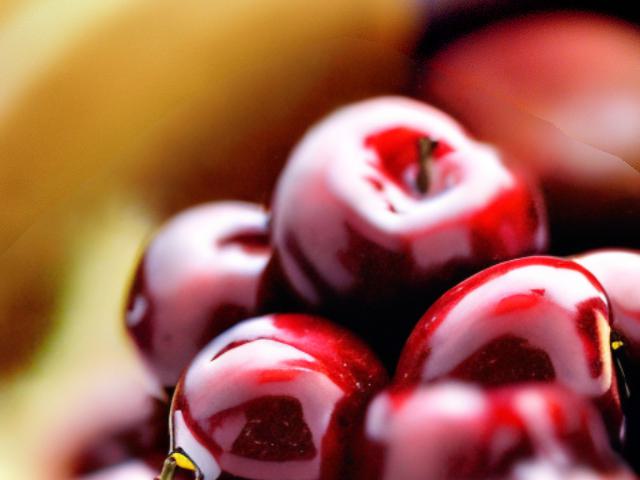}
            \caption{``vibrant red cherries to create a fruity collage effect''}
        \end{subfigure}
        \hfill
        \begin{subfigure}{0.32\textwidth}
            \includegraphics[width=0.49\linewidth]{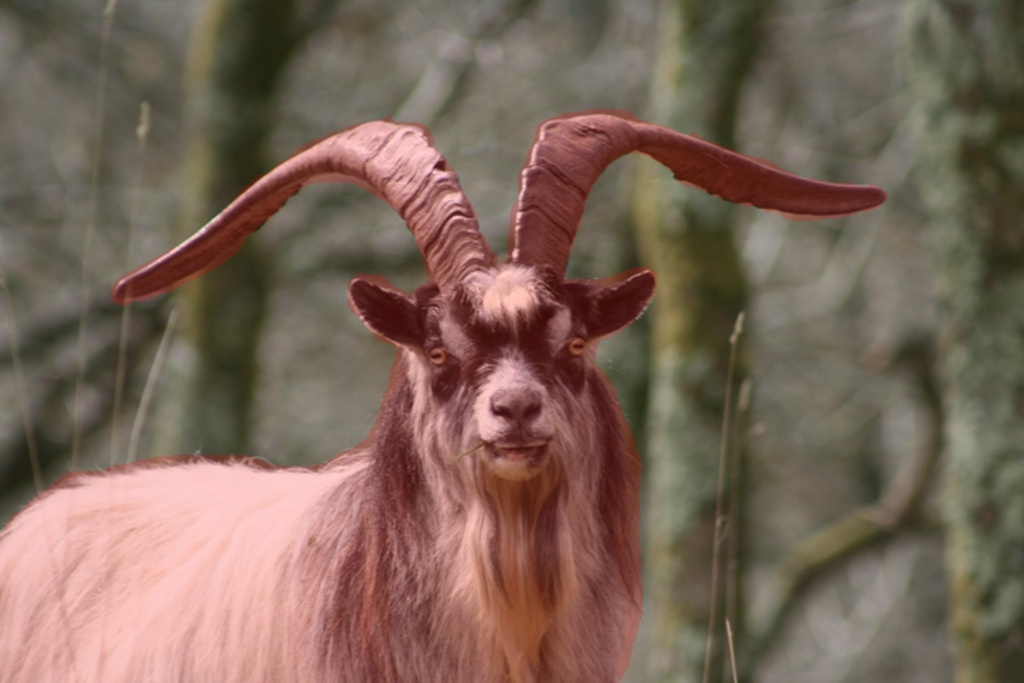}
            \includegraphics[width=0.49\linewidth]{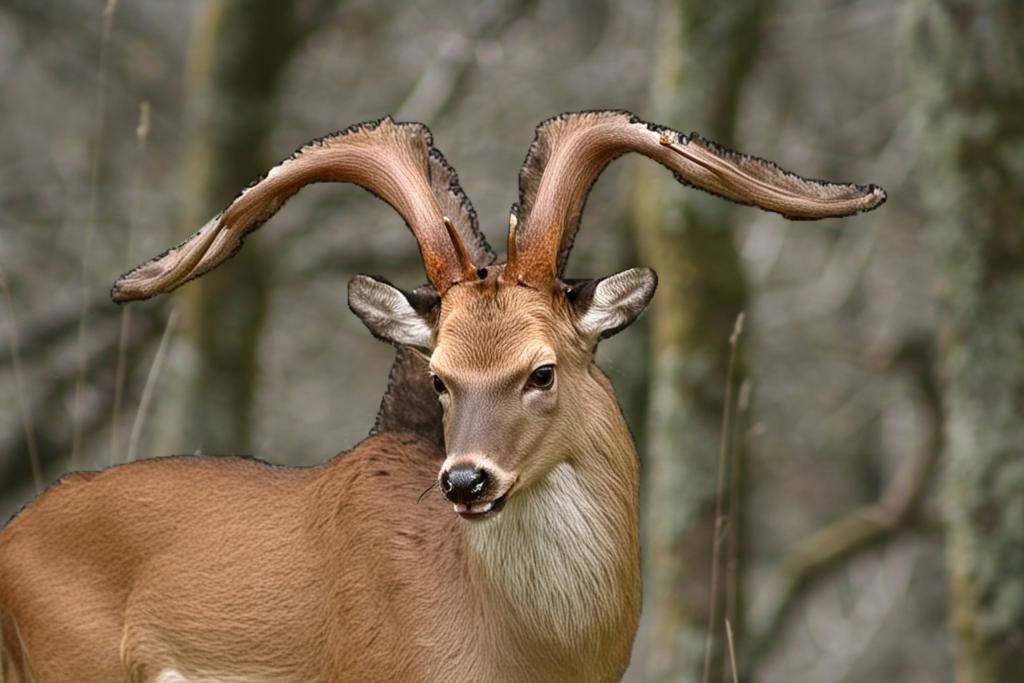}
            \caption{``a majestic deer with large ant''}
        \end{subfigure}
        \hfill
        \begin{subfigure}{0.32\textwidth}
            \includegraphics[width=0.49\linewidth]{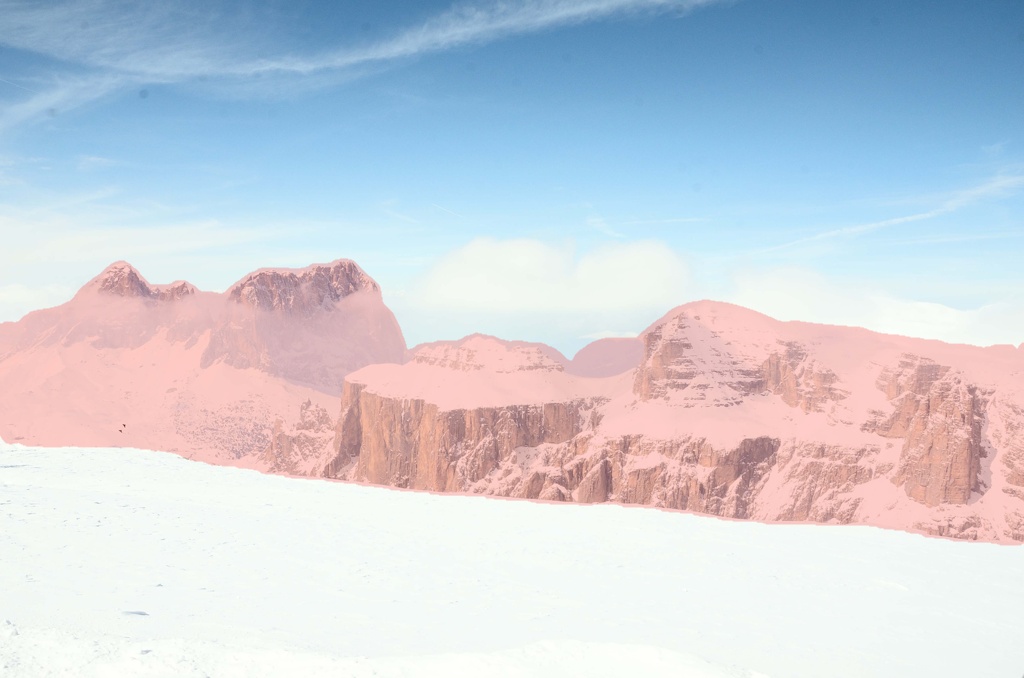}
            \includegraphics[width=0.49\linewidth]{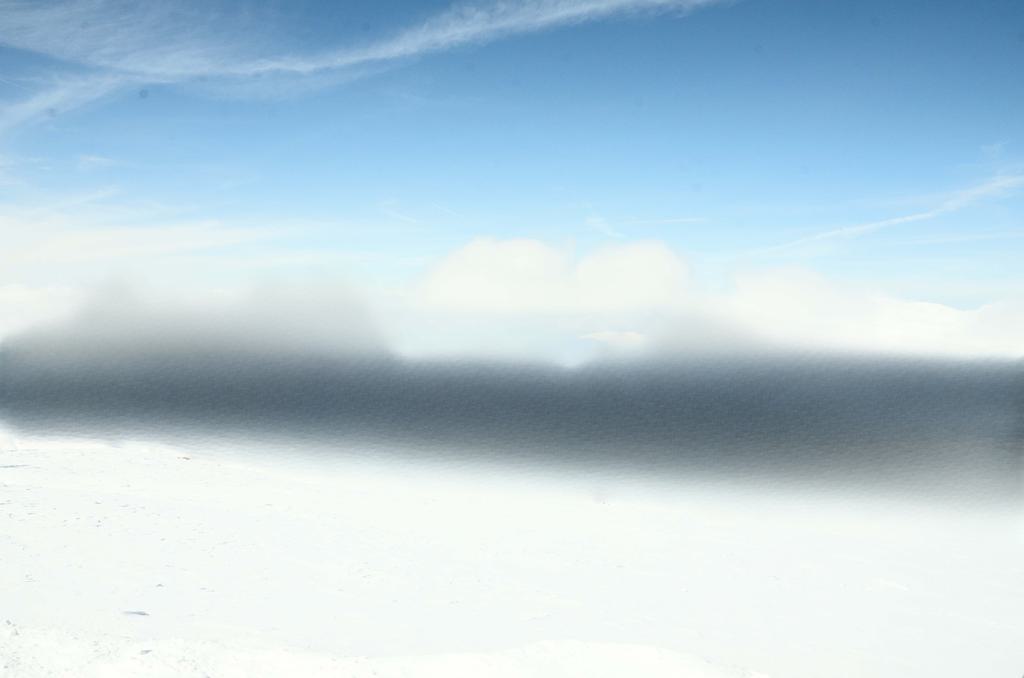}
            \caption{No prompt}
        \end{subfigure}
    \end{subfigure}
    
    \vspace{1em}
    
    \begin{subfigure}{\textwidth}
        \begin{subfigure}{0.32\textwidth}
            \includegraphics[width=0.49\linewidth]{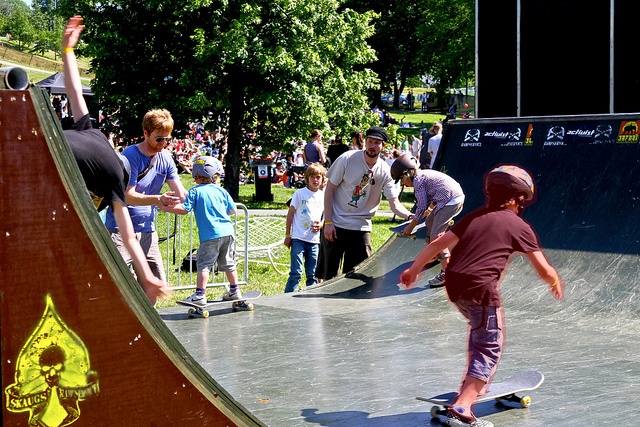}
            \includegraphics[width=0.49\linewidth]{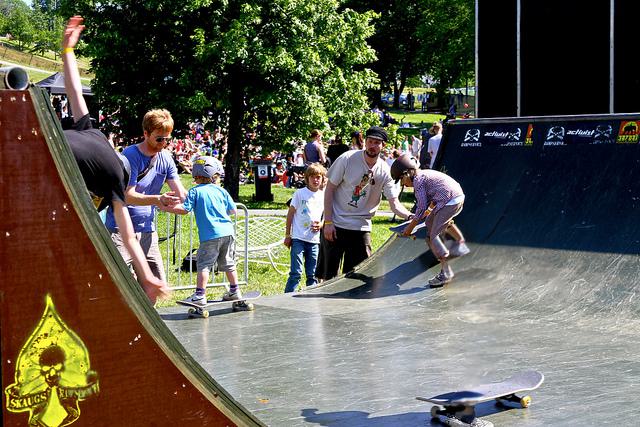}
            \caption{``a group of young adults playing frisbee''}
        \end{subfigure}
        \hfill
        \begin{subfigure}{0.32\textwidth}
            \includegraphics[width=0.49\linewidth]{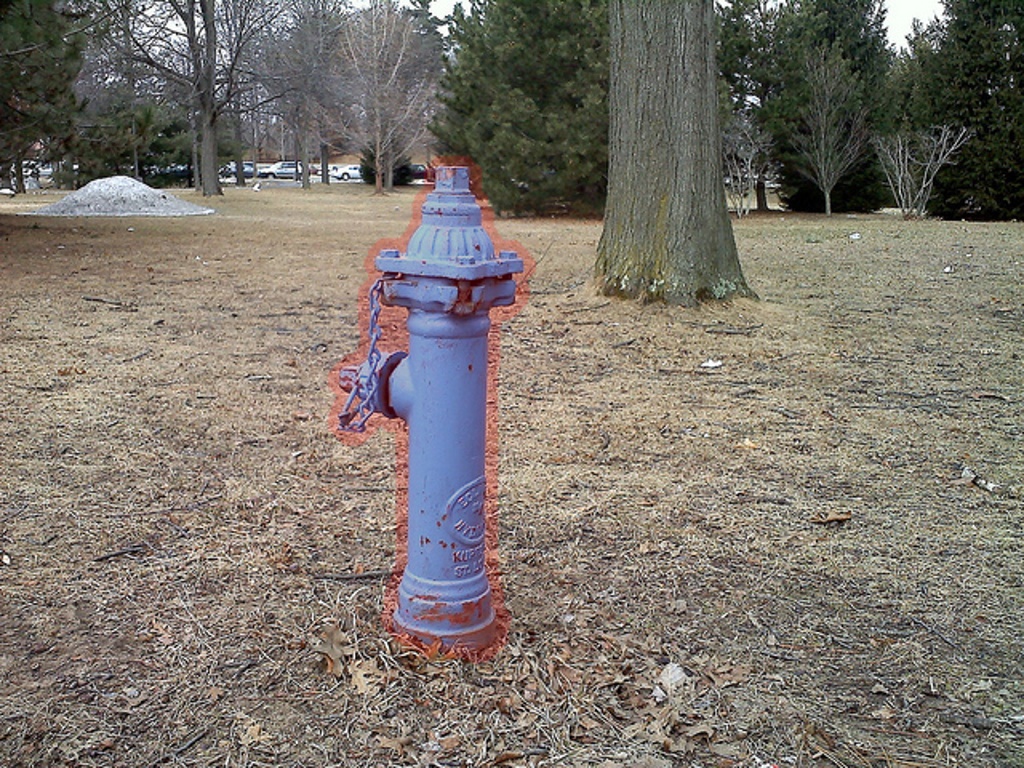}
            \includegraphics[width=0.49\linewidth]{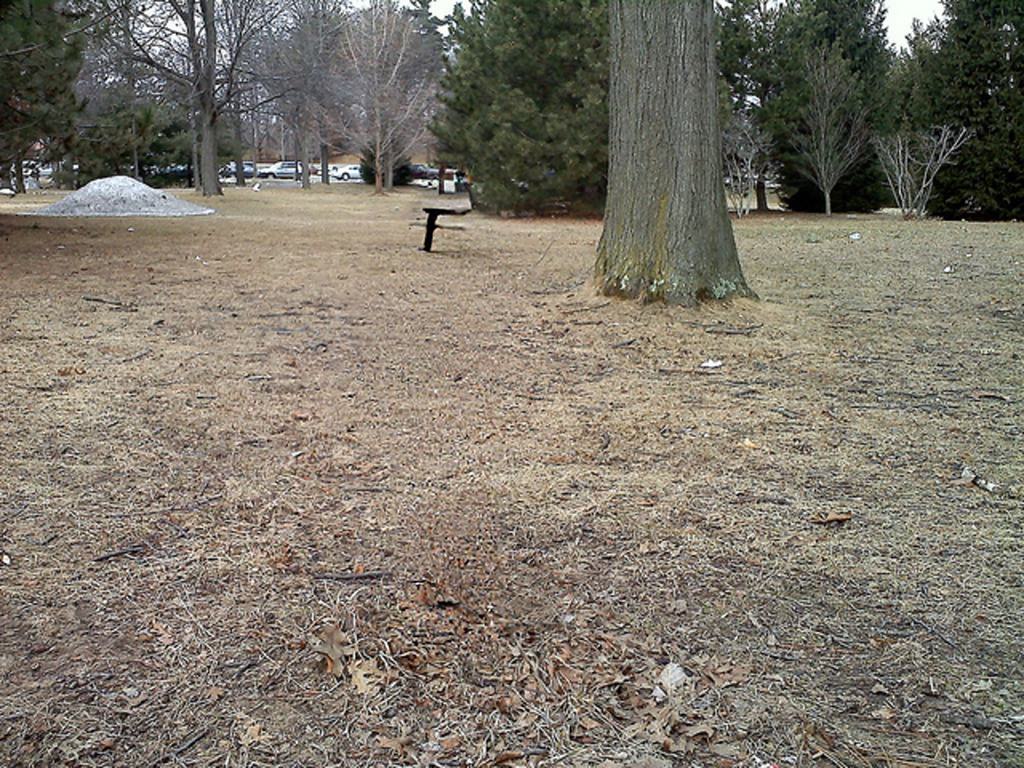}
            \caption{``a bright red mailbox to blend seamlessly into the park scene''}
        \end{subfigure}
        \hfill
        \begin{subfigure}{0.32\textwidth}
            \includegraphics[width=0.49\linewidth]{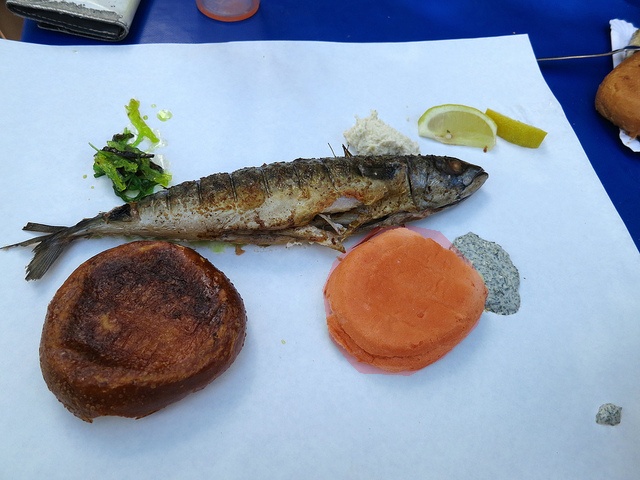}
            \includegraphics[width=0.49\linewidth]{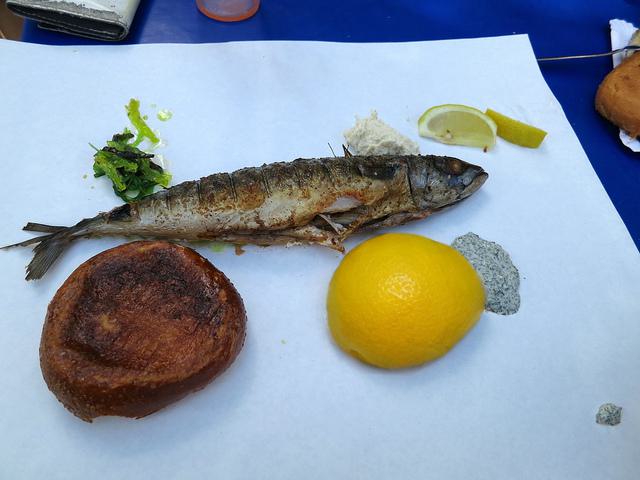}
            \caption{``a colorful bowl of fruit salad''}
        \end{subfigure}
    \end{subfigure}
    
    \vspace{1em}
    
    \begin{subfigure}{\textwidth}
        \begin{subfigure}{0.32\textwidth}
            \includegraphics[width=0.49\linewidth]{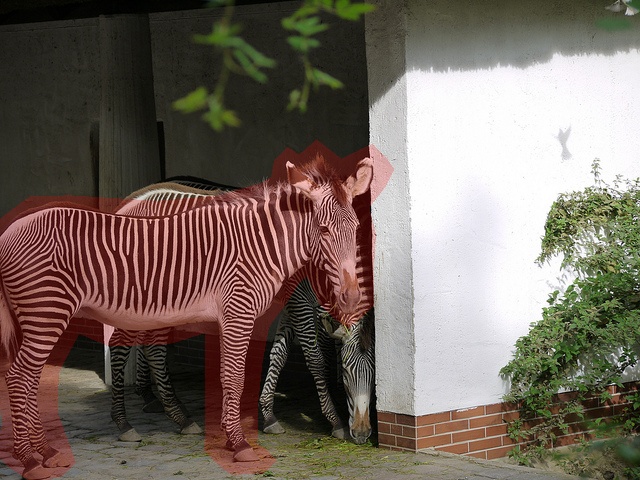}
            \includegraphics[width=0.49\linewidth]{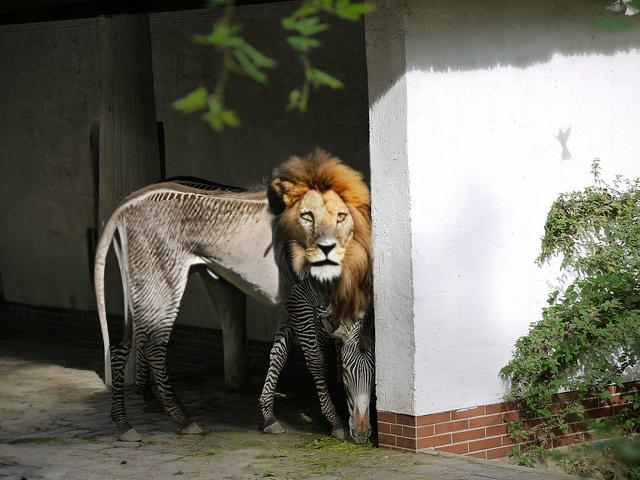}
            \caption{``a majestic lion standing proudly in the savanna''}
        \end{subfigure}
        \hfill
        \begin{subfigure}{0.32\textwidth}
            \includegraphics[width=0.49\linewidth]{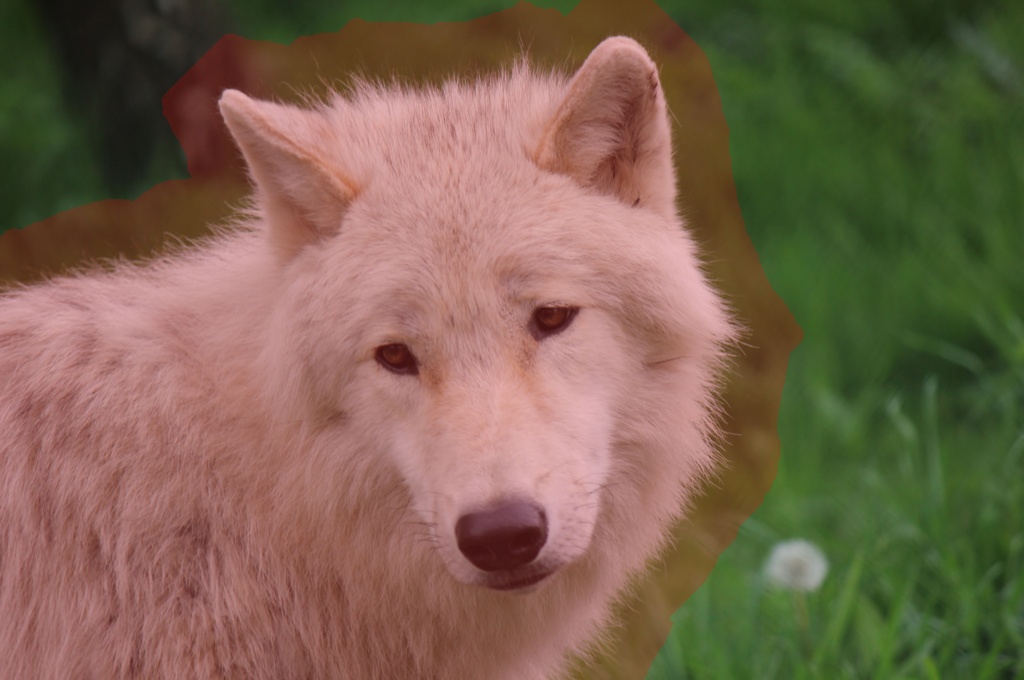}
            \includegraphics[width=0.49\linewidth]{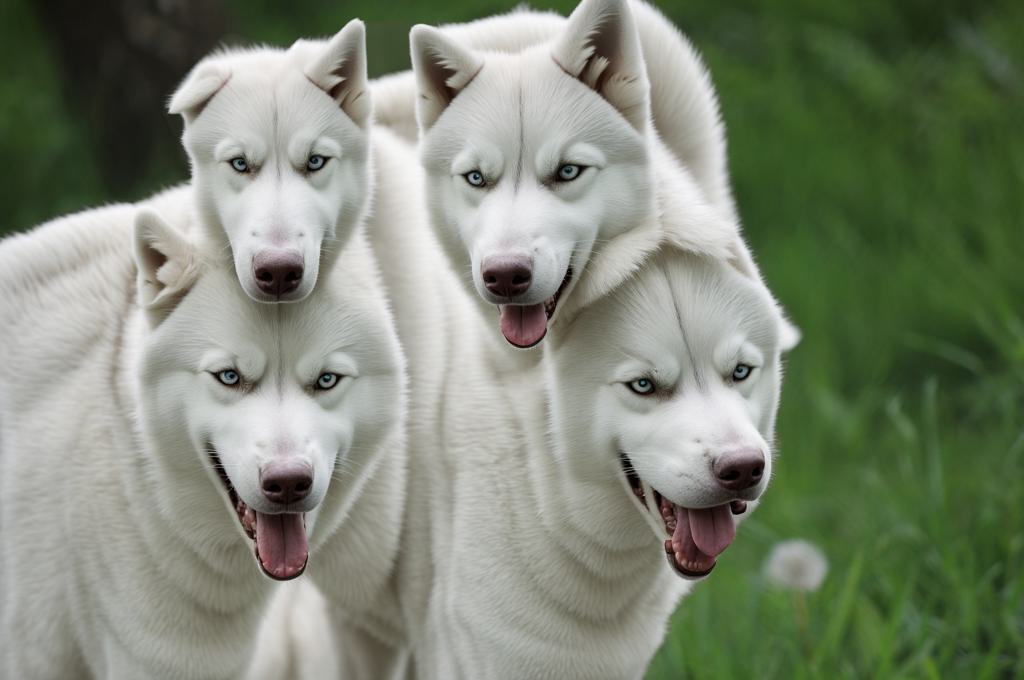}
            \caption{``a large, majestic white husky standing''}
        \end{subfigure}
        \hfill
        \begin{subfigure}{0.32\textwidth}
            \includegraphics[width=0.49\linewidth]{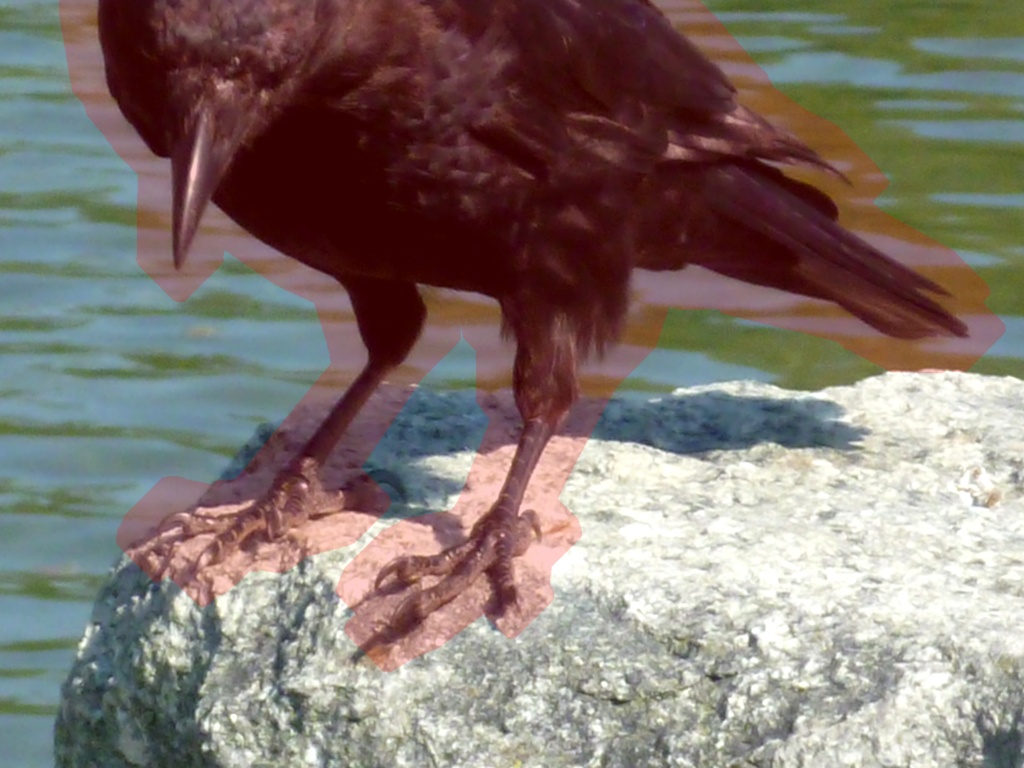}
            \includegraphics[width=0.49\linewidth]{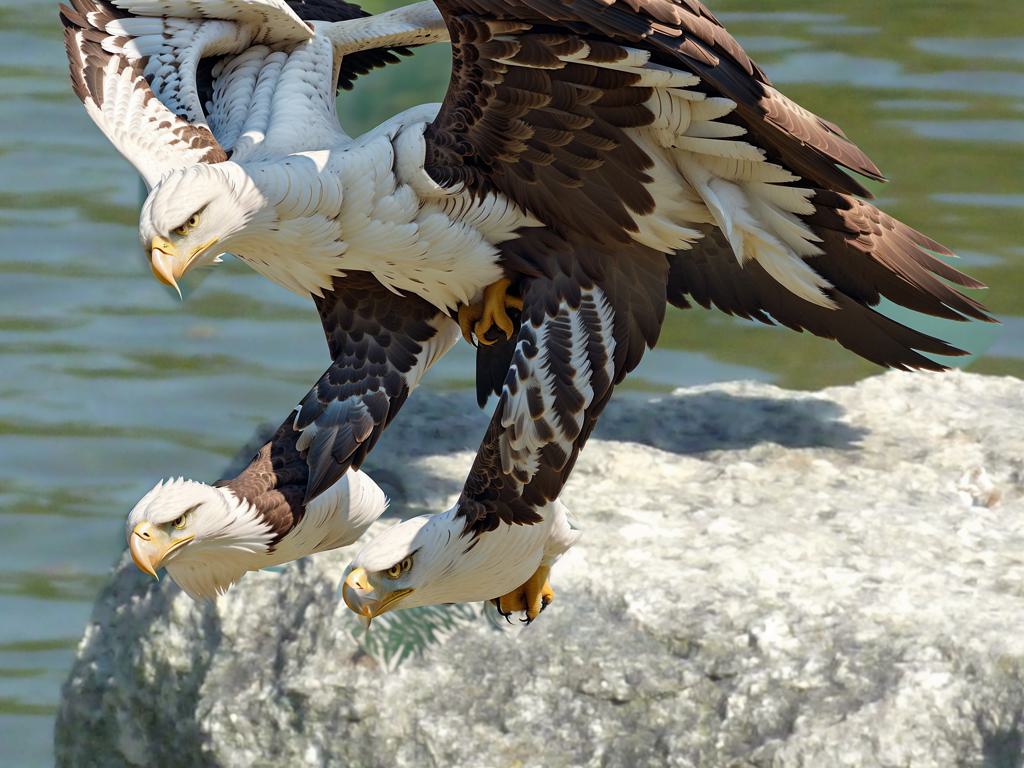}
            \caption{``a majestic eagle perched on''}
        \end{subfigure}
    \end{subfigure}
    
    \vspace{1em}
    
    \begin{subfigure}{\textwidth}
        \begin{subfigure}{0.32\textwidth}
            \includegraphics[width=0.49\linewidth]{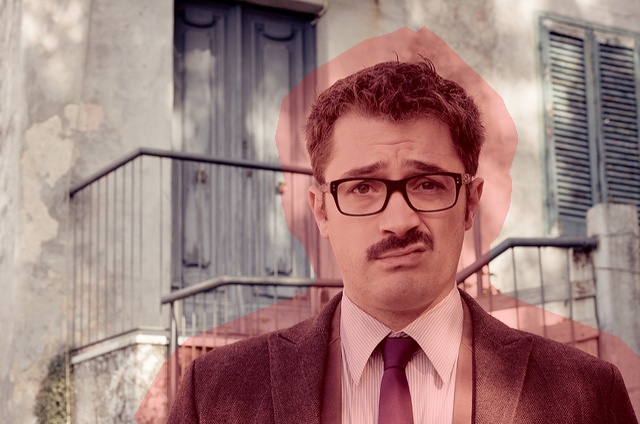}
            \includegraphics[width=0.49\linewidth]{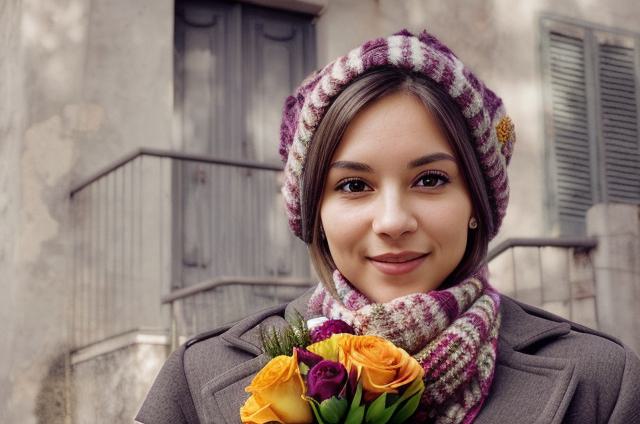}
            \caption{``a woman wearing a scarf and holding a bouquet of flowers''}
        \end{subfigure}
        \hfill
        \begin{subfigure}{0.32\textwidth}
            \includegraphics[width=0.49\linewidth]{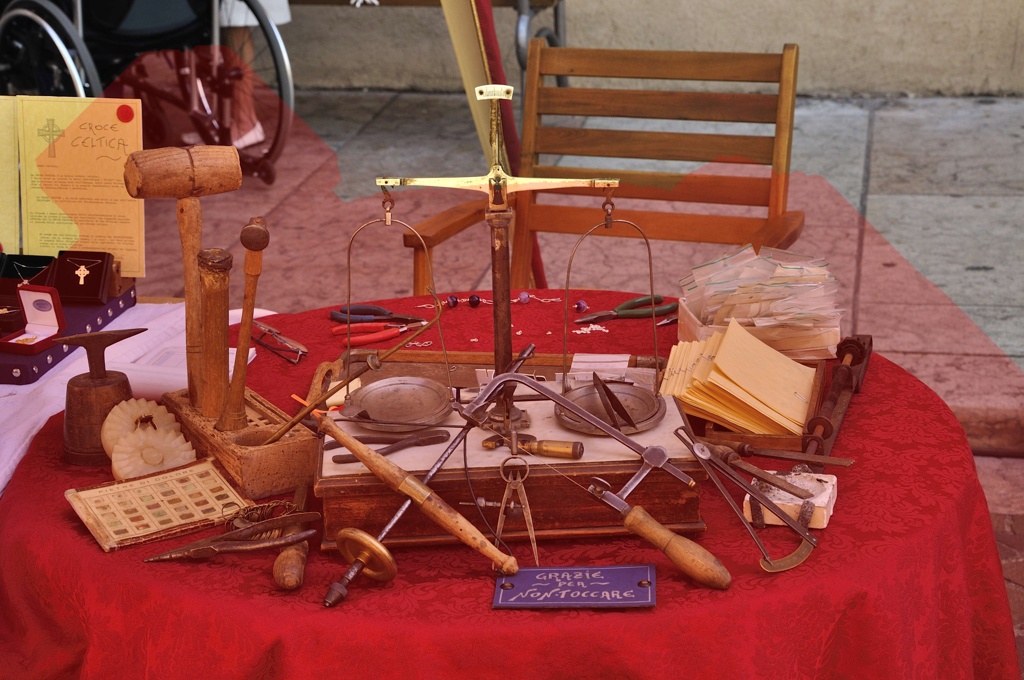}
            \includegraphics[width=0.49\linewidth]{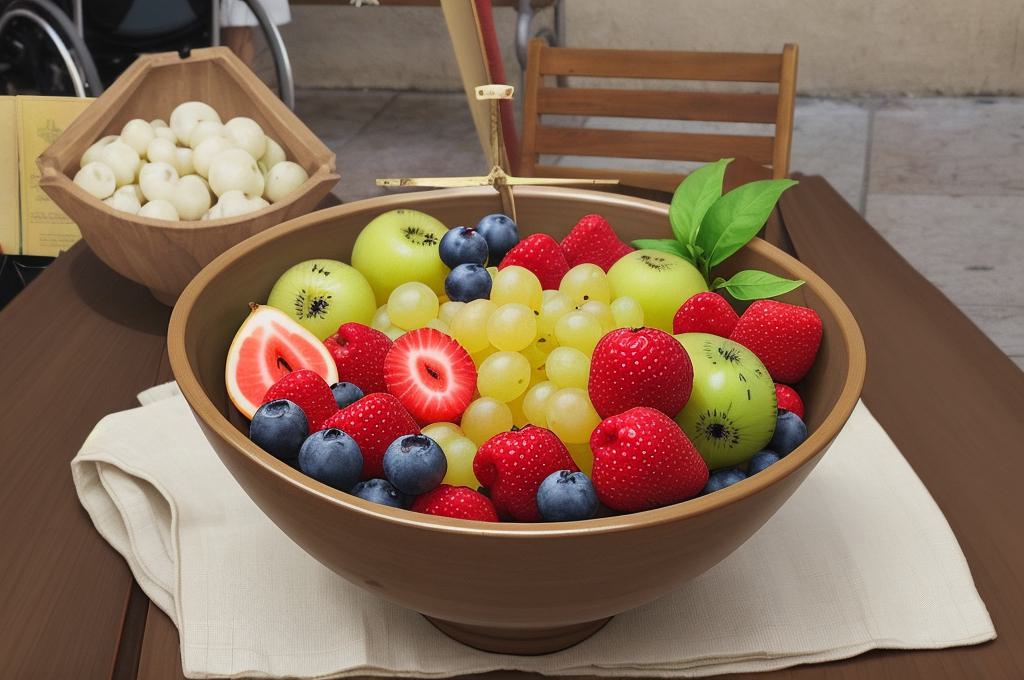}
            \caption{``a bowl of fresh fruits''}
        \end{subfigure}
        \hfill
        \begin{subfigure}{0.32\textwidth}
            \includegraphics[width=0.49\linewidth]{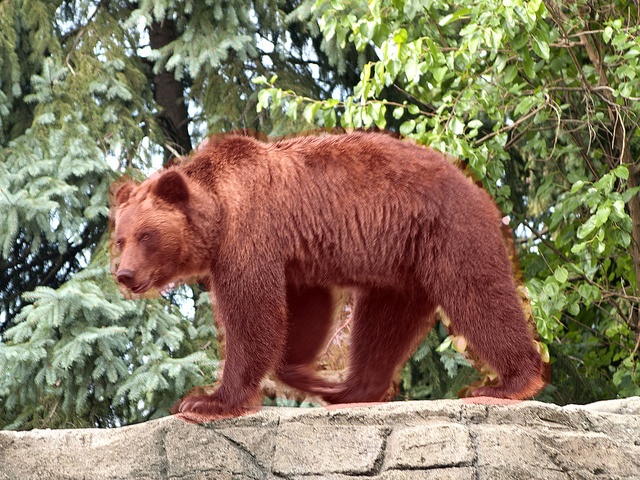}
            \includegraphics[width=0.49\linewidth]{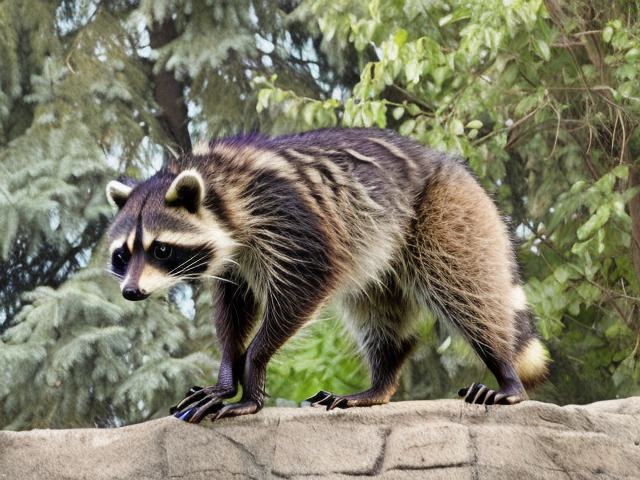}
            \caption{``a friendly raccoon walking across a stone wall near trees''}
        \end{subfigure}
    \end{subfigure}
    
    \caption{Examples of failure cases in inpainting. Row 1: LLM-generated prompts that fail to match the image context. Row 2: Technically sound inpainting results that generate improbable real-world scenarios. Row 3: Results with visible artifacts and blurs. Row 4: Realistic inpaintings that don't follow the given prompts. Row 5: Cases where the inpainting fails to produce coherent results. Row 6: Realistic but uncanny results that human observers can potentially identify as artificial.}
    \label{fig:failure_cases}
\end{figure*}

Figures \ref{fig:model_comparison_examples} and \ref{fig:failure_cases} present a qualitative analysis of some cases from our dataset. In Figure \ref{fig:model_comparison_examples}, we show successful inpainting examples across different models and datasets (COCO, RAISE, and OpenImages), where the models correctly follow the prompts while producing realistic results.

Figure \ref{fig:failure_cases} presents different failure modes. The top two rows reveal problems with LLM-generated prompts, showing cases where prompts either fail to match the scene context or lead to technically sound but unrealistic results. Row 3 demonstrates technical limitations with visible artifacts and blurs. Row 4 presents cases where the inpainting appears realistic but deviates from the given prompt. Row 5 shows examples of poor inpainting quality where the models fail to generate coherent content. Finally, row 6 illustrates a subtle failure mode where the inpainting is technically well-executed but produces results that appear unnatural to human observers upon closer inspection. While these cases might be easily identifiable as manipulated by careful observers, they could potentially deceive viewers who are not actively looking for signs of manipulation, highlighting the importance of including such examples in inpainting datasets for developing robust detection methods.

\end{document}